\documentclass[a4paper,12pt,letterpaper]{report}

\usepackage[round]{natbib}
\usepackage{url}
\usepackage{microtype}
\usepackage[utf8]{inputenc}
\usepackage[T1]{fontenc}
\usepackage[frenchb]{babel}
\usepackage[left=25mm,right=20mm,headheight=14.61858pt,top=25mm,bottom=25mm,inner=2.5cm,bindingoffset=0cm]{geometry}
\usepackage{graphicx}
\usepackage{subfig}
\usepackage{indentfirst}
\usepackage{pifont}
\usepackage{multirow}
\usepackage{xcolor}
\usepackage[nottoc]{tocbibind}
\usepackage[Lenny]{fncychap}
\usepackage{amssymb}
\usepackage[flushleft]{threeparttable}

\usepackage{fancyhdr} 
\begin{document}

\begin{titlepage}
\center\footnotesize RÉPUBLIQUE TUNISIENNE\\
\vspace{0.1em}\footnotesize MINISTÈRE DE L'ENSEIGNEMENT SUPÉRIEUR ET DE LA RECHERCHE SCIENTIFIQUE\\
\vspace{0.1em}\footnotesize UNIVERSITÉ DE TUNIS EL MANAR\\
\vspace{0.1em}\footnotesize FACULTÉ DES SCIENCES DE TUNIS\vspace{0.1em}

\begin{figure}[h!]
\centering
\includegraphics[width=0.2\textwidth]{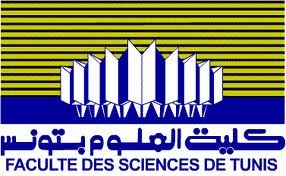}
\end{figure}

\vspace{3em}
{\bfseries\huge Mémoire de Mastère} \\
\vspace{0.25em}
{\slshape\normalsize Présenté en vue de l’obtention du diplôme de } \\
\vspace{0.25em}
{\textrm{\normalsize\textbf{Mastère de recherche en Informatique}}} \\ 
\vspace{1.9em}
{\slshape\normalsize Par } \\
\vspace{1.9em}
{\bfseries\large Inès \textsc{Osman}} \\
\vspace{5em}\hrule

{\textrm{\Huge\textbf{Proposition d'une nouvelle méthode pour l'intégration sémantique des ontologies OWL en utilisant des alignements}}}\medskip

\hrule

\vspace{7.5em}
\textbf{Soutenu le 19/02/2018 devant le jury composé de :} \\ \vspace{1em}

\begin{tabular}{p{6cm}p{6.5cm}l}
\textbf{Mme Hajer \textsc{Baazaoui}}&Professeur à l’ISAMM&Présidente\vspace{2px}\\
\textbf{M. Sami \textsc{Zghal}}&Maître-Assistant à la FSEGJ &Rapporteur\vspace{2px}\\
\textbf{M. Sadok \textsc{Ben Yahia}}&Professeur à la FST&Directeur de mémoire\vspace{2px}
\end{tabular}

\vspace{5mm}
\textbf{Au sein du laboratoire : LIPAH}

\vspace{7.5em}

\center\large{2016-2017}
\end{titlepage}


\chapter*{}
\thispagestyle{empty}

{\fontfamily{pzc}\selectfont
\hfill À mes parents

\hfill À ma sœur

\hfill À toute ma famille

\hfill Et mes amis
}
\chapter*{Remerciements}
\thispagestyle{empty}

Je remercie Monsieur Sadok \textsc{Ben Yahia}, Professeur à la Faculté des Sciences de Tunis et directeur du Laboratoire d'Informatique en Programmation, Algorithmique et Heuristique (LIPAH), pour la confiance qu’il m’a accordée en acceptant de diriger mes travaux de mastère. Je remercie sa disponibilité continue et je voudrais lui éprouver toute mon admiration.\bigskip

Je remercie aussi Monsieur Marouen \textsc{Kachroudi}, Assistant à l'Institut Supérieur des Langues Appliquées et d’Informatique de Béja, pour les idées qu'il m'a données tout au long de ce mémoire. C’est grâce à son aide que ce manuscrit a pu voir le jour.\bigskip

Je tiens à exprimer ma gratitude à Madame Hajer \textsc{Baazaoui}, Professeur à l'Institut Supérieur des Arts Multimédia de la Manouba, pour l’honneur qu’elle m’a fait en acceptant de présider le jury de soutenance.\bigskip

Mes remerciements s’adressent également à Monsieur Sami \textsc{Zghal}, Maître-Assistant à la Faculté des Sciences juridique Économique et de Gestion de Jendouba, d’avoir accepté de rapporter mon travail avec patience et pertinence.

\newpage
\pagenumbering{roman}
\setcounter{tocdepth}{2}
\tableofcontents
\listoffigures
\listoftables

\chapter*{Liste des acronymes}
\addcontentsline{toc}{chapter}{Liste des acronymes}
\markboth{LISTE DES ACRONYMES}{}

\begin{center}
\begin{tabular}{p{3.5cm}l}
\textbf{ALCOMO} & \textbf{A}pplying \textbf{L}ogical \textbf{C}onstraints \textbf{o}n \textbf{M}atching \textbf{O}ntologies
\\
\textbf{AML} & \textbf{A}greement\textbf{M}aker\textbf{L}ight
\\
\textbf{API} & \textbf{A}pplication \textbf{P}rogramming \textbf{I}nterface
\\
\textbf{AROMA} & \textbf{A}ssociation \textbf{R}ule \textbf{O}ntology \textbf{M}atchinh \textbf{A}pproach
\\
\textbf{CIDER} & \textbf{C}ontext and \textbf{I}nterface base\textbf{D} ontology align\textbf{ER}
\\
\textbf{ContentMap} & Logi\textbf{C}-based \textbf{ON}tology in\textbf{TE}gratio\textbf{N} \textbf{T}ool using \textbf{MAP}pings
\\
\textbf{DTD} & \textbf{D}ocument \textbf{T}ype \textbf{D}efinition
\\
\textbf{FITON} & \textbf{F}ramework for \textbf{I}n\textbf{T}egrating \textbf{ON}tologies
\\
\textbf{F-logic} & \textbf{F}rame logic
\\
\textbf{FMA} & \textbf{F}oundational \textbf{M}odel of \textbf{A}natomy
\\
\textbf{HTML} & \textbf{H}yper\textbf{T}ext \textbf{M}arkup \textbf{L}anguage
\\
\textbf{ILIADS} & \textbf{I}ntegrated \textbf{L}earning \textbf{I}n \textbf{A}lignment of \textbf{D}ata and \textbf{S}chema
\\
\textbf{IRI} & \textbf{I}nternationalized \textbf{R}esource \textbf{I}dentifier
\\
\textbf{KIF} & \textbf{K}nowledge \textbf{I}nterchange \textbf{F}ormat
\\
\textbf{LGPL} & \textbf{L}esser \textbf{G}eneral \textbf{P}ublic \textbf{L}icence
\\
\textbf{LOD} & \textbf{L}inked \textbf{O}pen \textbf{D}ata
\\
\textbf{LogMap} & \textbf{L}ogic-based \textbf{M}ethods for \textbf{O}ntology \textbf{M}apping
\\
\textbf{NCI} & \textbf{N}ational \textbf{C}ancer \textbf{I}nstitute \textbf{T}hesaurus
\\
\textbf{OAEI} & \textbf{O}ntology \textbf{A}lignment \textbf{E}valuation \textbf{I}nitiative
\\
\textbf{OIM-SM} & \textbf{O}ntology \textbf{I}ntegration \textbf{M}ethod based on \textbf{S}emantic \textbf{M}apping
\\
\textbf{OLA} & \textbf{O}WL \textbf{L}ite \textbf{A}lignment
\\
\textbf{OWL} & \textbf{O}ntology \textbf{W}eb \textbf{L}anguage
\\
\textbf{RDF} & \textbf{R}esource \textbf{D}escription \textbf{F}ramework
\\
\textbf{RDFS} & \textbf{RDF}-\textbf{S}chema
\\
\textbf{SWRL} & \textbf{S}emantic \textbf{W}eb \textbf{R}ule \textbf{L}anguage
\\
\textbf{SNOMED CT} & SNOMED \textbf{C}linical \textbf{T}erms
\\
\textbf{URI} & \textbf{U}niform \textbf{R}esource \textbf{I}dentifier
\\
\textbf{URL} & \textbf{U}niform \textbf{R}esource \textbf{L}ocator
\\
\textbf{WWW} & \textbf{W}orld \textbf{W}ide \textbf{W}eb
\\
\textbf{W3C} & \textbf{WWW} \textbf{C}onsortium
\\
\textbf{XML} & \textbf{E}xtensible Markup \textbf{L}anguage
\\
\textbf{YAM}++ & \textbf{Y}et \textbf{A}nother \textbf{M}atcher
\\

\end{tabular}
\end{center}

\pagenumbering{arabic}
\chapter*{Introduction générale}
\addcontentsline{toc}{chapter}{Introduction générale}
\markboth{INTRODUCTION}{}

L’intégration des données est un vaste domaine qui permet d'unifier les données provenant des sources hétérogènes partageant des informations en commun, ou qui permet de les transférer d’une représentation à une autre, pour but de faire l’échange entre différents systèmes. Elle concerne des sources de données telles que les bases de données, les fichiers textes, et les ontologies, \textit{etc.}\medskip\vspace{4px}

Pour ce faire, un sous-domaine a fait son apparition. Il s'agit de l'intégration des schémas tels que les schémas relationnels, orientés objet, XML (DTD, XML Schema), \textit{etc.}). Rappelons que le schéma ou le modèle des données permet de décrire avec précision la structure d'un document, les conventions de structuration, de typage, et de nommage de ses données.\medskip\vspace{4px}

Par ailleurs, les ontologies ont été reconnues comme une composante essentielle pour la concrétisation de la vision du Web Sémantique. En définissant et décrivant les termes associés à des domaines particuliers, elles permettent d'annoter ou d'attacher les termes de multiples documents avec les mêmes termes propres à elles, ainsi elles arrivent à intégrer le contenu de différentes sources des données telles que les pages Web, les documents XML, les bases de données relationnelles, \textit{etc}. L'utilisation de ces terminologies partagées permet un certain degré d'interopérabilité entre ces sources de données.\medskip\vspace{4px}

Cependant, cela ne résout pas complètement le problème d'intégration des données, car nous ne pouvons pas s'attendre à ce que tous les individus et toutes les organisations dans le Web sémantique s'accordent sur l'utilisation d'une terminologie ou d'une ontologie commune. Par conséquent, il est peu probable qu'une ontologie globale couvrant l'ensemble des systèmes distribués puisse être développée; au contraire, un domaine donné pourrait avoir plusieurs ontologies concurrentes, chacune incomplète ou couvrant le domaine d'une certaine perspective. En effet, dans la pratique, les ontologies de différents systèmes sont développées indépendamment les unes des autres, par des communautés différentes, et pour des buts différents.\medskip\vspace{4px}

Suite à ce problème d'hétérogénéité, le domaine de l'intégration des ontologies, qui est aussi un sous-domaine de l'intégration des données, a fait son apparition. D'ailleurs, il ressemble énormément au domaine de l'intégration des schémas des bases de données, car les approches récentes de ces deux domaines se composent toutes les deux de deux étapes principales : l'étape de matching qui va réconcilier les différences en déterminant des correspondances (les similarités et les différences), puis l'étape de fusion (ou d'union) qui va exploiter le résultat du matching.\medskip\vspace{4px}

L’intégration des ontologies de différents domaines vise à la construction d’une nouvelle ontologie pour un nouveau domaine plus large composé des domaines des ontologies intégrées. Elle est aussi appelée "composition" d'ontologies. L’intégration des ontologies de mêmes domaines vise à les unifier pour obtenir une ontologie plus complète qui couvre mieux ce même domaine. Elle est appelée "fusion" d’ontologies.\medskip\vspace{4px}

En général, les ontologies peuvent couvrir des domaines différents, ou bien des domaines identiques, proches (liés), complémentaires, ou interdisciplinaires dans lesquels les termes se chevauchent et les niveaux de détail (de leur conceptualisation) diffèrent. Ainsi, si les connaissances et les données doivent être partagées (\textit{e.g.} dans le Web, ou par des entreprises en collaboration), il faudrait au moins établir des correspondances sémantiques ou des liens entre les ontologies qui les décrivent (matching des ontologies).\medskip\vspace{4px}

La tâche d'intégration des ontologies (composée d’une étape de matching, puis d’une étape d’union) est particulièrement importante dans les systèmes d'intégration puisqu'elle autorise la prise en compte conjointe des ressources décrites par des ontologies différentes. Ce thème de recherche a donné lieu à de très nombreux travaux.\medskip\vspace{4px}

Dans un contexte plus large, les ontologies produites sur le Web peuvent être graduellement d'une très grande taille. Ainsi, le processus d’intégration des ontologies nécessitera l'utilisation de mécanismes de prise en charge pour le passage à l'échelle des techniques d'intégration.\medskip\vspace{4px}

Pour conclure, l'idée de base est de réconcilier et d'intégrer des ontologies ou des fragments d'ontologies pour en fédérer d'autres qui encapsulent les données des ontologies initiales. Ainsi, l'objectif de ce mastère est de proposer une nouvelle méthode d'intégration des ontologies qui cherche à mettre en place une solution capable de raisonner sur des ontologies déjà existantes pour en produire une nouvelle en utilisant des techniques d'intégration.

\section*{Motivations et contributions :}
Les techniques actuelles d'intégration des ontologies sont encore non efficaces en termes de temps d'exécution, semi-automatiques (qui reposent beaucoup sur l’intervention humaine), non extensibles (non scalables), générant une ontologie de mauvaise qualité (ayant énormément de contradictions sémantiques / logiques), et non complètes (avec perte d'informations précieuses, car ils n’arrivent pas à préserver toutes les connaissances des ontologies sources surtout les disjonctions).\medskip\vspace{4px}

Dans ce mémoire, nous proposons de développer une méthode automatique d’intégration des ontologies (OIA2R) ayant pour but d’intégrer deux ou plusieurs ontologies (de toute taille) en utilisant les mappings (ou les alignements) entre elles, pour former à la fin une nouvelle ontologie qui conserve toutes les informations des ontologies sources et des mappings, tout en les personnalisant (refactoring). Autrement dit, il s’agit d’une ontologie de pont qui englobe les ontologies d’entrée et les "bridging" axiomes qui les réconcilient. Notre algorithme produit une ontologie de sortie de bonne qualité en des temps d’exécution compétitifs. Nous proposons aussi une nouvelle classification des terminologies ambiguës utilisées dans le domaine de l'intégration des ontologies, car il y a une très grande confusion dans les appellations de l'état de l'art.

\section*{Structure du mémoire :}

Ce mémoire se compose de quatre chapitres :\medskip\vspace{4px}

Le \textbf{premier chapitre} introduit les notions essentielles pour cerner le champ d’étude. En effet, une étude bibliographique sur le Web sémantique, l’ontologie, ses constituants, ainsi que le langage OWL qui contribue à sa description. La section suivante élucide la notion d’ingénierie des ontologies et liste ses différents domaines dont l’intégration des ontologie fait partie.\medskip\vspace{4px}

Le \textbf{deuxième chapitre} présente les notions de fusion et d’intégration des ontologies, leurs différentes définitions et leurs approches existantes dans la littérature. Ensuite, nous introduisons les problèmes qui peuvent résulter de ces deux processus.\medskip\vspace{4px}

Le \textbf{troisième chapitre} est consacré à la présentation de notre nouvelle méthode. Il présente les différents modules qui la constitue en s'aidant par des schémas explicatifs, puis montre les critères nécessaires pour avoir de meilleurs résultats.\medskip\vspace{4px}

Le \textbf{quatrième chapitre} présente les critères d’évaluation du domaine de l’intégration des ontologies, ainsi que l’environnement de réalisation du prototype développé, et les expérimentations qui s’intéressent à l’analyse et la critique du résultat de notre méthode.\medskip\vspace{4px}

Le mémoire se termine par une conclusion générale qui résume l’ensemble de nos travaux et qui présente quelques perspectives futures de recherche.
\chapter{Fondement du Web sémantique}

\section*{Introduction}
Dans ce chapitre, nous rappelons les grandes étapes de l’évolution du Web, citons la définition du terme \textit{Web sémantique} et \textit{Ontologie}, présentons le langage OWL, rappelons les types d’hétérogénéité entre les ontologies, et présentons tous les domaines de l’ingénierie des ontologies et leurs applications dans le monde réel. Enfin, nous terminons ce chapitre par une conclusion qui introduit les causes du recours aux domaines de l’intégration et la fusion des ontologies dans lesquels nous allons entrer en détail dans le chapitre 2.

\section{Notion du Web sémantique}
\subsection{Introduction au Web sémantique}
Le Web actuel est un ensemble de documents (données et pages) dédiés aux humains, stockés et manipulés d’une façon purement syntaxique. Voici les deux principaux problèmes du Web actuel.\medskip\vspace{4px}

D’une part, il y a énormément de sources de données, du fait que n’importe qui peut facilement publier un contenu (sachant qu’il n’a pas la moindre idée sur la probabilité que ce contenu soit trouvé par autrui) ; il n’a qu’à l’annoter, ainsi les moteurs de recherche à base de mot-clé auront la tâche de l’indexer pour pouvoir l’afficher aux utilisateurs lorsqu’ils font une recherche. Par conséquent, l'information sur Internet est tellement énorme que l’utilisateur a du mal à la retrouver.\medskip\vspace{4px}

D’une autre part, les résultats de recherche sont imprécis, très sensibles au vocabulaire, et assez longs à trouver. En effet, les moteurs de recherche ne sont capables de répondre qu’à deux questions principales :
\begin{itemize}
\item Quelles sont les pages contenant ce terme ? et ;
\item Quelles sont les pages les plus populaires à ce sujet ?\medskip\vspace{4px}
\end{itemize}
Le Web est essentiellement syntaxique, et l’Homme est le seul à pouvoir interpréter son contenu (des documents et des ressources) inaccessible et non interprétable par la machine ; lui seul doté de la capacité de comprendre ce qu’il a trouvé et décider en quoi cela se rapporte à ce qu’il veut vraiment chercher. Finalement, nous ne pourrons pas se passer de l’intervention humaine pour naviguer, chercher, faire le tri des documents manuellement, interpréter, et combiner les résultats.\medskip\vspace{4px}

Pour conclure, le Web actuel ne peut pas être manipulé de façon intelligente par les programmes informatiques car il y a un vrai manque de sémantique.\medskip\vspace{4px}

Voici un exemple qui illustre ces problèmes : Supposons que nous voulons rechercher un fabricant de portes et de fenêtres pour construire une maison, nous tapons les mots "gates" et "windows" dans Google, nous aurons des résultats non satisfaisants concernent en grande partie Bill Gates et Microsoft Windows. Idéalement, les résultats devraient contenir les deux sens équitablement, ou selon le contexte de l’utilisateur.

\subsection{Objectifs du Web sémantique}
L'intérêt croissant porté à la recherche d'information sur le Web a donné lieu à l'initiative du Web sémantique. De nos jours, le souci du Web n'est plus vraiment l’augmentation continuelle de sa taille d’informations, mais plutôt l’amélioration de la recherche dans cette énorme masse d’informations, et la réalisation de systèmes permettant de filtrer et délivrer les informations de façon "intelligente".\medskip\vspace{4px}

Le but ultime du Web de troisième génération est de permettre aux utilisateurs d’exploiter tout le potentiel du Web en s’aidant par les machines qui pourront accomplir les tâches encore réalisées par l’Homme comme la recherche ou l'association d'informations, et ainsi atteindre un Web intelligent qui regroupera l'information de manière utile et qui apportera à l’utilisateur ce qui cherche vraiment.

\subsection{Définition du Web sémantique}
En 1993, Tim Berners-Lee a fournit une solution au problème du partage de connaissances entre les applications Web à l’aide d’un mécanisme à base d’ontologies qui structure les données d’une manière compréhensible par la machine. En 2001, il a envisagé un WWW accessible pour les machines et les humains, de telle sorte qu’ils soient mis dans une position égale.\medskip\vspace{4px}

Le Web sémantique (nommé aussi Web intelligent ou Web des données) est un ensemble de connaissances, où toutes les machines peuvent lier sémantiquement les données du Web, ainsi comprendre leurs significations, y accéder plus intelligemment, pour améliorer le dialogue entre les applications et l'interaction avec l’utilisateur en lui offrant une meilleure qualité des tâches de recherche (d'association des informations, et d'apprentissage, \textit{etc.}).\medskip\vspace{4px}

Il peut être vu aussi comme une couche supplémentaire de connaissances (au-dessus du Web actuel) ou une extension du Web actuel.\medskip\vspace{4px}

De la même manière que le Web actuel, le Web sémantique est construit principalement autour des identifiants (URIs) et du protocole HTTP, mais il est par contre basé sur le langage RDF et non plus sur le HTML, pour but de séparer l’information qui décrit le sens et le contexte des données, de l’information qui décrit la présentation des données. Il est basé principalement sur les bases de connaissances et non seulement sur les bases de données. La recherche aussi va s’affecter et devenir une recherche par concept, non plus par mot clé.\medskip\vspace{4px}

Dans le Web sémantique, toutes les données du Web, textuelles ou multimédia, doivent être annotées sémantiquement par des métadonnées pertinentes, car les machines (les agents logiciels) ne pourront comprendre les données et prendre des décisions qu’à travers une explication plus spécifique du contenu, et cela en utilisant un mark-up sémantique nommé "méta-données". L’annotation de ces ressources d’information repose sur l’accès à des représentations de connaissances (des ontologies) partagées sur le Web.\medskip\vspace{4px}

Pour résumer, le Web sémantique donnera naissance à un nouvel aspect intelligent basé sur la recherche, le raisonnement, et la prise de décision automatique, faisant ainsi croître la productivité et les capacités des moteurs de recherche.

\section{Ontologie}
\subsection{Ontologies et Web}
Le domaine des ontologies est né d’une volonté de pallier les limites du Web (déjà évoquées). Les ontologies font partie intégrante des normes du W3C pour le Web sémantique, car elles sont indispensables pour représenter la sémantique des documents (les connaissances) qui coexistent dans le Web, en structurant et en définissant la signification des termes actuellement collectées et normalisées. En effet, les ressources du Web telles que les pages Web, les bases de données, ou les documents XML, \textit{etc.} sont annotées par (attachées à) la signification des termes (concepts) de sorte que nous aurons besoin du même concept de la même ontologie pour représenter la même chose dans l'indexation de ces différentes ressources. C'est ici que se manifeste le rôle et l'utilité des ontologies.\medskip\vspace{4px}

Elles sont utilisées pour publier des bases de connaissances réutilisables et faciliter l'interopérabilité entre plusieurs systèmes hétérogènes et bases de données. Ainsi, nous pouvons considérer les ontologies comme une représentation pivot qui a pour but d’intégrer les sources de données hétérogènes.\medskip\vspace{4px}

Elles sont utilisées dans beaucoup de filières telles que : la gestion des connaissances, l’intelligence artificielle, ou le Web sémantique. Et elles aident à réaliser de nombreuses applications comme la recherche d'informations, la réponse aux requêtes, la recherche documentaire, et la synthèse de texte, \textit{etc.}\medskip\vspace{4px}

Nous pouvons conclure que l’ontologie est un outil essentiel permettant l’exploitation automatique (le traitement machine) des connaissances, et la concrétisation des principes de réutilisabilité et du partage de l’information entre différentes sources de données, et cela grâce au vocabulaire commun fourni pour un domaine de connaissances réel ou imaginaire.\medskip\vspace{4px}

\cite{caldarola2016} constatent que les ontologies disponibles dans la littérature sont en train de devenir de plus en plus volumineuses en termes de nombre d’entités, à un tel point qu'elles peuvent être considérées comme de la Big Data.

\subsection{Revue des définitions d’une ontologie}
L’ontologie est un terme qui est apparu dans la Métaphysique avec Aristote qui considérait que l’ontologie est une \textit{"Science qui étudie l’être en tant qu’être et les attributs qui lui appartiennent essentiellement"}. Dans ce contexte, élaborer une ontologie, revient à faire l’étude philosophique de la nature de l’être et de l’existence, \textit{i.e.} l’étude des propriétés générales de ce qui existe, en définissant l’ensemble des connaissances sur le monde.\medskip\vspace{4px}

Pendant la dernière décennie, les informaticiens ont repris le terme "Ontologie" qui est devenu très utilisé dans le domaine de l’informatique. C’est au début des années 90 qu’il est apparu pour la première fois dans le cadre des recherches sur les Systèmes à Base de Connaissances (SBC).\\

Une des premières définitions a été donnée par \cite{neches1991enabling} : \textit{"Une ontologie définit les termes et les relations de base comportant le vocabulaire d’un domaine, aussi bien que les règles pour combiner ces termes et ces relations afin de définir des extensions du vocabulaire"}.\\

\cite{studer1998knowledge} ont conclu qu’\textit{"une ontologie est une spécification formelle et explicite d’une conceptualisation partagée d'un domaine de connaissances"}.\vspace{4px}

\begin{itemize}
\item[$\diamond$] Le terme \textit{"conceptualisation"} ou conceptualiser un domaine veut dire faire une abstraction décrivant un phénomène quelconque du monde réel de ce domaine ; faire les choix quant à la manière de décrire ce domaine particulier (par des entités).\vspace{4px}
\item[$\diamond$] Une \textit{"spécification"} est une conceptualisation représentée dans une forme concrète. Une spécification de la conceptualisation est par conséquent une définition formelle des termes qui décrivent un domaine, des relations entre eux, et des axiomes qui les contraignent (Nous en parlerons en détail juste après).\vspace{4px}
\item[$\diamond$] Le terme \textit{"formelle"} signifie qu’une ontologie doit être interprétable et lisible par la machine.\vspace{4px}
\item[$\diamond$] Le terme \textit{"explicite"} veut dire que les entités et les axiomes doivent être explicitement définis.\vspace{4px}
\item[$\diamond$] Le terme \textit{"partagée"} indique qu’une ontologie doit annoter multiples sources de données, être consensuelle et accessible par tous les utilisateurs
d’une communauté particulière.\\
\end{itemize}

\cite{gruber2009encyclopedia} définissent une ontologie comme suit : \textit{"Dans le contexte des sciences de l’informatique et de l’information, une ontologie définit un jeu de primitives représentatives avec lequel un domaine de connaissance ou un univers de discours peut être modélisé"}.\vspace{4px}

\begin{itemize}
\item[$\diamond$] Le terme "jeu de primitives" est la traduction la plus fidèle possible du monde réel à représenter.
\end{itemize}

\subsection{Constituants d’une ontologie}

Une ontologie est une collection structurée de termes, de relations entre les termes, et d'un ensemble de règles d'inférence sur ces termes. Elle est nommée avec un IRI. Et puisqu’elle est un document Web, elle est ainsi référencée par un URI (IRI physique) qui doit pointer sur (coïncider avec) la localisation de l’URL choisi pour la publier.\medskip\vspace{4px}

Dans la syntaxe abstraite, une ontologie OWL est une séquence d'axiomes (de règles ou de contraintes) logiques et non logiques (y compris les faits), et éventuellement de références à d'autres ontologies (des importations) qui sont considérées incluses dans l'ontologie.\medskip\vspace{4px}

La particularité des ontologies réside dans l'existence d'une sémantique (de théorie) de logique mathématique. En effet, les relations entre les entités peuvent être formellement modélisées par la logique de description formelle de premier ordre.\medskip\vspace{4px}

L’ontologie est formalisée par des entités pouvant avoir chacune un IRI qui est une référence d'URI. Il existe cinq types d’entités : les \textbf{concepts} (ou classes), les \textbf{propriétés} (relations, attributs, slots, rôles, ou actions), les \textbf{individus} (objets, instances, ou extensions des classes), les \textbf{types de données}, et les \textbf{valeurs de données}. La déclaration de ces entités dans l’ontologie est faite par des axiomes non logiques :\\

$\bullet$ Les \textbf{concepts sémantiques} de l’ontologie correspondent aux abstractions d’une partie de la réalité (du domaine). Ce sont les concepts auxquels nous nous référons, choisis en fonction des objectifs que nous nous donnons et de l’application envisagée pour l’ontologie. Ils sont les entités principales d’une ontologie. Ils peuvent représenter des concepts abstraits (une notion, une intention, une idée, une croyance, un sentiment, \textit{etc.}), ou bien des concepts spécifiques (un objet matériel, un ensemble ou un groupe d’individus de caractéristiques similaires, \textit{etc.}).\vspace{4px}

\ding{220} Les concepts sont organisés hiérarchiquement à travers la relation conceptuelle "Sous classes de" ou "is a" d’héritage ou de spécialisation, utilisée pour construire une taxonomie / hiérarchie de concepts ; Cette relation peut aussi signifier une relation d’agrégation ou de composition "Partie de" ou "has a". D’autres relations prédéfinies telle que l’équivalence et la disjonction peuvent également lier les concepts pour véhiculer plus de sémantique.\\

$\bullet$ Les \textbf{propriétés} permettent de définir des liens pour les individus présents dans le domaine. Les propriétés sont des relations non prédéfinies et non taxonomiques utilisées pour exprimer la sémantique qui relie deux concepts, et c’est justement l’apport des ontologies qui peuvent définir d’autres relations spécifiques non prédéfinies.\medskip\vspace{4px}

\begin{itemize}
\item[$\circ$] Le premier type de propriétés, nommé \textbf{propriété d’objet}, est défini tel que le premier argument de la relation corresponde au domaine (un concept pour lequel est définie la propriété) et que le deuxième argument corresponde au co-domaine / à l’image (un concept relié au domaine par la propriété). Ainsi, il définit une relation entre deux individus.\vspace{4px}

\item[$\circ$] Le deuxième type de propriétés, nommé \textbf{propriété de type de données}, est utilisé pour exprimer les attributs des concepts. Les attributs sont des relations dans lesquelles le domaine est un concept, et le co-domaine / l’image est un type de donnée (un littéral) tel que "String", "Integer", "Double", "Date", \textit{etc.} Ainsi, il définit une relation entre un individu d’une classe et une valeur de donnée. \vspace{4px}
\end{itemize}

\ding{220} Ces deux types de propriétés peuvent être organisés hiérarchiquement, et liés par des relations conceptuelles prédéfinies telles que l’équivalence, la disjonction, et beaucoup d’autres. Ces propriétés sont instanciées à l’aide de la relation d’affectation qui leur associe une valeur de domaine (un individu) et une valeur de co-domaine (un individu ou une valeur de donnée).\medskip\vspace{4px}

\begin{itemize}
\item[$\circ$] Le troisième type de propriétés, nommé \textbf{propriété d’annotation}, ne se conforme pas à la définition des propriétés décrite ci-dessus. Le rôle de cette propriété est d’annoter les entités ou les ontologies. Son domaine peut être une entité (classe, propriété, ou individu) ou une ontologie, et son co-domaine peut être une entité, un littéral généralement de type "String", ou une ontologie.\vspace{4px}
\end{itemize}

\ding{220} Les propriétés d’annotation peuvent également être organisées hiérarchiquement.\\

$\bullet$ Les \textbf{individus} constituent la définition extensionnelle / l’extension des concepts, et ainsi l’extension (les données) de l’ontologie. Les IRIs des individus sont utilisés pour faire référence aux ressources. Ce sont des objets particuliers instanciés par les concepts à l’aide de la relation d’instanciation prédéfinie "Instance de" ou "is kind of" ou "type". Ils peuplent les classes et véhiculent les connaissances à propos du domaine. (Il existe aussi des \textbf{individus anonymes} qui sont des individus non utilisés en dehors de l’ontologie. Ils sont identifiés par un ID local plutôt qu'un IRI global).\vspace{4px}

\ding{220} Les instances peuvent être liées par des relations conceptuelles d’identité et de différence.\\

$\bullet$ Les \textbf{types de données} sont des parties particulières du domaine qui spécifient des valeurs. Les ontologies référencent des types de données intégrés de XML Schema (des littéraux) au moyen d'une référence URI à ce type de données.\\

$\bullet$ Les \textbf{valeurs de données} sont, contrairement aux autres entités, des valeurs simples qui n'ont pas d'IRI. Ce sont les valeurs des types de données.\\

$\bullet$ Les \textbf{axiomes logiques} constituent des assertions liées aux entités. Au lieu de compter sur les labels et les termes des entités (qui sont destinés aux humains) pour transmettre la sémantique, le concepteur d’ontologies doit contraindre l’interprétation possible des entités à travers une utilisation judicieuse d’axiomes logiques pour rende leurs sens beaucoup plus précis.\vspace{4px}

\ding{220} Ils sont aussi utilisés pour vérifier la consistance de l’ontologie, car ils permettent à un "raisonneur" d’inférer des connaissances additionnelles qui ne sont pas déclarées directement. Plus les axiomes exprimés dans les ontologies sont complexes, plus ils transportent des connaissances implicites qui peuvent être inférées par le raisonneur.\\

$\bullet$ Les \textbf{faits} sont des axiomes qui énoncent des informations sur les individus, telles que les classes auxquelles les individus appartiennent, et les propriétés et les valeurs des propriétés de ces individus.

\subsection{Définitions formelles d'une ontologie}

Selon \cite{kalfoglou2003ontology}, une approche algébrique plus formelle identifie une ontologie comme étant une paire <S, A>, où S est la \textbf{signature} des entités de l'ontologie (modélisée par une structure mathématique comme un treillis ou un ensemble non structuré) et A est l'\textbf{ensemble des axiomes} ontologiques qui spécifient l'interprétation voulue de la signature dans un domaine donné.\medskip\vspace{4px}

\cite{udrea2007leveraging}, les ontologies modélisent la structure des données (\textit{i.e.}, les ensembles de classes et de propriétés), la sémantique des données (sous la forme d’axiomes tels que les relations d'héritage ou les contraintes sur les propriétés), et les instances des données (les individus). Ainsi, les entités d'une ontologie se composent d’une partie "\textbf{structure}", et d’une partie "\textbf{donnée}".\medskip\vspace{4px}

Selon \cite{cheatham2017semantic}, les informations des classes, des propriétés, et des axiomes qui restreignent leur interprétation, sont appelées la "structure", le "schéma", ou "\textbf{T-box}" (comme Terminologie) de l’ontologie, et les informations des instances et leurs axiomes sont appelées "données", "données d’instances" ou \textbf{"A-box"} (comme Assertions) et contiennent des assertions sur des instances utilisant des données du T-box.\medskip\vspace{4px}

D’après \cite{zhang2017oim}, une ontologie est un modèle en arbre, à cause du principe de l’hyponymie (la subsomption – is-a –) qui fait que chaque entité (classe ou propriété) soit héritée d’une seule super-entité directe, formant ainsi une structure de graphe acyclique enracinée \cite{raunich2012towards}. Mais dans le cas d'un héritage multiple, l'ontologie devient un modèle en réseau qui peut contenir des cycles et dans lequel plusieurs chemins peuvent mener à une entité.\\ \\

\begin{figure}[h]
\begin{center}
\includegraphics[scale=0.84]{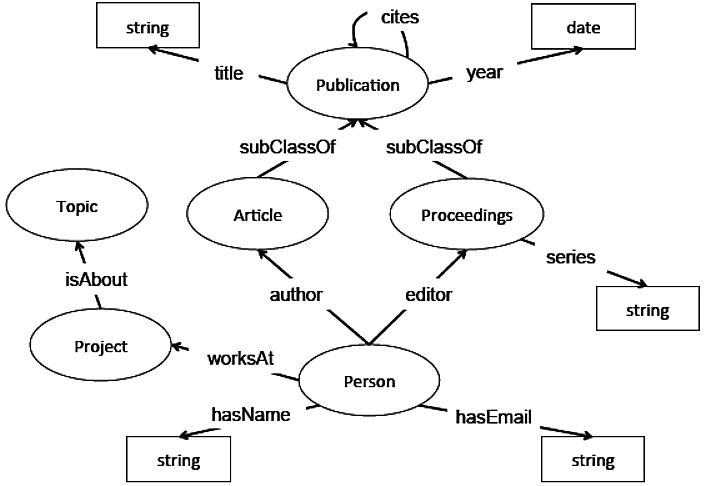}
\vspace{15px}
\caption[Fragment d'une ontologie]{Fragment d'une ontologie \cite{cheatham2017semantic}}
\end{center}
\end{figure}

\section{Langage d'ontologie OWL}
Il existe une grande variété de langages pour exprimer les ontologies. Quelques exemples de langages incluent RDF, RDFS, OWL, KIF, F-Logic, UML, SQL DDL, ou XML Schema, \textit{etc.}\medskip\vspace{4px}

Le défi du Web sémantique est de fournir un langage qui exprime à la fois des règles, des structures, et des données sur lesquelles il va raisonner (à l’aide de ces règles). Par la suite, les règles de n'importe quel système de représentation de connaissances pourront être exportées dans le Web sémantique.\medskip\vspace{4px}

\textbf{OWL} (\textbf{W}eb \textbf{O}ntology \textbf{L}anguage) est un langage de représentation de connaissances qui permet d’écrire (de construire) des ontologies Web, et tout comme RDF, il est un langage profitant de l'universalité syntaxique de XML. OWL devient une recommandation du W3C en 2004, et OWL 2 le devient en 2009.\medskip\vspace{4px}

A part sa capacité de définir et de décrire des classes, des propriétés, et des individus de classes, OWL permet aussi de définir des relations entre les classes (union, intersection, disjonction, équivalence, subsomption \textit{etc.}), des contraintes de cardinalité pour les valeurs des propriétés (nombre minimum, maximum, ou exact), des relations spéciales pour les propriétés (transitive, symétrique, fonctionnelle, inverse, réflexive, \textit{etc.}), et des restrictions sur le domaine et le co-domaine des propriétés, \textit{etc.} Par conséquent, OWL possède une logique très développée qui permet le raisonnement sémantique sur ces règles.\medskip\vspace{4px}

Comparé aux langages RDF et RDFS, OWL offre aux machines une plus grande capacité d'interprétation du contenu Web, grâce à son vocabulaire riche et sa sémantique formelle.\medskip\vspace{4px}

\shorthandoff{:}
Les classes définies par l'utilisateur sont toutes des enfants de la superclasse « owl:Thing » (qui représente l’ensemble de tous les individus) et des parents de la sous-classe « owl:Nothing » (qui représente l’ensemble vide).\medskip\vspace{4px}

Les propriétés d’objet et de type de données définies par l'utilisateur sont toutes respectivement des enfants des super propriétés « owl:TopObjectProperty » et « owl:TopDataProperty », et des parents des sous-classes « owl:BottomObjectProperty » et « owl:BottomDataProperty ». Les propriétés d’annotation telles que « owl:versionInfo », « rdfs:label », « rdfs:comment », « rdfs:seeAlso », « owl:priorVersion » \textit{etc.} sont des constructeurs intégrés dans OWL.\medskip\vspace{4px}

Un individu peut ne pas avoir de classe(s) qui l’instancie(nt) ; dans ce cas, il sera implicitement une instance de la classe « owl:Thing ».\medskip\vspace{4px}
\shorthandon{:}

Il y a une variété de syntaxes (formats) pour persister, partager, et éditer des ontologies OWL, telles que Functional OWL, RDF/XML, Turtle, OWL/XML, Manchester OWL, OBO, KRSS, \textit{etc.} La spécification OWL décrit ce qui constitue une ontologie d'un point de vue structurel de haut niveau, qui est ensuite mappée en diverses syntaxes concrètes. RDF/XML est la syntaxe d'échange officiellement recommandée par W3C, que tout outil OWL doit pouvoir prendre en charge.

\section{Types d’hétérogénéité}

La diversité du monde réel est une source de richesse et d'hétérogénéité. En effet, dans les systèmes ouverts et distribués, tels que le Web sémantique, l'hétérogénéité ne peut pas être évitée. Plusieurs ontologies de mêmes domaines ou de domaines proches peuvent exister, à cause du développement déconnecté qui se focalise sur des applications particulières de différents buts et intérêts. Par ailleurs, les concepteurs ont des habitudes et des pré-requis différents, et modélisent les connaissances avec des niveaux de détails différents et des outils différents. Tout cela va influencer de différentes manières leurs décisions de conception. Par conséquent, la conception des ontologies ne peut jamais être un processus déterministe ; même deux ontologies de même domaine ne vont pas être identiques. Toutes ces raisons mènent à diverses formes d'hétérogénéité.\medskip\vspace{4px}

Prenons l’exemple du domaine biomédical. Il y a neuf ontologies qui décrivent une maladie neurologique, allant des ontologies très spécifiques couvrant une seule maladie (\textit{e.g.} l’épilepsie, l’Alzheimer) à des ontologies couvrant toutes sortes de maladies telles que la "Disease Ontology". Il en résulte plusieurs ontologies qui décrivent les mêmes concepts sous des modèles légèrement différents.\medskip\vspace{4px}

\cite{klein2001combining} distingue deux niveaux d’hétérogénéité qui peuvent exister entre les ontologies :
\subsection{Hétérogénéité des langages}
Ce sont les différences au niveau du langage, du méta-modèle, ou des primitives du langage utilisées pour spécifier une ontologie. Ce sont des différences entre les mécanismes (à partir desquels les entités vont être définies). Nous pouvons classifier ces différences en quatre catégories de difficulté croissante :

\subsubsection{\underline{Syntaxe}}
Les différents langages d'ontologie utilisent souvent des syntaxes différentes, \textit{e.g.}, pour définir la classe de chaises dans RDF Schema (RDFS), nous utilisons {\NoAutoSpacing<rdfs:Class ID="Chair">}. Dans LOOM, l'expression (defconcept Chair) est utilisée pour définir la même classe. L’exemple typique d'incompatibilité de "syntaxe seulement" est quand un langage d'ontologie a plusieurs représentations syntaxiques, comme les différentes syntaxes de OWL.

\subsubsection{\underline{Représentation logique}}
La différence de représentation des notions logiques, \textit{e.g.} dans certains langages, il est possible d'indiquer explicitement que deux classes sont disjointes (A disjoint B), alors que dans d’autres langages, il est nécessaire d'utiliser la négation dans des instructions de sous-classes (A subclass-of (NOT B), (B subclass-of (NOT A)) pour indiquer la disjonction.

\subsubsection{\underline{Sémantique des primitifs}}
Une différence plus subtile au niveau du méta modèle est la sémantique des constructions du langage. Malgré le fait que parfois le même nom est utilisé comme un constructeur dans deux langages, la sémantique peut différer, \textit{e.g.} il existe plusieurs interprétations de A equalTo B. Même lorsque deux langages d'ontologie semblent utiliser la même syntaxe, la sémantique des constructeurs peut différer.

\subsubsection{\underline{Expressivité des langages}}
C’est la différence fondamentale qui a le plus d'impact. Cette différence implique que certains langages sont capables d'exprimer des choses qui ne sont pas exprimables dans d'autres langages, \textit{e.g.} certains langages ont des constructions pour exprimer la négation, d'autres non ; également pour le support des listes, des ensembles, et des valeurs par défaut, \textit{etc.}

\subsection{Hétérogénéité des modèles du domaine}
Ce sont des différences dans la façon dont le domaine est modélisé. Elles sont décrites par \cite{visser1998assessing} :

\subsubsection{\underline{Différence de conceptualisation (de sémantique)}}
C’est une différence dans la façon dont un domaine est interprété (conceptualisé / modélisé), ce qui entraîne différents concepts, différentes relations entre les concepts, ou différentes instances des concepts. Elle est classée en deux catégories :

\paragraph{Portée}
Quand il s’agit de deux classes qui semblent représenter le même concept mais qui n'ont pas exactement les mêmes instances (extensions) bien que l’ensemble de leurs instances se croise (se chevauche), \textit{e.g.} les concepts "Student" et "TaxPayer".

\paragraph{Couverture et granularité du modèle}
C'est une différence dans la partie du domaine couverte par les deux ontologies (\textit{e.g.}, les ontologies des employés universitaires et des étudiants), ou une différence dans le niveau de détail avec lequel le modèle est modélisé / couvert (\textit{e.g.}, une ontologie peut avoir un concept "Person" alors qu'une autre peut distinguer "YoungPerson", "MiddleAgedPerson" et "OldPerson").\medskip\vspace{4px}

Tenons l'exemple d'une ontologie sur les voitures :

\begin{itemize}
\item Une ontologie peut modéliser des voitures mais pas des camions.
\item Une autre pourrait représenter les camions mais les classer seulement dans quelques catégories.
\item Alors qu'une troisième pourrait faire des distractions très fines entre les types de camions en se basant sur leur structure physique générale, poids, but, \textit{etc.}
\end{itemize}

\subsubsection{\underline{Différence d’explications}}
C'est une différence dans la façon dont la conceptualisation est spécifiée. Elle se base sur la manière d'exprimer les entités. Elle est classée en trois catégories :

\paragraph{Style de modélisation}
Une différence dans le style de modélisation qui résulte des choix explicites du modélisateur :

\subparagraph{\textit{Paradigme}}
De différents paradigmes peuvent être utilisés pour représenter certains concepts tels que le concept du temps (\textit{e.g.}, représentation basée sur les "intervalles" \textit{vs} représentation basée sur les "points"), l'action, les plans, la causalité, les attitudes propositionnelles, \textit{etc.}

\subparagraph{\textit{Description des concepts ou convention de modélisation}}
Les différences dans la description des concepts ou les conventions de modélisation peuvent se manifester par l’utilisation de différentes structures pour représenter des informations identiques ou similaires, \textit{e.g.} une distinction entre deux classes peut être modélisée en utilisant un attribut qualificatif (une propriété), ou en introduisant une autre (sous-) classe. Un autre choix dans les descriptions des concepts est la manière dont la hiérarchie is-a "<" est construite, en effet, les entités peuvent être augmentées ou réduites dans la hiérarchie, \textit{e.g.} la classe "thèse" ou "dissertation" peut être modélisée comme dissertation < livre < publication scientifique < publication, ou comme dissertation < livre scientifique < livre < publication, ou même comme sous-classe de "livre" et de "publication scientifique".

\paragraph{Terminologie}
Une différence terminologique peut 	voir deux cas :

\subparagraph{\textit{Termes synonymes}}
Lorsque deux concepts sont équivalents, mais représentés en utilisant des noms différents. Un exemple trivial est l'utilisation du terme "Car" dans une ontologie et du terme "Automobile" dans une autre. Un type particulier de ce problème est le cas où le langage naturel avec lequel les ontologies sont décrites diffère.

\subparagraph{\textit{Termes homonymes}}
Lorsque le même nom est utilisé pour des concepts différents, \textit{e.g.} dans le domaine musical, le terme "Conductor" (le chef d’orchestre) a une signification différente que celle dans le domaine de l’ingénierie électrique (le conducteur électrique).

\paragraph{Différence de codage}
Une différence de codage se produit lorsque les valeurs dans les différentes ontologies sont codées dans différents formats, \textit{e.g.} une date peut être représentée par "jj/mm/aaaa" ou "mm-jj-aa", la distance peut être décrite en "mile" ou en "kilomètre", le poids peut être décrit en "gramme" ou en "pound", le prix peut être décrit en différentes monnaies, \textit{etc.}

\section{Ingénierie ontologique}

L'ingénierie des ontologies est un contexte dans lequel les utilisateurs sont confrontés à des ontologies hétérogènes. Et plus généralement, c’est la tâche de concevoir, mettre en œuvre, et maintenir des applications basées sur les ontologies \cite{euzenat2007ontology}. Elle doit traiter plusieurs ontologies distribuées et évolutives.\medskip\vspace{4px}

Dans le but d'atténuer l'hétérogénéité croissante et la complexité des ontologies modernes, plusieurs domaines de recherche connexes ont vu le jour au cours des dernières années, tels que le "matching", le "mapping", l’"alignement", l’"intégration", la "fusion", le "versionning", et l'"évolution" des ontologies qui sont les domaines les plus répandus.\cite{caldarola2016} Nous n'allons expliciter que les domaines liés à notre thème.

\subsection{Médiation}
La médiation des ontologies est un vaste domaine de recherche qui vise à déterminer et réconcilier les différences entre les ontologies afin de permettre leur réutilisation dans différentes applications hétérogènes dans le Web sémantique. \cite{de2006ontology} distinguent deux types principaux de médiation ontologique : le mapping et la fusion. \cite{leung2014new} distinguent trois types de médiation : le matching, la fusion, et l'intégration (qui sont basées sur le matching).

\subsection{Réconciliation}

La réconciliation des ontologies est un processus qui harmonise le contenu de deux ou de plusieurs ontologies. Il exige typiquement de faire un matching entre deux ontologies, et des changements dans un des deux côtés, ou dans les deux côtés \cite{euzenat2007ontology}. Dans ce cas, il ne s'agit pas d'une fusion ou d’une intégration d'ontologies, mais plutôt d'une \textbf{coévolution}. Sachant que la réconciliation des ontologies peut être effectuée pour le but de fusionner ou d’intégrer deux ontologies.

\subsection{Matching (Appariement)}
Le matching des ontologies peut être une solution au problème de l’hétérogénéité sémantique des systèmes car il permet que la connaissance et les données exprimées dans les ontologies correspondues soient interopérables. C’est le processus de découverte des relations sémantiques ou des correspondances entre des entités provenant de deux différentes ontologies (ou de plusieurs ontologies dans le cas du matching multiple). Ces entités sont généralement des entités nommées (des classes, des propriétés, ou des individus), mais elles peuvent aussi être des entités anonymes \textit{i.e.} des expressions plus complexes (des formules).\medskip\vspace{4px}

Le matching des ontologies peut concerner des ontologies entières (\textit{i.e.} tout type d’entités : T-box et A-box), ou bien uniquement la partie "schéma" (la structure) des ontologies (\textit{i.e.} T-box : seulement les classes et les propriétés) \cite{cheatham2017semantic}.\medskip\vspace{4px}

La correspondance est la relation sémantique détenue ou supposée être détenue entre deux entités des différentes ontologies. La relation entre les deux entités n’est pas limitée à la relation d’équivalence, elle peut être plus sophistiquée, \textit{e.g.} la subsomption, la disjonction, l’instanciation, et même des relations floues. Certains auteurs, utilisent le terme mapping, au lieu de correspondance.\medskip\vspace{4px}

Le résultat du matching, l'"alignement" (éventuellement le "mapping"), exprime, avec de différents degrés de précision, les relations sémantiques entre les ontologies mises en correspondance. Plusieurs auteurs utilisent le terme "alignement" (qui est le résultat du matching), au lieu de "matching", et utilisent le terme "mapping" au lieu d’"alignement" (que nous expliquerons juste après) \cite{euzenat2007ontology}.\medskip\vspace{4px}

Formellement, le processus de matching peut être vu comme une fonction $f$ qui, à partir d’une paire d’ontologies $O$ et $O'$ à mettre en correspondance, un ensemble de paramètres $p$, et un ensemble de ressources externes $r$, retourne en sortie un alignement $A$ (éventuellement un mapping) entre ces deux ontologies : $A = f (O, O', p, r)$.

\begin{figure}[!ht]
\centering
\includegraphics[scale=0.6]{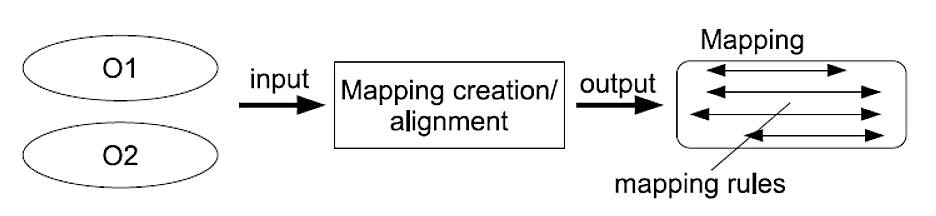}
\caption[Matching]{Matching \cite{de2006ontology}}
\end{figure}

Le type le plus simple de relations à trouver est l’équivalence ou la disjonction (l’exclusion) un à un (1-à-1) entre deux entités appartenant chacune à une ontologie. Le niveau de complexité suivant est la relation de subsomption (d’inclusion) 1-à-1. Pour trouver des relations 1-à-1, une recherche exhaustive doit comparer chaque entité de la première ontologie avec chaque entité de la deuxième ontologie, ce qui peut être réalisable pour de petites ontologies, mais infaisable pour des ontologies contenant des millions d’entités. C’est pour cela que les systèmes de matching peuvent employer une étape de filtrage ou de hachage pour déterminer les entités qui valent la peine d'être comparées \cite{cheatham2017semantic}.\medskip\vspace{4px}

Les relations un-à-plusieurs (1-à-m) sont encore plus difficiles à trouver. Tenons comme exemple une relation d’équivalence entre une classe de la première ontologie et l’union de trois classes de la deuxième ontologie. Ce type de relation cause un problème de complexité. Pour trouver des relations 1-à-m, une approche exhaustive aurait besoin de comparer chaque entité de la première ontologie avec toutes les combinaisons possibles des m entités de la deuxième ontologie, ce qui n'est pas possible \cite{cheatham2017semantic}.\medskip\vspace{4px}

Trouver des relations plusieurs-à-plusieurs (n-à-m) arbitraires est la tâche d’alignement la plus complexe. Une relation arbitraire signifie tout type de relation, non seulement l’équivalence, la disjonction, et la subsomption \cite{cheatham2017semantic}.\medskip\vspace{4px}

\ding{220} Les systèmes de matching actuels traitent l’identification des relations 1-à-1. Ils sont devenus très compétents dans la découverte des relations d’équivalence 1-à-1 entre les classes et les instances, mais moins performants dans la découverte des relations entre les propriétés. Leur compétence et leur exactitude est due principalement aux mesures de similarité syntaxiques (de chaînes de caractères).\medskip\vspace{4px}

\ding{220} Les travaux qui traitent un matching multiple sont très spécifiques pour le moment, et seul un petit nombre d'algorithmes le considère.\medskip

Voici les travaux qui ont été menés au sein de notre laboratoire LIPAH concernant le matching des ontologies : \cite{zghal2007soda, zghal2007nouvelle, zghal2007edola, zghal2007new, zghal2011oacas, zghal2010contributions,  kachroudi2011ldoa, kachroudi2012exploring, kachroudi2013parametrage, kachroudi2013using, kachroudi2013ontopart, kachroudi2014bridging, kachroudi2015f, kachroudi2016initiating, kachroudi2017composition, kachroudi2017oaei, djeddi2015xmap, el2015clona}.

\subsubsection{\underline{Méthodes de matching}}

Pour évaluer la similarité des entités, les systèmes de matching utilisent différentes approches. Ils peuvent utiliser zéro ou plusieurs approches de mesure de similarité, soit en combinant leurs valeurs pour former une seule mesure, soit en les appliquant en série pour filtrer les correspondances et ne mesurer que les correspondances candidates \cite{cheatham2017semantic}.\medskip\vspace{4px}

La similarité reflète à quel point deux entités ont des choses en commun, c’est une mesure du degré qu’une entité puisse être utilisée à la place d’une autre. En général, la mesure ou le degré de confiance nous renseigne à quel point la correspondance est correcte et fiable. Plus elle est élevée, plus la relation qui la détient est solide. Généralement, c’est un nombre réel appartenant à un ensemble ordonné qui varie dans l’intervalle [0 1], mais il existe des systèmes qui utilisent simplement les booléens "vrai" et "faux" où le plus grand élément (1) est interprété en tant que "vrai", et le plus faible élément (0) est interprété en tant que "faux" \cite{euzenat2007ontology}. Un seuil peut être mis pour ne pas afficher les correspondances de mesure de similarité inférieure à ce seuil.\medskip\vspace{4px}

Parmi les méthodes utilisées par les approches de matching des ontologies, nous citons \cite{abels2005identification} :

\paragraph{Méthode basée sur les chaînes}
Elle compare deux entités en se basant sur les chaînes de caractères associées à elles. Les chaînes de caractères sont généralement les labels de l’entité, mais ils peuvent aussi inclure les commentaires et d’autres annotations de l’entité. Plus les chaînes sont similaires, plus elles sont susceptibles de désigner les mêmes concepts.\medskip\vspace{4px}

\ding{220} Cette approche souffre lorsque les concepts sémantiquement identiques sont modélisés avec des noms différents, \textit{i.e.} lorsqu’il s’agit de synonymes \cite{fahad2010disjoint}.

\paragraph{Méthode linguistique}
Telle que la suppression de mots inutiles (stop-words), la tokenisation, la stemmatisation du texte, la considération des préfixes ou des suffixes, \textit{etc.} pour gérer les noms des entités, \textit{e.g.} cette méthode détecte que les classes "house" et "houses" sont identiques.

\paragraph{Méthode sémantique}
Elle tente d’utiliser les sens des labels de l’entité, plutôt que leurs orthographes. Des ressources linguistiques externes comme les lexiques, les thésaurus, les dictionnaires, les encyclopédies, et les moteurs de recherche du Web sont souvent utilisées afin d'identifier les synonymes, les hyperonymes (is-a), ou les hyponymes (is-a). Il est courant d'utiliser la base de données lexicale WordNet, l’ontologie de référence (UMLS), ou les règles d'articulation (les mappings), pour identifier les relations entre les entités \cite{cheatham2017semantic}.\medskip\vspace{4px}

\ding{220} L’inconvénient de cette méthode c’est qu’elle est spécifique au domaine particulier de la ressource externe utilisée, et ne produit des résultats efficaces que lorsqu'elle est utilisée pour des ontologies dans ce même domaine. Elle manquerait de précision si elle aurait été appliquée à des ontologies de domaine plus général ou totalement différent \cite{fahad2010disjoint}.

\paragraph{Méthode taxonomique / structurelle}
Elle ne considère que la relation de spécialisation (héritage). Son intuition est que la spécialisation (is-a) relie des termes qui sont déjà similaires (étant interprétés comme un sous-ensemble ou un sur-ensemble de l'autre), par conséquent, leurs voisins peuvent aussi être en quelque sorte similaires. Elle examine le voisinage de deux entités pour déterminer leur similarité \cite{euzenat2007ontology}.

\paragraph{Méthode basée sur les attributs / propriétés}
Elle examine les attributs de deux concepts pour déterminer leur similarité \cite{fahad2010disjoint}.\medskip\vspace{4px}

\ding{220} Son inconvénient est qu’elle produit des correspondances inexactes lorsque de différents concepts ont les mêmes attributs, \textit{e.g.} le concept "Person" et "Company" sont supposés être les mêmes sur la base des labels de leurs attributs qui sont identiques, tels que les attributs "name", "adress", et "phone", \textit{etc.}

\paragraph{Méthode extensionnelle}
Elle se base sur l’intuition qui dit : si deux classes ont les mêmes instances, alors ce sont des classes similaires.\medskip\vspace{4px}

\ding{220} L’inconvénient majeur de cette méthode se manifeste lorsque des concepts sémantiquement distincts ayant des instances en commun sont considérés comme identiques \cite{fahad2010disjoint}.

\paragraph{Méthode basée sur les graphes}
Cette méthode interprète la représentation graphique de la structure de deux ontologies et regarde les chemins, les enfants et les feuilles pour identifier leurs structures similaires en recherchant leurs parties identiques. Elle se base sur l’intuition qui dit : si deux nœuds de deux ontologies sont similaires, alors leurs voisins doivent aussi être plus ou moins similaires \cite{euzenat2007ontology}, \textit{e.g.} deux entités qui ont la même superclasse et qui partagent quelques instances en commun, sont considérées plus similaires que deux entités n’ayant pas ces choses en commun ; deux classes de deux ontologies sont similaires ou identiques si elles ont les mêmes attributs et les mêmes classes voisines.\\

Nous pouvons trouver une première classification qui groupe ces méthodes de matching en des approches \textbf{syntaxiques}, \textbf{structurelles}, et \textbf{sémantiques} ; et une autre classification qui les groupent en des approches \textbf{élémentaires} (qui calculent les correspondances en analysant les entités isolément, en ignorant leurs relations avec les autres entités), et des approches \textbf{structurelles} (qui calculent les correspondances en analysant l’apparition des entités ensemble dans une structure).\medskip\vspace{4px}

En pratique, il n’existe aucun système de matching automatisé qui peut générer des alignements complètement corrects. En effet, les alignements manqueront toujours quelques correspondances correctes, contiendront quelques correspondances incorrectes, ou bien les deux en même temps \cite{cheatham2017semantic}.

\subsubsection{\underline{Concernant notre sujet}}
Ces quinze dernières années, la grande majorité des recherches sur l'intégration des ontologies s’est concentrée surtout sur l'étape de matching des ontologies et a négligé la partie de fusion des ontologies qui vient après \cite{raunich2014target}. En effet, la résolution de l'hétérogénéité par les stratégies de matching des ontologies est considérée comme une phase interne nécessaire et très importante pour l'intégration (ou la fusion) des ontologies en une nouvelle ontologie les regroupant. L’amélioration du processus de matching va améliorer considérablement les résultats de l’intégration (et de la fusion) des ontologies \cite{umer2012semantically}.\medskip\vspace{4px}

Les systèmes de matching peuvent faire à la fin une vérification d’incohérence et une réparation à l’alignement (ou le mapping) produit, en supprimant les correspondances incorrectes ou incohérentes \textit{i.e.} celles qui sont correctes mais qui causent une incohérence logique dans l’ontologie produite suite à l’intégration ou la fusion des ontologies sources à l'aide de cet alignement-là (Nous détaillerons ce volet dans le chapitre 3).\\

\begin{figure}[ht]
\begin{center}
\includegraphics[width=\textwidth]{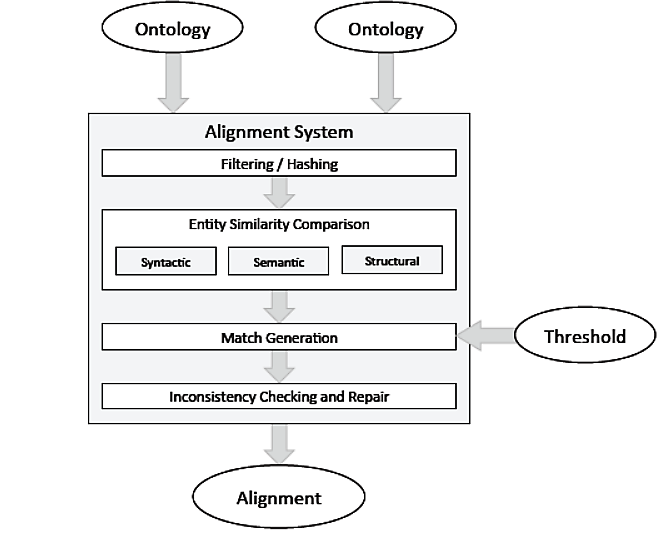}
\caption[Processus général du matching]{Processus général du matching \cite{cheatham2017semantic}}
\end{center}
\end{figure}

\subsection{Résolution de coréférence}
Les systèmes de matching des ontologies se concentrent généralement sur la recherche des relations entre les entités de schéma / T-box (les classes et les propriétés), alors que les systèmes de résolution de coréférence se concentrent sur l’identification des mêmes individus qui sont référencés par différents URIs \cite{cheatham2017semantic}.\medskip\vspace{4px}

Les relations cherchées par les algorithmes de résolution de coréférence sont uniquement des identités 1-à-1, car deux individus ne peuvent être qu’identiques ou distincts, tandis que les matchings (de schéma / T-box) impliquent (aussi) des classes et des propriétés, et ainsi peuvent avoir toute relation traditionnelle qui existe entre deux ensembles comme la subsomption, l’exclusion (la disjonction), \textit{etc.}\medskip\vspace{4px}

Le nombre d’instances (A-box) d’un data set (dans les linked data du Web) est souvent beaucoup plus grand que le nombre de ses entités de schéma (T-box), ainsi ce n’est pas faisable de comparer chaque individu d’un data set avec chaque individu d’un autre data set pour déterminer s’ils sont identiques ou pas. Par conséquent, une méthode de filtrage est utilisée pour décider si deux individus sont suffisamment proches pour valoir la peine d’être comparés ; s’ils le sont, un algorithme de comparaison va se produire en mesurant la similarité entre les individus, ou bien entre les individus et les noms des propriétés auxquelles elles sont liées. La mesure de similarité la plus utilisée est la similarité syntaxique (de chaînes de caractères). Enfin, le système doit prendre le résultat de la comparaison de deux individus et décider s’ils sont identiques ou pas en spécifiant souvent un seuil (une valeur empirique malheureusement) \cite{cheatham2017semantic}.\medskip\vspace{4px}

Le matching (des schémas) a un plus grand historique de recherche que celui de la résolution de coréférence qui vise l’intégration des linked data \cite{cheatham2017semantic}.

\subsection{Alignement}
\subsubsection{Définition 1}
Les auteurs \cite{noy1999algorithm} pensent que dans l’alignement, les ontologies sources (généralement de domaines complémentaires) doivent être toujours séparées et consistantes les unes avec les autres, tout en ayant des liens entre elles. Les auteurs \cite{zhu2009research} le pensent aussi et définissent l’alignement par le processus qui combine deux ontologies et qui établit ensuite une collection de relations binaires (correspondances) entre elles.


\begin{figure}[!ht]
\centering
\subfloat[\cite{noy1999algorithm}]{
\includegraphics[scale=0.45]{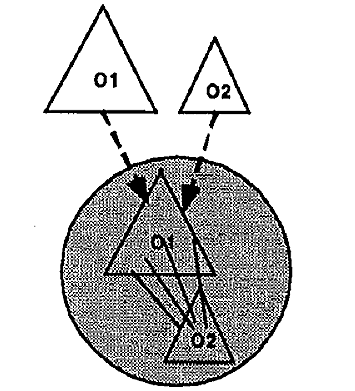}}
\quad
\subfloat[\cite{zhu2009research}]{
\includegraphics[scale=0.525]{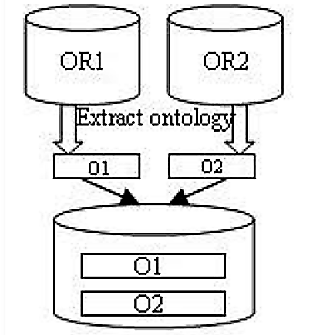}}
\caption[Alignement 1]{Alignement}
\end{figure}

\subsubsection{Définition 2 (consensuelle)}
Etant donné deux ou plusieurs ontologies (dans le cas d’un matching multiple), l’alignement est un ensemble de correspondances (relations) sémantiques entre des paires d’entités appartenant à différentes ontologies. Rappelons que l’alignement est la sortie du processus de matching des ontologies \cite{euzenat2007ontology}.\medskip\vspace{4px}

Plusieurs auteurs, utilisent le terme "mapping" au lieu d’"alignement". Dans le reste de ce mémoire, nous utiliserons le mot "alignement" dans ce sens.\medskip\vspace{4px}

Puisque la relation est une relation binaire valable dans les deux sens et pouvant être décomposée en une paire de fonctions totales, \cite{kalfoglou2003ontology} supposent que l'alignement des ontologies peut être décrit au moyen d'une paire de mappings (chacun contenant des correspondances dans un seul sens). Ils introduisent la notion de l’ontologie intermédiaire commune $O_0$ (ou l’articulation) qui peut être créée à travers cet alignement.\medskip\vspace{4px}

Formellement, étant donné deux ontologies $O$ et $O'$ (ayant les langages $L$ et $L'$) et un ensemble de relations d’un alignement A, une correspondance est un triplé $(e, e', r)$, tel que $e \in O$, $e' \in O'$, et $r \in \Theta$. La correspondance $(e, e', r)$ déclare que la relation bidirectionnelle $r$ relie les entités $e$ et $e'$ ; mais elle est souvent accompagnée aussi par un identifiant et une confiance, ainsi, elle sera représentée généralement par un tuple $(id, e, e', r, n)$ où $id$ est son identifiant unique, et $n$ est sa mesure de confiance \cite{euzenat2007ontology}. L'alignement peut avoir des correspondances ayant la même entité source, \textit{i.e.} une entité source peut avoir plus qu'une relation avec des entités cibles.\medskip\vspace{4px}

Les alignements peuvent être utilisés dans des tâches variées, telles que la réponse aux requêtes, la liaison des données, la navigation dans le Web sémantique, la transformation des ontologies, l’intégration et la fusion des ontologies, et le raisonnement sur les ontologies.

\begin{figure}[!ht]
\begin{center}
\includegraphics[scale=1.03]{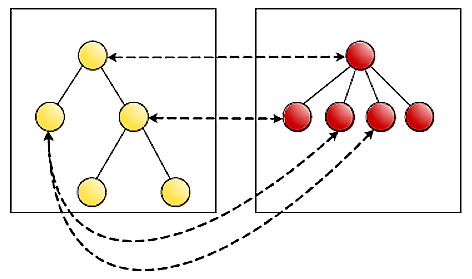}
\caption[Alignement 2]{Alignement \cite{abels2005identification}}
\end{center}
\end{figure}

\subsubsection{\underline{Concernant notre sujet}}
Il est possible d’utiliser des relations à partir d’un langage ontologique pour exprimer un alignement. Tenons l’exemple du langage OWL qui peut être considéré comme un langage d’expression de correspondances entre les ontologies. En effet, dans OWL, les primitifs "\textit{equivalentClass}", "\textit{equivalentProperty}" et "\textit{sameAs}" ont été introduits initialement pour lier les éléments des ontologies de même domaine ; d’ailleurs, dès qu’une ontologie OWL implique des entités provenant d’autres ontologies, elle exprime implicitement des alignements. Par conséquent, il est possible d’utiliser ces constructeurs pour relier les entités de deux ontologies mises en correspondance ou pour créer une ontologie OWL intermédiaire.

\subsection{Mapping}
\subsubsection{Définition 1}
Selon \cite{umer2012semantically}, le "mapping" des ontologies est une approche pour l’intégration des ontologies où l’ontologie intégrée $O$ contient les règles de correspondance entre les entités des ontologies A et B. \cite{klein2001combining} considère également le mapping comme une intégration virtuelle. (C’est la même notion d’ontologie intermédiaire $O_0$ ou d’articulation rencontrée dans la partie "alignement").\medskip\vspace{4px}

Selon \cite{ziemba2015integration}, le mapping permet d'obtenir un résultat similaire à l'ontologie de pont (c'est l'ontologie que nous allons réaliser dans ce mémoire). Cependant, dans l'ontologie de pont, par opposition au mapping, les ontologies sources et les connexions entre elles sont stockées ensemble, or que dans le mapping, les connexions sont à part. Le mapping entre les ontologies forme des "ponts sémantiques" \cite{de2006ontology}.

\subsubsection{Définition 2 (consensuelle)}
Le mapping est la version orientée d’un alignement où une entité d’une première ontologie est correspondue à une entité d’une deuxième ontologie, et pas l’inverse. Il assigne chaque entité d’une ontologie à au plus une (exactement une ou aucune) entité de l’autre ontologie. Il se conforme à la définition mathématique d’une fonction totale (une relation unidirectionnelle (injective)), et non pas à la définition d’une relation générale bidirectionnelle (bijective). Selon \cite{euzenat2007ontology}, cette définition mathématique exige que l’entité mise en correspondance soit égale à son image, \textit{i.e.} que la relation soit une relation sémantique d’équivalence ou d’identité.\medskip\vspace{4px}

Selon \cite{flouris2006classification}, un mapping peut être perçu comme une collection de règles (ou d’axiomes) toutes orientées dans la même direction, de telle sorte que les entités de l’ontologie source et cible apparaissent au maximum une fois. Ils ajoutent que les deux ontologies mappées doivent partager le même domaine de discours (ou des domaines proches). Ceci est implicite sinon nous n’aurons pas de mapping. Dans le reste de ce mémoire, nous utiliserons le terme "mapping" dans ce sens.\medskip\vspace{4px}

D’après \cite{de2006ontology}, le mapping, comme l’alignement, est stocké séparément des deux ontologies, ainsi il n'est pas incorporé dans les définitions de ces ontologies.

\begin{figure}[ht]
\begin{center}
\includegraphics[scale=0.8]{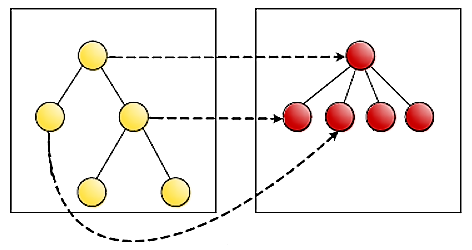}
\caption{Mapping}
\end{center}
\end{figure}

\subsubsection{\underline{Utilisations du mapping}}
D’après \cite{choi2006survey}, le mapping des ontologies est utilisé principalement dans trois situations :

\paragraph{Processus d’intégration des ontologies}
Il s'agit du mapping entre une ontologie globale et des ontologies locales dans le processus de l'intégration des ontologies : Il décrit les relations (les correspondances) entre l’ontologie globale (cible) et les ontologies locales (sources) la composant. Il peut être aussi utilisé pour exprimer une entité de l’ontologie globale dans une vue ou une requête sur les autres ontologies (approche global-centric), ou l’inverse (approche local-centric).

\paragraph{Processus de fusion ou d'alignement}
Il s'agit du mapping entre des ontologies sources dans le processus de fusion ou d'alignement (définition 1) des ontologies. Il identifie les similitudes (synonymies) entre les différentes ontologies pour pouvoir les fusionner ou les aligner.

\paragraph{Processus de transformation des ontologies}
Il s'agit du mapping entre deux ou plusieurs ontologies sources dans le processus de transformation des ontologies : Il peut être utilisé pour transformer les entités sources en des entités cibles en se basant sur leurs correspondances, \textit{i.e.} leurs relations d’équivalence sémantique dans le mapping. Il fournit une interopérabilité entre les différentes ontologies qui ne peuvent pas être intégrées ou fusionnées à cause d’une inconsistance mutuelle de leurs informations.\\

\ding{220} Les utilisations du mapping dans une ontologie intermédiaire, une ontologie de pont, une transformation des ontologies, une interconnexion des données, ou une requête, s’avèrent très utiles pour les environnements dynamiques, ouverts et distribués, et évitent également la complexité et les coûts de l'intégration ou de la fusion des ontologies sources. En effet, le mapping forme une sorte de couche ou d’interface commune entre les ontologies.

\subsection{Mophisme}
Selon \cite{flouris2006classification}, le morphisme des ontologie est une collection de correspondances sous forme de fonctions qui relient non seulement les signatures (les vocabulaires, les entités) de deux ontologies, mais aussi leurs axiomes (les syntaxes, les formalismes, les constructeurs des langages).\medskip\vspace{4px}

Selon \cite{euzenat2007ontology}, le terme "morphisme" est utilisé pour représenter un mapping entre différents types de modèles\textbf{*}. Il contient des relations binaires sur deux ensembles d'identificateurs d'objets (OIDs) et il peut être inversé et composé.\medskip\vspace{4px}

\textbf{*} Les modèles, tels que les schémas relationnels ou les schémas XML, sont représentés implicitement (intérieurement) sous la forme de graphes étiquetés et orientés, dans lesquels les nœuds désignent les éléments du modèle (les relations et les attributs). Chacun de ces éléments est identifié par un identifiant d'objet (OID) \cite{euzenat2007ontology}.

\subsection{Transformation}
La transformation des ontologies est le processus de changement / de traduction des entités (vocabulaire, signature) d’une ontologie par les entités d’une autre ontologie. Elle est utile quand nous voulons exprimer une ontologie par rapport à une autre. En général, les deux ontologies initiales sont inchangées et une troisième ontologie (le résultat de la transformation de la première ontologie par rapport à la deuxième) est créée. Les conséquences de la première ontologie sont aussi les conséquences du résultat de la transformation \cite{euzenat2007ontology}.\medskip\vspace{4px}

Ce terme est très confondu à la notion de traduction (que nous allons expliquer juste après). Formellement, la transformation des ontologies à l’aide d’un alignement (mapping) $A$ entre deux ontologies $O$ et $O'$, génère une ontologie $O''$ qui transforme les entités de $O$ par celles de $O'$ suivant les correspondances dans $A$. Elle peut être exprimée par l’opérateur suivant : $Transform(O, A) = O''$. Les opérations de transformation sont orientées, \textit{i.e.} la transformation a une source et une cible identifiées, ainsi, à partir d’un alignement, il est possible de générer deux opérations (dans les deux sens) selon la source et la cible.\medskip\vspace{4px}

La transformation des ontologies n’est pas bien supportée par les outils. Elle peut être particulièrement utile dans la connexion d’une ontologie à une autre ontologie (réconciliation des ontologies), ou dans la connexion d’une ontologie locale à une ontologie globale dans le cadre de l’\textbf{intégration} ou la \textbf{fusion des ontologies}. Elle est utilisée aussi pour importer des données sous une autre ontologie sans importer l’ontologie elle-même.

\subsection{Traduction}
La traduction des ontologies est le processus qui transforme la représentation formelle de l'ontologie d'un langage (d’un formalisme de représentation) à un autre, tout en préservant la sémantique, \textit{e.g.} de "Ontolingua" à "Prolog". Elle change la forme syntaxique des axiomes, mais pas le vocabulaire (pas la signature) de l’ontologie \cite{klein2001combining, kalfoglou2003ontology, flouris2006classification, euzenat2007ontology}.

\subsection{Interconnexion des données}
L'interconnexion des données est le processus qui consiste à établir des liens explicites, principalement des déclarations de la relation d'identité "owl:sameAs" entre les instances de deux ensembles de données RDF différents dans le Web de données (Linked Data). Il est possible de traiter les alignements en tant que spécifications de liaison, ainsi l’interconnexion des données pourrait être exprimée par l'opérateur $Interlink (d, d', A) = L$ dans lequel un alignement $A$ résultant de chaque couple d’ontologies ($O$ et $O'$) sous lesquels deux ensembles de données ($d$ et $d'$) sont exprimés, est utilisé pour les lier, et générer un ensemble de liens $L$ entre les ressources (les URIs des instances) de ces deux data sets \cite{euzenat2007ontology}.\shorthandoff{:} Bien qu’il y a une quantité très énorme de liens de type "\textit{owl:sameAs}" entre les instances des data sets du LOD, il n’existe que quelques liens rares de type "\textit{owl:equivalentClass}" ou "\textit{owl:equivalentPropery}" entre leurs classes et leurs propriétés \cite{zhao2014ontology}.\medskip\vspace{4px}
\shorthandon{:}

Dans le Web des données, le "matching" et la "résolution de coréférence" sont utiles dans l'aide à la génération de ces liens qui fournissent le contexte nécessaire pour rendre les données plus utiles \cite{cheatham2017semantic}. Ils sont effectués hors ligne et sans contraintes de temps de telle sorte que les correspondances résultantes soient correctes, mais pouvant être non exhaustives (non complètes).\medskip

Dans ce contexte, citons le travail de \cite{hamdi2015linking}, membre de notre laboratoire LIPAH, qui a exploité l'ontologie du Web de données FOAF pour les réseaux sociaux.

\subsection{Intégration / Fusion}
Nous allons l’expliquer en détail dans le chapitre 2

\subsection{Raisonnement}
Le raisonnement consiste en l'utilisation des alignements comme des règles pour raisonner sur les ontologies mises en correspondance. Les "\textit{bridging}" axiomes utilisés dans l'intégration (l'ontologie de pont) sont des règles. Cet ensemble de règles est vu comme une ontologie $O$ qui doit être écrite dans un langage ontologique supportant les règles ou les expressions des axiomes de pont. C’est la même notion de l’ontologie intermédiaire ou l’articulation des ontologies (définition 1 d'un mapping). Le raisonnement peut être décrit par la fonction $TransformAsRules(A) = O$ où $A$ est un alignement entre deux ontologies $O'$ et $O''$ \cite{euzenat2007ontology}.\medskip\vspace{4px}

Toute transformation des alignements sous une forme adaptée au raisonnement, telle que SWRL ou OWL, peut être utilisée par les moteurs d'inférence (les raisonneurs) de ces langages, tels que Pellet ou HermiT.

\subsection{Enrichissement}
L'enrichissement est le processus qui cherche de nouvelles entités (généralement à partir de ressources textuelles externes) et les place correctement au sein de l'ontologie à enrichir.\medskip

Voici quelques travaux d'enrichissement d'ontologies réalisés au sein de notre laboratoire LIPAH : \cite{kamoun2010evolution, hamdi2012enriching, kamoun2012novel, kamoun2012automatic, kamoun2012information, kamoun2014stability}.

\section*{Conclusion}
A présent, les entreprises ont migré vers l'adoption des stratégies de mondialisation et d'internationalisation. En effet, traditionnellement, les entreprises partageaient seulement les biens physiques en collaboration, mais maintenant elles ont aussi besoin de partager et intégrer leurs connaissances. C'est pourquoi la notion de l'interopérabilité s'impose car elle permet aux systèmes informatiques hétérogènes de communiquer, interpréter et traiter l'information échangée.\medskip\vspace{4px}

Pour ce faire, les ontologies se présentent comme le meilleur outil pour communiquer et partager des connaissances en fournissant une compréhension commune d'un domaine donnée. Malheureusement, les concepteurs des ontologies eux-mêmes appliquent des visions différentes du même domaine lors du développement des ontologies, et ceci engendre le problème de l'\textbf{hétérogénéité sémantique} qui est l'un des principaux obstacles de l'interopérabilité sémantique ; Il se produit lors de l'utilisation des ontologies de même domaine, \textit{i.e.} quand des ontologies hétérogènes réutilisent les mêmes connaissances. L'intégration sémantique est indispensable pour remédier à ce problème. Elle se base sur la sémantique des systèmes inter-opérants pour comparer leurs différents concepts et déduire leurs correspondances, et éventuellement les associer et créer des bases de connaissances intégrées. Par conséquent, l'intégration sémantique mènera inévitablement à un matching inter ontologique qui est une étape essentielle dans l'intégration des ontologies.
\chapter{Revue sur la fusion et l’intégration des ontologies}

\section*{Introduction}
A présent, il existe une très grande confusion dans l’utilisation des termes "intégration" et "fusion" dans la littérature. En effet, il arrive que nous trouvons des travaux sur la fusion que les auteurs nomment "intégration", et des travaux sur l’intégration que les auteurs nomment "fusion" ; il y a des cas où les auteurs utilisent les deux termes comme synonymes et choisissent l’un d’entre eux comme titre de l’article (ils choisissent généralement le terme "intégration" car il parait plus général, ainsi vrai dans les deux cas). Par ailleurs, plusieurs auteurs utilisent le terme "intégration" dans le titre de leurs travaux, sans pour autant faire une intégration ; ils se contentent d’un matching ; et même s’ils font une intégration, ils ne l’explicitent et ne l’évaluent pas, ils se concentrent seulement sur l’évaluation du matching.\medskip\vspace{4px}

Toutes les définitions et les approches qui vont suivre vont être ordonnées chronologiquement dans chaque section.\medskip\vspace{4px}

Dans ce chapitre, nous citons les différentes définitions et approches du terme \textit{fusion des ontologies} dans la littérature. Puis, nous citons les différents types d’intégration des ontologies, les définitions du terme \textit{intégration des ontologies} dans la littérature, et ses principales approches existantes. Par la suite, nous évoquons les avantages de l’intégration et la fusion, leurs différences et leurs points communs, et nous expliquons les causes des erreurs qui peuvent se produire suite à ces deux processus. Ensuite, nous consacrons une petite section pour parler de ce qui nous intéresse parmi toutes ces définitions et approches évoquées. Enfin, nous clôturons ce chapitre par une conclusion qui résume le tout.

\section{Fusion des ontologies}

Puisque de nombreuses ontologies se réfèrent au même domaine et aux mêmes objets, il existe un besoin croissant de les fusionner et les organiser. En effet, le but ultime de la fusion est de représenter une meilleure perspective des connaissances d’un domaine. \medskip\vspace{4px}

En général, la fusion des ontologies est utilisée dans le domaine de l’intégration des données, mais elle peut être aussi perçue comme une technique utilisée dans le domaine de l'enrichissement des ontologies (de domaine) qui consiste à insérer dans l'ontologie des connaissances connexes en moins de temps et de coût.\medskip\vspace{4px}

La façon dont le processus de fusion est effectué est encore très peu claire. En effet, il n'y a pas de consensus sur la méthodologie à suivre pour fusionner les ontologies. La seule phase commune est la phase initiale qui prend en entrée un ensemble d’ontologies (deux ou plus). Certains commencent directement par toutes les ontologies à fusionner (méthode non incrémentale), d’autres commencent par un groupe initial sélectionné d'ontologies (généralement par une seule ontologie) qui est élargi ensuite de manière incrémentielle par les autres ontologies (méthode incrémentale) \cite{pinto2004ontologies}.

\subsection{Définitions de la fusion}
\subsubsection{\underline{Fusion de Noy}}
D’après \cite{noy1999algorithm}, dans la fusion, une ontologie unique qui est une version fusionnée des ontologies d'entrée est créée. Souvent, les ontologies sources couvrent des domaines similaires ou liés. C’est une définition très vague (et qui peut convenir aussi au terme "intégration des ontologies").

\subsubsection{\underline{Fusion de Pinto (consensuelle)}}
Selon \cite{pinto1999towards}, dans la fusion, nous avons, d'une part, un ensemble d’ontologies (au moins deux) qui vont être fusionnées ($O_1$, $O_2$, …, $O_N$) et, d'autre part, l'ontologie résultante du processus de fusion ($O$). Ainsi, cette méthode est non incrémentale. Le sujet des ontologies sources et de l’ontologie résultante est le même ($S$), bien que certaines ontologies sources soient plus générales que d'autres (leur niveau de généralité peut ne pas être le même). Le but est de remplacer les ontologies existantes, portant sur un sujet particulier, par une ontologie plus riche et plus large qui couvre mieux ce même sujet en fusionnant leurs connaissances (les terminologies, les définitions, et les axiomes des ontologies sources).\medskip\vspace{4px}

Selon eux, dans la fusion (l’unification qui est le troisième cas d’intégration de \cite{sowa1997electronic}), les ontologies sources sont unifiées en une seule. Dans certains cas, les connaissances des ontologies sources sont homogénéisées et modifiées par l'influence d'une ontologie source sur une autre (à l'aide des opérations d’abstraction, de généralisation, de transformation (mapping)). Dans d'autres cas, les connaissances provenant d'une ontologie source particulière sont dispersées et mêlées avec les connaissances des autres sources.\medskip\vspace{4px}

\cite{malik2010ontology} donnent une autre définition proche qui considère que la fusion est le fait de former des ontologies mieux modélisées à partir d'ontologies mal définies ou plus petites (\textit{i.e.} qui ne couvrent pas tout le domaine).

\begin{figure}[p]
\begin{center}
\includegraphics[scale=0.7]{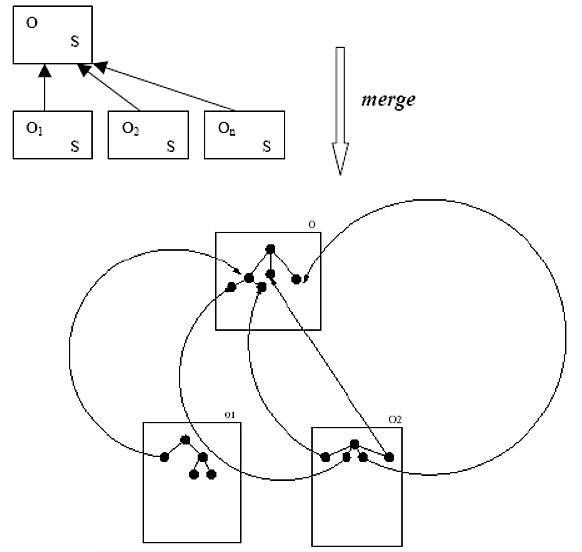}
\caption[Fusion 1]{Fusion de \cite{pinto1999towards} interprétée par \cite{keet2004aspects}}\vspace{37px}\bigskip

\includegraphics[scale=0.7]{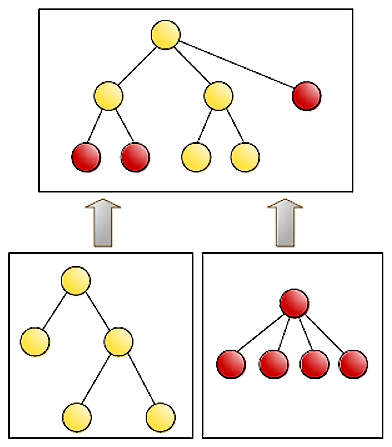} \caption[Fusion 2]{Fusion de \cite{abels2005identification}}
\end{center}
\end{figure}

\newpage
\subsubsection{\underline{Fusion comme étant une ontologie intermédiaire}}
Selon \cite{kalfoglou2003ontology}, la "forte" notion de fusion peut être détendue en prenant l'articulation (l’alignement des deux ontologies $O'$ et $O''$) avec laquelle une ontologie  $O$ pourrait être définie.

\subsubsection{\underline{Fusion de De Bruijn \textit{et al.}}}
D’après \cite{de2006ontology}, la fusion des ontologies est la création d’une nouvelle ontologie qui unie deux ou plusieurs ontologies en se basant sur les correspondances entre elles. Selon eux, la nouvelle ontologie doit capturer toutes les connaissances des ontologies sources et refléter toutes les correspondances entre elles pour pouvoir les remplacer. Nous notons qu'ils n'évoquent pas les domaines des ontologies sources (différents ou similaires). Ils distinguent deux approches distinctes dans la fusion des ontologies :\\

\begin{itemize}
\item[$\bullet$] \textbf{Fusion complète} (Full Merge) : Chaque paire d'entités équivalentes est fusionnée en une seule entité.
\item[$\bullet$] \textbf{Ontologie de pont} : Nous allons l’expliquer tout de suite.
\end{itemize}

\begin{figure}[!h]
\begin{center}
\includegraphics[scale=0.72]{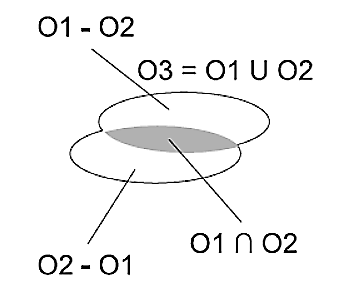}
\caption[Fusion complète]{Fusion complète \cite{de2006ontology}}
\end{center}
\end{figure}

\subsubsection{\underline{Fusion dans la symétrie et l’asymétrie}}
Selon \cite{raunich2014target}, la fusion peut être symétrique ou asymétrique par rapport aux ontologies d'entrée. Ils exigent, dans ces deux types d'union, que la propriété de "\textit{préservation de l'égalité}" soit assurée, ce qui signifie que les entités correspondues (comme prescrit dans le mapping entre les deux ontologies d’entrée) doivent être fusionnées dans la même entité afin qu'elles ne soient représentées qu'une seule fois dans l’ontologie résultante. Selon eux, en fusionnant les entités équivalentes ainsi, ils réduisent le chevauchement sémantique (même si que l’héritage multiple est aussi une source de conflits), ainsi le résultat de la fusion sera plus compact et moins redondant qu'une simple union directe des ontologies d'entrée (\textit{i.e.} une ontologie de pont avec des "bridging" axiomes).

\paragraph{Approche symétrique}
L’approche symétrique est la plus courante et vise à fusionner les ontologies d’entrée avec la même priorité (en préservant toutes les entités de toutes les ontologies). C’est une approche "Full Merge" qui prend l'union des ontologies d'entrée et qui combine leurs entités équivalentes (en une seule entité). Mais elle engendre une quantité importante de conflits sémantiques due à l’organisation hétérogène des mêmes concepts dans les ontologies d’entrée et à l’introduction de l’héritage multiple dans les entités fusionnées, ce qui génère des chemins redondants (plusieurs chemins conduisant à une même entité), réduisant ainsi la compréhensibilité de l’ontologie résultante.

\paragraph{Approche asymétrique}
L’approche asymétrique, prend l'une des ontologies d'entrée comme cible, dans laquelle les autres ontologies sources seront fusionnées (d’une façon incrémentale) pour l’étendre, donnant la préférence uniquement à l'ontologie cible dont toutes les entités à elle seule doivent être préservées. Les entités des ontologies sources ne doivent pas obligatoirement faire partie de l’ontologie résultante (cible). Ici, nous n’aurons plus affaire à l’héritage multiple, ainsi nous n’aurons pas (ou presque pas) de conflits sémantiques dans l’ontologie résultante qui aura bien une structure d’arbre (où un seul chemin conduit à une entité).\medskip\vspace{4px}

\ding{220} Mais d’après nous, cette approche asymétrique est un enrichissement de l’ontologie cible, plutôt qu'une fusion des ontologies sources.

\subsubsection{\underline{Fusion comme étant synonyme à l'intégration}}
\cite{raunich2012towards} et \cite{zhang2017oim} utilisent les termes "fusion" et "intégration" comme des synonymes.

\subsubsection{\underline{Fusion comme étant une intégration}}
\cite{chatterjee2017ontology} considèrent la création d'une ontologie à l’aide de la fusion comme étant un processus incrémental où des ontologies de petites tailles, de différents domaines, et de développement indépendant, devraient être fusionnées en une seule ontologie pour former un domaine (interdisciplinaire) plus vaste. (C’est la définition de la composition / l’intégration).\medskip\vspace{4px}

Ils donnent l’exemple du domaine de l’agriculture qui peut se composer de plusieurs sous domaines tels que les pesticides et les engrais, la récolte, la terre (le sol), les prévisions météo, l’infrastructure d’irrigation, la gestion de la sécheresse, la gestion du bétail, l’infrastructure de marketing, le suivi des régimes et des programmes, \textit{etc.}

\subsubsection{\underline{Fusion comme étant une ontologie de pont}}

Selon \cite{de2006ontology}, dans la seconde approche de fusion (l’ontologie de pont), les ontologies originales ne sont pas remplacées, elles sont conservées après l’opération de fusion, c’est plutôt une "vue", appelée "Bridge Ontology", qui est créée. Elle importe les ontologies originales et spécifie des correspondances entre elles pour relier les entités de ces ontologies par des axiomes de pont. Ces "Bridging" axiomes sont des règles de transformation utilisées pour connecter la partie de chevauchement des ontologies sources.\medskip\vspace{4px}

D’après \cite{euzenat2007ontology}, la fusion des ontologies est la création d'une nouvelle ontologie $O''$ qui lie les différentes entités de deux ontologies $O$ et $O'$ (qui se chevauchent) par des axiomes de pont ou des axiomes d'articulation, comme prescrit dans l'alignement entre $O$ et $O'$. Ils expriment la fusion par l'opérateur suivant : $fusion(O, O', A) = O''$. Selon eux, les ontologies sources sont inchangées et l'ontologie résultante est supposée contenir les connaissances des ontologies initiales de sorte que les conséquences de chaque ontologie source soient les conséquences de la fusion.\medskip\vspace{4px}

Dans la fusion de \cite{abbas2013creating}, une nouvelle ontologie peut être créée à partir d'ontologies sources, en établissant des correspondances entre les ontologies sources (un matching), puis en les combinant avec ces correspondances trouvées. Ils ne spécifient pas également les domaines des ontologies sources.\medskip\vspace{4px}

\ding{220} \textbf{C’est l’approche que nous allons suivre.}

\begin{figure}[!h]
\begin{center}
\includegraphics[scale=0.47]{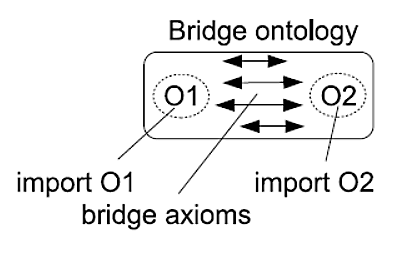}
\caption[Ontologie de pont]{Ontologie de pont \cite{de2006ontology}}
\end{center}
\end{figure}

\subsection{Principaux outils de fusion et leurs limites}

\subsubsection{\underline{Outils célèbres}}
Les approches les plus connues de fusion des ontologies telles que \textbf{PROMPT*} \cite{noy1999algorithm}, \textbf{Chimaera} \cite{mcguinness2000chimaera}, et \textbf{FCA-Merge} \cite{stumme2001fca} sont des Full Merge, semi-automatiques (nécessitant l'intervention d'experts et introduisant un effort manuel important, surtout pour les grandes ontologies) qui ne se basent pas sur les mappings, \textit{i.e.} elles n'appliquent pas une séparation entre le matching et la fusion \cite{raunich2012towards}.\medskip\vspace{4px}

En général, leurs algorithmes consistent en une itération de trois étapes principales :
\begin{enumerate}
\item Trouver un endroit où il y a un chevauchement dans les deux ontologies (trouver des entités candidates identiques ou apparentées);
\item Relier ces entités (qui sont sémantiquement proches) via des relations d'équivalence ou de subsomption ; ou les fusionner (après avoir transformé et uni les ontologies).
\item Vérifier la consistance, la cohérence et la non-redondance de la nouvelle structure de l’ontologie résultante, et les résoudre (trouver des solutions possibles à ces conflits).\\
\end{enumerate}

\textbf{*} Le plugin de PROMPT est dépassé et non fonctionnel maintenant (il n’est plus disponible pour le téléchargement).

\subsubsection{\underline{L’outil de Caldarola et Rinaldi}}
\noindent Le Framework de \cite{caldarola2016} contient quatre blocs principaux :
\begin{enumerate}
\item Le bloc de récupération des ontologies
\item Le bloc de normalisation des ontologies
\item Le bloc de matching des ontologies :\\Ce bloc est responsable de l’obtention d’un ensemble d’alignements ($A$) qui sont des correspondances entre les entités des ontologies d'entrée et de l’ontologie cible. Le matcher implique trois types d'opérations de matching (à base de chaînes, sémantique, et linguistique) qui vont également aider à faire une analyse automatique qui évalue et découvre les ontologies d’entrée les plus similaires à l'ontologie cible (\textit{i.e.} les ontologies pertinentes) parmi lesquels les ingénieurs de connaissances vont sélectionner les ontologies locales à réutiliser en les intégrant dans l’ontologie cible.
\item Le bloc de fusion des ontologies :\\Il est responsable de l'intégration des ontologies d'entrée sélectionnées dans une ontologie OWL globale qui sera plus riche. Selon les mesures des correspondances contenues dans l'ensemble des alignements ($A$), les opérations suivantes seront effectuées sur les entités appariées des ontologies :
\begin{itemize}
\item[$\diamond$] Si deux ou plusieurs entités (concepts ou relations) des ontologies sources sont équivalentes à une certaine entité cible, elles seront automatiquement fusionnées pour former une seule entité dans l'ontologie cible;
\item[$\diamond$] Si une entité source est subsumée par une entité cible, elle sera importée dans l'ontologie cible avec le consensus des experts du domaine ;
\item[$\diamond$] La même approche sera appliquée si une entité source subsume une autre entité cible ;
\item[$\diamond$] Si une entité source est disjointe à toutes les entités cibles, elle peut être non pertinente et ainsi rejetée, ou peut être considérée comme une nouvelle entité qui enrichit éventuellement l’ontologie cible.
\end{itemize}
\end{enumerate}

\noindent$\circleddash$ Mais ce processus nécessite beaucoup de travail manuel.

\subsubsection{\underline{L’outil de Chatterjee \textit{et al.}}}
Dans leur expérimentation, \cite{chatterjee2017ontology} ont choisi de créer une nouvelle ontologie dans le domaine de l’agriculture, en fusionnant des ontologies de sous-domaines (de la récolte, les engrais, la terre (le sol), et la météo).\medskip\vspace{4px}

\ding{220} Ce travail est en réalité une intégration (non pas une fusion), car les ontologies d’entrée appartiennent à différents domaines, et l’ontologie résultante est de domaine (interdisciplinaire) plus large qui englobe ces sous-domaines (c’est une composition de sous-domaines).\medskip\vspace{4px}

Ils font le parsing des fichiers .owl des ontologies d'entrée et extraient leur ensemble d’entités (en utilisant la bibliothèque "Owlready" en Python), puis ils font le matching de chaque couple d’ontologies (en combinant plusieurs techniques de matching (\textit{i.e.} de niveau élémentaire et structurel). A l’aide des alignements générés par le matching, ils appliquent une fusion complète des entités similaires et les mettent dans l’ontologie résultante $O_M$, puis ils copient les entités restantes (non fusionnées) des ontologies sources dans $O_M$, et génèrent un fichier .owl correspondant à $O_M$.\medskip\vspace{4px}

$\circleddash$ Bien qu'ils déclarent que leur outil assure que chaque aspect des ontologies sources soit présent dans l'ontologie de sortie, ils n’expliquent pas la manière avec laquelle ils ont fait les mises à jour de tous les axiomes sources qui appellent les entités nouvellement modifiées suite à la fusion (en effet, deux ou plusieurs entités similaires formeront une nouvelle entité).\\

$\circleddash$ Ils n’ont pas évoqué non plus le traitement des conflits sémantiques (les incohérences) générés certainement suite à la fusion.

\subsubsection{\underline{L’outil OIM-SM}}
Pour \cite{zhang2017oim}, le processus de fusion des ontologies avec la méthode OIM-SM prend deux ontologies et retourne une nouvelle ontologie (sous forme d’arbre, \textit{i.e.} sans héritage multiple). Il se compose des étapes suivantes :

\begin{enumerate}
\item Le matching d’équivalence sémantique entre les concepts des deux ontologies.
\item La fusion de toutes les paires de concepts équivalents, pour produire un nouveau concept à la place de chaque paire. Concernant les instances et les propriétés de chaque couple de concepts ($A$ et $B$), ils ont proposé d’appliquer deux règles de fusion pour former le nouveau concept $C$ :
\begin{itemize}
\item[$\ast$] L’ensemble des instances de $C$ est l’union de l’ensemble des instances de $A$ et de $B$.
\item[$\ast$] L’ensemble des propriétés de $C$ est l’intersection de l’ensemble des propriétés de $A$ et de $B$.\\
\ding{239} La sortie de ces deux étapes est un modèle en réseau où toutes les paires de concepts équivalents sont fusionnées générant ainsi un héritage multiple. Selon eux, il ne s’agit plus d’une ontologie (qui doit être un modèle en arbre), mais plutôt d’un réseau. C’est pourquoi ils ont ajouté deux autres étapes pour transformer le modèle initial de fusion en une structure d’arbre  
\end{itemize}
\item La décomposition du modèle (de réseau) en plusieurs blocs dont les concepts fusionnés sont les frontières.
\item La reconstitution de ce modèle, de sorte que les concepts contenus dans les blocs (à part les concepts fusionnés) soient réorganisés pour former un seul chemin acyclique entre les deux concepts fusionnés. Cette réorganisation va être réalisée à l’aide d’un matching de subsomption / d’inclusion (is-a) entre les concepts de chaque bloc.\\
\end{enumerate}

\begin{figure}[!ht]
\begin{center}
\includegraphics[scale=0.58]{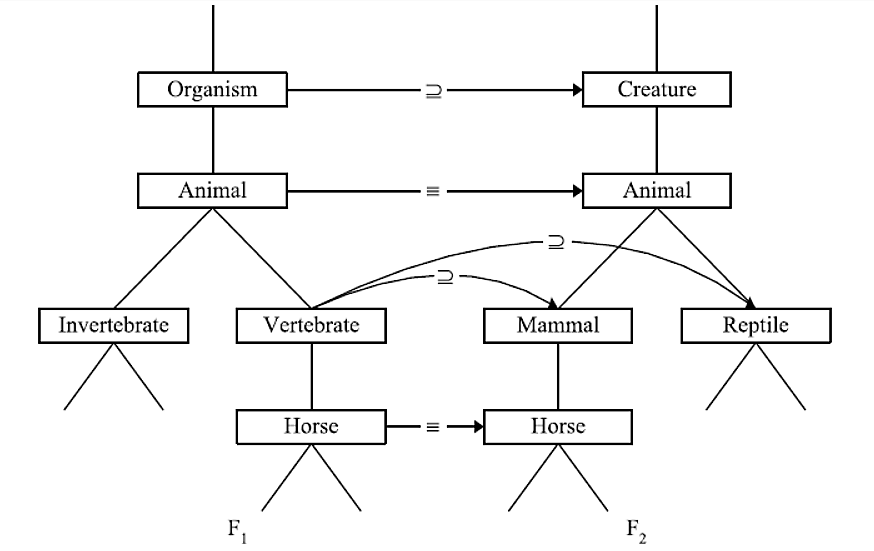}
\caption[]{Correspondances entre deux ontologies \cite{zhang2017oim}}
\label{oim}
\end{center}
\end{figure}

\begin{figure}[!ht]
\begin{center}
\includegraphics[scale=0.58]{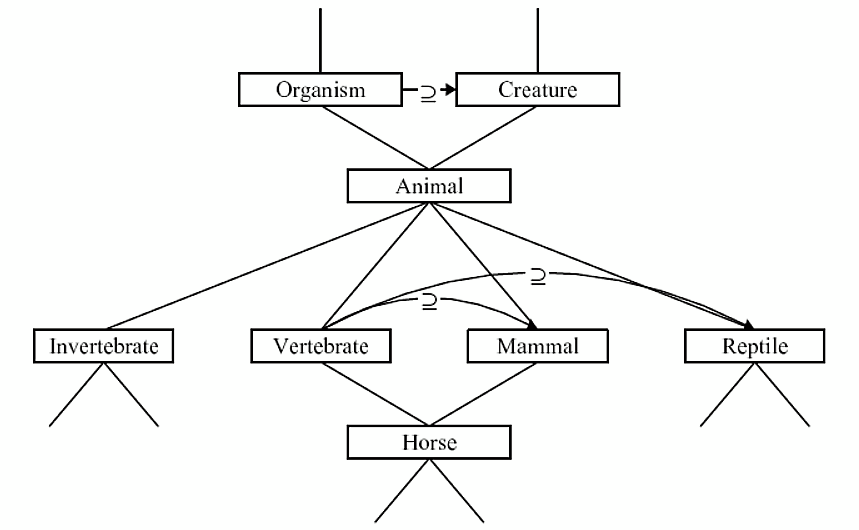}
\caption[]{Fusion initiale des fragments d’ontologies \cite{zhang2017oim}}
\label{oim2}
\end{center}
\end{figure}

\begin{figure}[!ht]
\centering
\includegraphics[scale=0.7]{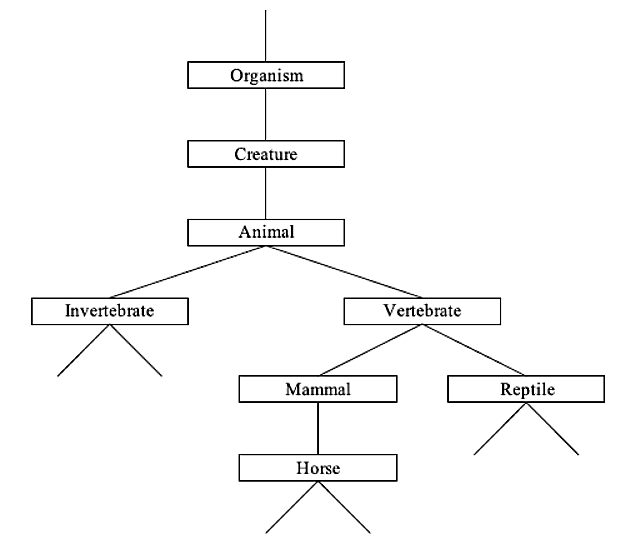}
\caption[]{Résultat final de la fusion de \cite{zhang2017oim}}
\label{oim3}
\end{figure}

Dans la figure \ref{oim}, ils donnent un exemple de correspondances entre deux fragments d’ontologies à fusionner. Dans l'image \ref{oim2}, ils donnent un modèle qui illustre le résultat des deux premières étapes, où les concepts fusionnés ont plus qu’un super-concept direct (héritage multiple), ce qui forme une structure de réseau. Dans la figure \ref{oim3}, ils donnent le modèle qui illustre le résultat des deux dernières étapes, où la sortie finale est une ontologie ayant les concepts réorganisés sous forme d’arbre.

\newpage
Dans les expérimentations, ils ont fusionné l’ontologie BCO (Biological Collections Ontology) qui contient 127 concepts, avec l’ontologie ACO (Animal in Context Ontology) qui contient 510 concepts, dans 9 minutes ; et ils ont utilisé comme référence une fusion retournée artificiellement.\medskip\vspace{4px}

$\circleddash$ Il s’agit bien de très petites ontologies fusionnées en un temps relativement long.

\section{Intégration des ontologies}
En général, les techniques d’intégration des ontologies sont utilisées dans le développement des ontologies ou dans le domaine de l’intégration des données.\medskip\vspace{4px}

L’intégration des ontologies joue un rôle important dans le développement des ontologies en réutilisant des ontologies publiques existantes pour construire une ontologie en cours de développement ; ce qui réduit le coût de l'ingénierie des ontologies et favorise la réutilisation des modules d'ontologies standards. Tenons l’exemple de la construction d'une ontologie de catalogage des bibliothèques qui peut nécessiter l'assemblage d'ontologies dans les domaines des personnes, des livres, des sujets, des coordonnées géographiques, des numéros d'identification des livres, \textit{etc.}

\subsection{Types d’intégration}
Voici les trois types d’intégration d’ontologies selon \cite{keet2004aspects} :

\subsubsection{\underline{Intégration sémantique}}
Elle se focalise sur le sens voulu des entités, \textit{e.g.} découvrir si le concept C1 dans l’ontologie $I$ est synonyme (ou hyponyme ou hyperonyme) au concept C2 dans l’ontologie $II$. \textbf{C'est le type auquel nous nous intéressons.}

\subsubsection{\underline{Intégration structurelle}}
Quand la sémantique est (convenue d’être) identique mais l’organisation des entités (la catégorisation, le schéma) ne l’est pas et doit ainsi être alignée et intégrée. Il faut noter que la distinction entre la sémantique et la structure n’est pas aussi claire que cela puisse paraître, car la structure transporte une interprétation sémantique de la conceptualisation.

\subsubsection{\underline{Intégration syntaxique}}
Elle se concentre sur la réalisation d'un formalisme uniforme à partir des formalismes avec lesquels les ontologies sources sont exprimées, tels que la description logique, KIF, OWL, F-logic \textit{etc.} Dans la méthodologie, ce type d’intégration vient logiquement en troisième position (après l’intégration sémantique et structurelle), car c’est inutile de faire correspondre les formalismes si le sens de ce qui est intégré n’est pas compatible. Cependant, ces traductions, telles que la représentation syntaxique d'un concept dans deux langages formels, peuvent être recherchées indépendamment du domaine de l’intégration (le domaine de traduction des ontologies).

\subsection{Définitions de l'intégration}
\subsubsection{\underline{Intégration comme étant une fusion}}
\begin{figure}[!ht]
\centering
\includegraphics[width=0.95\textwidth]{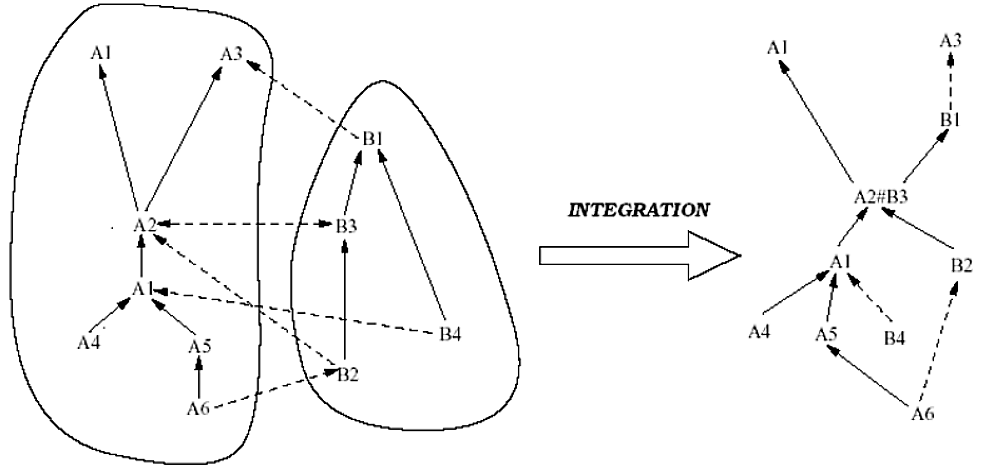}
\caption[Intégration 1]{Intégration des ontologies \cite{mena1996managing}}
\label{mena}
\end{figure}

Comme le montre la figure \ref{mena}, pour \cite{mena1996managing}, l’intégration relie les entités des ontologies à intégrer, en traversant les hyponymes, les hyperonymes, et les synonymes entre eux. C’est en effet une fusion des ontologies.


\subsubsection{\underline{Intégration de Sowa}}
Selon \cite{sowa1997electronic}, l’intégration est "le processus de recherche de points communs entre deux ontologies $A$ et $B$ et de dérivation d'une nouvelle ontologie $C$ facilitant l'interopérabilité entre les systèmes informatiques basés sur les ontologies $A$ et $B$. La nouvelle ontologie $C$ peut remplacer $A$ ou $B$, ou peut être utilisée comme intermédiaire entre un système basé sur $A$ et un autre basé sur $B$". Il n’a pas spécifié les domaines des ontologies à intégrer.\medskip\vspace{4px}

Selon la quantité de changement nécessaire pour dériver $C$ de $A$ et $B$, \cite{sowa1997electronic} distingue trois niveaux d'intégration qui ressemblent un peu à la classification de \cite{heflin2000dynamic} :

\paragraph{Alignement}
C'est la définition 1 de l'alignement (expliquée dans le chapitre 1). Il s'agit du plus bas niveau d’intégration qui ne nécessite aucun changement dans $A$ et $B$. Il supporte l’interopérabilité la plus limitée (le mapping de \cite{heflin2000dynamic}).

\paragraph{Compatibilité partielle}
Elle nécessite plus de changements dans $A$ et $B$, et permet une interopérabilité moyenne. Toute inférence exprimée dans une ontologie en utilisant seulement les entités alignées, peut être traduite en une inférence équivalente dans l'autre ontologie (les révisions de mappings et l'intersection des ontologies de \cite{heflin2000dynamic}).

\paragraph{Unification (Compatibilité totale)}
Elle nécessite des changements ou des réorganisations majeures dans $A$ et $B$, pour entraîner l’interopérabilité la plus complète (le plus haut niveau d’intégration). En effet, tout ce qui peut être fait avec une ontologie peut être fait d'une manière exactement la même avec l'autre. Autrement dit, toute inférence exprimée dans une ontologie, peut être traduite en une inférence équivalente dans l’autre (C’est la fusion de \cite{pinto2004ontologies}).

\subsubsection{\underline{Intégration de Heflin et Hendler}}

Selon \cite{heflin2000dynamic}, l'intégration des ontologies implique généralement l'identification des correspondances entre deux ontologies, la détermination des différences dans les définitions des entités, et la création d'une nouvelle ontologie qui résout ces différences.\medskip\vspace{4px}

\begin{figure}[!h]
\centering
\includegraphics[scale=0.56]{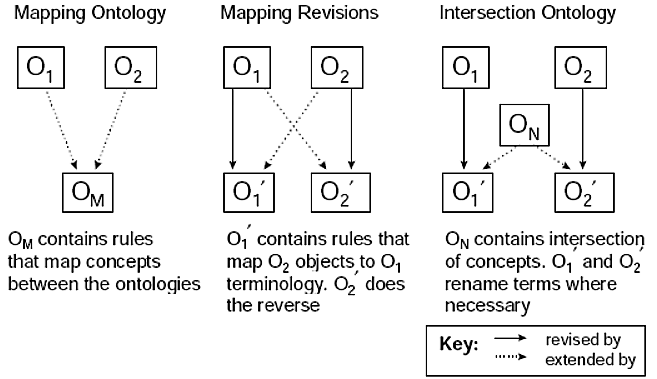}
\caption[Intégration 2]{Intégration de \cite{heflin2000dynamic}}
\label{hef}
\end{figure}

Selon eux, la simple création d'une nouvelle ontologie intégrée ne résout pas le problème d'intégration de l'information sur le Web. En effet, puisque d'autres ontologies et pages Web dépendent des ontologies intégrées, tous les objets dépendants devraient être révisés pour refléter la nouvelle ontologie. Vu que cette tâche est impossible, ils ont suggéré trois façons d'incorporer les résultats de l'intégration dans le Web comme le montre la figure \ref{hef} :

\paragraph{Mapping des ontologies}
C'est la définition 1 du mapping (expliquée dans le chapitre 1). Il s'agit d'une nouvelle ontologie $O_M$, positionnée entre les deux ontologies $O_1$ et $O_2$, qui contient les règles / les axiomes (de transformation) pour mettre en correspondance des entités entre $O_1$ et $O_2$.

\paragraph{Révisions de Mapping}
où $O_1'$, une nouvelle version de $O_1$, contient (à part les entités et les axiomes de $O_1$) les règles qui mettent en correspondance les entités de $O_1$ par rapport à $O_2$, et $O_2'$, une nouvelle version de $O_2$, contient (à part les entités et les axiomes de $O_2$) les règles qui mettent en correspondance les entités de $O_2$ par rapport à $O_1$ (A ne pas confondre avec la notion de révision de mapping qui veut dire le débogage ou la réparation de mapping !!!). \medskip\vspace{4px}

\ding{220} Ils pensent que l’inconvénient de ces deux premières approches est que les concepts partagés entre deux domaines pourraient être également utilisés dans plusieurs autres domaines connexes, ainsi chaque nouveau domaine aurait besoin d'un ensemble de règles pour le mapper à tous les autres domaines en chevauchement. Et cela peut devenir très lourd.

\paragraph{Intersection des ontologies}
où une nouvelle ontologie $O_N$ fusionne et standardise les termes des entités en correspondance de $O_1$ et $O_2$, tout en les renommant dans $O_1'$ et $O_2'$ (qui sont des nouvelles versions de $O_1$ et $O_2$) par les termes fusionnés et standardisés. (C’est la transformation des termes des entités correspondues en des termes communs).\medskip\vspace{4px}

\ding{220} Ils considèrent cette troisième approche comme l’approche la plus naturelle d’intégration des ontologies car elle a l'avantage que l'équivalence des termes soit déterminée dans la phase de pré-traitement plutôt que lors de l'exécution de la requête.\\ \\ \\


\newpage
\subsubsection{\underline{Intégration de Pinto (consensuelle)}}
Dans l’intégration ou la composition des ontologies, \cite{pinto1999towards} considère que nous avons, d'une part, une (ou plusieurs) ontologies ($O_1$, $O_2$, ..., $O_N$) qui vont être intégrées dans l’ontologie cible, et d'une autre part, l'ontologie cible en cours de construction ($O$) qui sera issue du processus de l'intégration. Ainsi, cette méthode est incrémentale. Les domaines des ontologies sources (à intégrer dans l’ontologie cible) sont généralement différents entre eux, et différents du domaine de l'ontologie cible, mais il peut y avoir une relation entre eux ($D_1$, $D_2$, ..., $D_k$).\medskip\vspace{4px}

Il s’agit de deux ou plusieurs ontologies sources de sujets différents (ou de sujets liés) qui vont tout simplement être assemblées, composées, agrégées, ou combinées pour former une ontologie résultante, peut-être après que les ontologies sources aient subi quelques changements, comme l’extension, la spécialisation, la transformation, ou l’adaptation.\medskip\vspace{4px}

L’intégration vise à créer une ontologie d’un nouveau domaine plus large composé de tous les domaines des ontologies d’entrée. C’est un processus de réutilisation qui vise à construire des ontologies à partir d'autres ontologies existantes.\medskip\vspace{4px}

Les ontologies à intégrer doivent répondre à certaines exigences avant de leur appliquer le processus d’intégration, \textit{e.g.}, le domaine, l’abstraction, le type, la généralité, la modularité, l’évaluation, \textit{etc}.\medskip\vspace{4px}

L’ontologie résultante doit avoir toutes les propriétés d'une bonne ontologie : consistante, cohérente, complète, ayant un niveau adéquat de détail, et décrivant seulement le vocabulaire nécessaire pour le domaine, \textit{etc.} Il ne devrait pas avoir une ontologie existante similaire à la résultante, sinon il faudrait tout simplement réutiliser l'ontologie existante.\medskip\vspace{4px}

\noindent Avant leur inclusion dans l’ontologie résultante, les entités de l’ontologie à intégrer peuvent être :
\begin{itemize}
\item[$\bullet$] incluses (utilisées telles quelles);
\item[$\bullet$] spécialisées (conduisant à une ontologie plus spécifique dans le même domaine);
\item[$\bullet$] augmentées (étendues) par de nouvelles entités manquantes (soit par des entités plus générales, soit par des entités de même niveau);
\item[$\bullet$] adaptées (modifiées) pour les corriger ou les améliorer en changeant :
\begin{itemize}
\item leurs terminologies (pour se conformer aux règles de normalisation des noms, ou introduire une terminologie standard ou plus usuelle),
\item leurs documentations (pour les mettre à jour ou améliorer leur clarté),
\item leurs définitions (pour les mieux représenter dans le domaine concerné);
\end{itemize}
\item[$\bullet$] retirées (à cause de leur non pertinence).
\end{itemize}
Ces adaptations transforment l'ontologie source choisie en l'ontologie voulue.\medskip\vspace{4px}

Ils précisent que des problèmes tels que la cohérence, la consistance, et le niveau de détail de l'ontologie résultante doivent être traités (\textit{i.e.} elle ne doit pas avoir des parties de niveau de détail exagéré et d'autres de niveau adéquat).\\

\begin{figure}[!ht]
\begin{center}
\includegraphics[width=\textwidth]{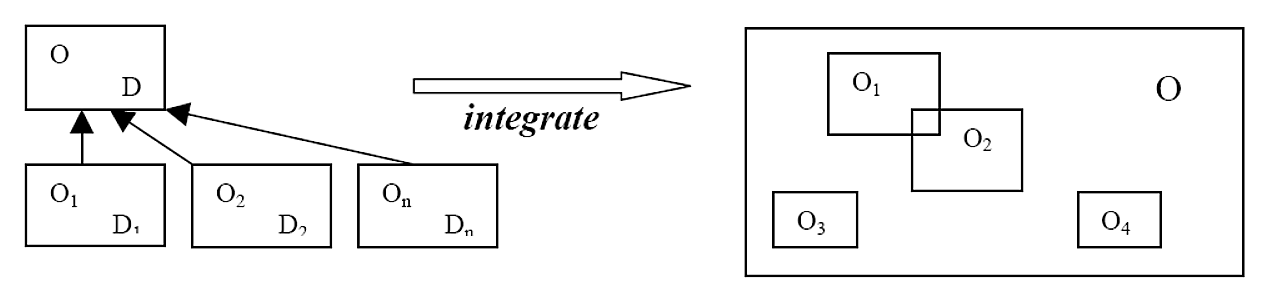}
\caption[Intégration 3]{Intégration de \cite{pinto1999towards} interprétée par \cite{keet2004aspects}}
\label{pint}
\end{center}
\end{figure}

La figure \ref{pint} illustre leur définition.\newpage

\subsubsection{\underline{Intégration comme étant synonyme à la fusion}}
\cite{klein2001combining} considère l'intégration et la fusion comme égales. Il les définit par la création d'une nouvelle ontologie à partir de deux ou plusieurs ontologies existantes qui se chevauchent. C’est une définition très vague qui ressemble à la définition de la fusion des ontologies de \cite{noy1999algorithm}.\medskip\vspace{4px}

\cite{raunich2012towards} et \cite{zhang2017oim} utilisent les termes "fusion" et "intégration" comme des synonymes.

\subsubsection{\underline{Intégration comme étant une ontologie de pont}}
Selon \cite{udrea2007leveraging}, l'intégration des ontologies est l'ajout des axiomes de l’alignement $A$ (entre $O1$ et $O2$) à l’union de $O1$ et $O2$ produisant à la fin une ontologie consistante et cohérente. Les correspondances de $A$ sont utilisées pour créer des liens logiques (des axiomes) qui représentent la sémantique des relations entre les différentes entités (l'équivalence, la subsumption, la disjonction \textit{etc.}).\medskip\vspace{4px}

Selon \cite{euzenat2007ontology}, l'intégration des ontologies est l'inclusion dans une ontologie $O$ d'une autre ontologie $O'$ et des assertions exprimant des liens entre ces deux ontologies (des axiomes de pont). L'ontologie résultante $O$ est censée contenir la connaissance des deux ontologies initiales ($O$ et $O'$). Il n'y a pas vraiment de différence entre leurs définitions de fusion et d'intégration, à part le fait que, selon eux, contrairement à la fusion qui ne modifie pas les ontologies d'entrée, dans l'intégration, l’ontologie source $O'$  est inchangée tandis que l’ontologie initiale $O$ est modifiée (plutôt augmentée par $O'$). En d'autres termes, l'intégration, selon eux, se fait d'une manière incrémentale, alors que la fusion se fait d'une manière non incrémentale. \medskip\vspace{4px}

\ding{220}\textbf{ C’est l’approche que nous allons implémenter.}

\subsubsection{\underline{Autres définitions}}
D’après \cite{kokla2006guidelines}, une intégration d’ontologies génère une nouvelle ontologie intégrée sans modifier les originales.\medskip\vspace{4px}

Selon \cite{umer2012semantically}, l’intégration des ontologies est un processus de construction d’une nouvelle ontologie en utilisant des ontologies disponibles. Elle peut être divisée en trois différents scénarios :
\begin{itemize}
\item Le mapping (définition 1 de l'alignement dans le chapitre 1);
\item L’intégration (la réutilisation);
\item La fusion.
\end{itemize}\medskip\vspace{4px}

Selon \cite{wroblewska2012methods}, il existe différents types d'intégration des ontologies :
\begin{itemize}
\item L’alignement, le mapping (définition 1 de l'alignement dans le chapitre 1)
\item La transformation
\item La fusion
\item L'intégration, \textit{etc.}
\end{itemize}

\subsection{Principaux outils d’intégration et leurs limites}
Les approches récentes d'intégration des ontologies suivent un schéma modulaire qui décompose ce problème en sous-problèmes indépendants tels que le matching et puis la composition. De cette façon, elles profitent des grands progrès déjà réalisés dans le domaine du matching automatique des ontologies qui identifie les entités à intégrer dans le deuxième sous-problème (la composition).\medskip\vspace{4px}

$\circleddash$ La plupart des outils actuels d’intégration des ontologies sont semi-automatiques, car les experts de domaine et les ingénieurs de connaissances sont souvent nécessaires pour aider à la prise de décisions.

\subsubsection{\underline{L’outil ILIADS}}
L’algorithme ILIADS de \cite{udrea2007leveraging} prend en entrée deux ontologies consistantes $O1$ et $O2$, et retourne en sortie un alignement $A$ entre $O1$ et $O2$, de telle sorte que l’intégration future de $O1$ et $O2$ à travers $A$ soit consistante et cohérente. Les auteurs combinent un algorithme de matching de similarité (lexicale, structurelle, extensionnelle, et de clusters) avec un algorithme d'inférence logique qui raisonne sur les conséquences des relations (des correspondances) de l’alignement. Les relations de l’alignement sont exprimées comme des axiomes OWL (\textit{{\NoAutoSpacing owl:equivalentClass}}, \textit{{\NoAutoSpacing owl:equivalentProperty}}, \textit{{\NoAutoSpacing owl:sameAs}}) ajoutés à l'ontologie résultante (qui compose $O1$ et $O2$).\\

Ils ont utilisé le raisonneur Pellet pour vérifier, à la fin de chaque expérience, si l'ontologie résultante est consistante ou pas. ILIADS a produit des inconsistances dans moins de 0,5\% de leurs essais.\medskip\vspace{4px}

$\circleddash$ Mais ils n’ont pas évoqué les incohérences (\textit{e.g.} le nombre de classes insatisfiables), car une ontologie peut être consistante tout en ayant une multitude de classes insatisfiables. (Nous expliquerons plus les notions d'inconsistance et d'incohérence dans le chapitre 4).\newpage

\subsubsection{\underline{L’outil ContentMap (le plus proche de notre travail)}}

Pour \cite{jimenez2009ontology}, un ensemble de correspondances (d’un Mapping) est représenté par une ontologie $M$, où les correspondances sont des axiomes de la forme $subClassOf(e, e')$, $equivalentClass(e, e')$, et $disjointWith(e, e')$ pour la relation de subsomption, d’équivalence, et de disjonction respectivement ; et les identifiants et les valeurs de confiance des correspondances sont des annotations d’axiomes qui n’ont aucun effet sur les inférences.\medskip\vspace{4px}

Leur approche qui consiste en un processus interactif semi-automatique composé de quatre étapes principales (le calcul des correspondances, le calcul des nouvelles inférences, la détection des erreurs, et la réparation des erreurs identifiées à l’aide de l’utilisateur) a été appliquée pour réaliser l’outil "ContentMap", un plugin téléchargeable destiné à être utilisé dans Protégé 4.\medskip\vspace{4px}

ContentMap permet à l’utilisateur de choisir un ou plusieurs outils d’alignement à sélectionner (comme OLA, AROMA, CIDER, \textit{etc.}) en leur attribuant différents poids. Il lui permet aussi de filtrer automatiquement les correspondances en choisissant un seuil de confiance minimal.\medskip\vspace{4px}

Le calcul des nouvelles inférences a été fait à l’aide de la notion de "différence déductive" qui compare les axiomes de l’ontologie résultante $U$ (la composition de $O1$, $O2$, et $M$ (Mapping)) avec les axiomes de chacune des ontologies initiales ($O1$, $O2$, et $M$) pour détecter ceux qui existent dans l’ontologie de sortie mais qui n’existent pas dans les ontologies initiales,\textit{ i.e.} le système calcule la différence logique entre les inférences avant et après l’application des correspondances. Pour aider l’utilisateur à comprendre les conséquences sémantiques de $U$, ils lui montrent les justifications et les explications derrière la manifestation de ces nouveaux axiomes.\medskip\vspace{4px}

Les inférences imprévues de l’ontologie de sortie $U$ (\textit{i.e.} les inférences qui se trouvent dans l’ontologie résultante $U$ mais qui ne se trouvent pas dans les ontologies individuelles $O1$, $O2$, et $M$) seront présentées à l’utilisateur qui aidera à réparer $U$ en choisissant la ou les source(s) d’axiomes à modifier ($O1$, $O2$, et/ou $M$) et en supprimant les axiomes non désirés (qui sont, selon lui, générateurs d’incohérences) :\vspace{4px}
\begin{itemize}
\item[$\bullet$] Il peut choisir de supprimer uniquement des axiomes provenant de l’ontologie $M$ et laisser les axiomes des ontologies initiales $O1$ et $O2$ intacts, puisque $M$ est généralement considérée la plus grande source d’erreurs.\vspace{4px}
\item[$\bullet$] Mais si deux entités correspondues sont originairement contradictoires dans $O1$ et $O2$, et leur relation dans $M$ est correcte, l’utilisateur sera en face d’un dilemme; soit supprimer l’axiome de la correspondance correcte, soit supprimer les axiomes originaux dans l’une ou les deux ontologies $O1$ et $O2$.
\end{itemize}\medskip\vspace{4px}

Par suite, le système exécute un algorithme de réparation qui essaie de supprimer le minimum d’axiomes tout en préservant les inférences jugées valides par l’utilisateur.\\

Dans leurs expérimentations, ils ont utilisé quatre petites ontologies qui décrivent toutes le domaine des références bibliographiques, mais qui sont développées séparément. Leur taille varie de 130 entités (58 classes, 46 propriétés d’objet, et 26 propriétés de données) à 49 entités (18 classes, 12 propriétés d’objet, et 19 propriétés de données) et de 235 axiomes à 96 axiomes.\medskip\vspace{4px}

O-INR est l’ontologie avec laquelle les trois autres ontologies (O-MIT, O-UMBC, et O-AIFB) ont été intégrées chacune à part. Il existe trois alignements de référence : un alignement pour O-MIT et O-INR contenant 119 correspondances, un alignement pour O-UMBC et O-INR contenant 83 correspondances, et un alignement pour O-AIFB et O-INR contenant 98 correspondances.\medskip\vspace{4px}

Ils ont intégré O-MIT, O-UMBC, et O-AIFB séparément avec O-INR en utilisant leurs alignements de référence correspondants, puis ils ont évalué les conséquences sémantiques de leurs ontologies résultantes. Dans tous les cas, ils ont trouvé un nombre signifiant d’inférences imprévues. Par exemple, lors de l’intégration de O-AIFB et O-INR, ContentMap a détecté 34 concepts insatisfiables (originaires des deux ontologies) pour lesquels il y a eu un nombre énorme de justifications (généralement complexes) qui ont rendu la réparation manuelle extrêmement difficile. Egalement, lors de l’intégration de O-MIT et O-INR, ContentMap a détecté des nouvelles subsomptions qui ont été identifiées et réparées automatiquement.\medskip\vspace{4px}

D’un point de vue temps, le goulot d’étranglement est le calcul de toutes les justifications des inférences imprévues. Une fois les justifications calculées, le temps de réparation est, selon eux, relativement court (ils ne l’ont pas précisé).\medskip\vspace{4px}

Ils ont remarqué aussi que l’utilisation des correspondances générées automatiquement a abouti à un plus grand nombre d’inférences imprévues, \textit{e.g.} quand ils ont intégré O-AIFB et O-INR en utilisant l’outil de matching CIDER avec un seuil de confiance égale à 0.1, ils ont trouvé 55 concepts insatisfiables et 34 subsomptions imprévues qui sont des erreurs causées principalement par des correspondances incorrectes.\medskip\vspace{4px}

$\circleddash$ Pour conclure, les auteurs ont fait une fusion (non pas une intégration), précisément une ontologie de pont, semi-automatique à de petites ontologies de même domaine, et malgré la réparation des alignements d'entrée, ils trouvent toujours énormément de classes insatisfiables dans l'ontologie de sortie. Ils n'évoquent pas le temps d'exécution de ces expérimentations.

\subsubsection{\underline{L’outil de Ziemba \textit{et al.}}}
L’algorithme d'intégration des ontologies de \cite{ziemba2015integration} utilise essentiellement les outils de refactoring et d’intégration de l’éditeur "Protégé" qui facilitent énormément leur processus d'intégration. Il est divisé en trois parties :
\begin{enumerate}
\item L'intégration de la première ontologie :
\begin{enumerate}
\item Créer une nouvelle ontologie vide dans l'éditeur Protégé (Ils ont choisi de lui donner l'IRI (l’identificateur) : "Intégrée");
\item Sélectionner la première ontologie source à intégrer; ouvrir l'ontologie cible "Intégrée" dans l'éditeur Protégé et importer l'ontologie source sélectionnée;
\item Utiliser l'option "Merge ontologies" de Protégé pour intégrer l'ontologie source dans la destination ("Intégrée"). De cette façon, l’ontologie source sera incluse dans le même fichier de l’ontologie cible;
\item Puis, en utilisant les outils de refactoring, changer l’IRI de tous les éléments de l'ontologie source en "Intégrée" qui est l’IRI de l'ontologie cible.
\end{enumerate}
\item La sélection, l’importation et le refactoring d’une nouvelle ontologie source dans l'ontologie cible, et faire l’alignement entre elles, en utilisant les dictionnaires, les thésaurus, et les outils de "Protégé", pour l’introduire ensuite sous forme de relations d'équivalence et de subsomption entre les entités des ontologies source et cible.
\item La vérification de la consistance et la cohérence de l’ontologie cible (en utilisant un raisonneur dans Protégé) et la vérification de l'absence de redondances, puis leur résolution en éliminant les relations "is-a" redondantes entre les entités (causées par les relations d’équivalence).
\end{enumerate}
Les étapes 2 et 3 sont itératives (elles se répètent pour chaque ontologie source à intégrer).\medskip\vspace{4px}

$\circleddash$ Cette approche est simple et claire, et bien que nous n’ayons pas la moindre idée sur la qualité du matching réalisé, son inconvénient majeur est qu’elle est toute manuelle (elle est difficile à appliquer pour les grandes ontologies).\medskip\vspace{4px}

Dans les expérimentations, ils ont créé une ontologie cible vide dans laquelle ils ont intégré l’ontologie source eQual. Puis ils ont intégré l’ontologie source Ahn dans l’ontologie cible (contenant déjà eQual) en reliant les paires d’entités mises en correspondance par des relations d’équivalence. De la même manière, ils ont intégré trois autres ontologies sources (SiteQual, Website Evaluation Questionnaire, et Web Portal Site Quality) dans l’ontologie cible. Toutes ces ontologies sources appartiennent au domaine de l’évaluation de la qualité des sites Web. Dans chacune de ces itérations, le raisonneur a détecté des incohérences et des relations de subsomption redondantes dans l’ontologie résultante qu’il fallait corriger.\medskip\vspace{4px}

$\circleddash$ Il s'agit d'une \textbf{fusion}, non pas d'une intégration. Dans ce travail, il n’y a pas une partie d’évaluation à l’aide d’une référence ou des résultats d’un travail concurrent pour se comparer avec.

\subsubsection{\underline{L’outil FITON pour l’intégration des data sets}}
\cite{zhao2014ontology} ont proposé un système semi-automatique, nommé FITON, qui prend en entrée des data sets (deux ou plusieurs ontologies) du LOD (Linked Open Data) et qui retourne en sortie une ontologie intégrée (et enrichie).\medskip\vspace{4px}

Dans ce cas, la problématique c’est qu’il existe très peu (ou pas) de liens d’équivalence préétablis entres les différentes classes et propriétés des data sets (contrairement aux liens \textit{"sameAs"} disponibles avec abondance entre leurs instances - des centaines de millions -) ; ce qui rend difficile l’extraction directe des classes et des propriétés équivalentes pour faire l’intégration de ces data sets.\medskip\vspace{4px}

Pour ce faire, ils ont intégré tout d’abord les instances identiques (informations déjà fournies sous forme de propriétés \textit{"{\NoAutoSpacing owl:sameAs}"} dans les data sets) et ont extrait les classes et les propriétés qui décrivent ces instances pour former un graphe nommé graphe \textit{"sameAs"} à partir duquel ils vont découvrir les similarités entre les différentes classes et propriétés contenues dans le graphe en combinant des méthodes de matching (terminologiques et sémantiques) pour parvenir enfin à intégrer tous les types d’entités des data sets.\medskip\vspace{4px}

Puisqu’il n’existe pas de benchmark pour les ensembles de données du LOD, un expert a créé manuellement un alignement de référence entre les deux data sets DBpedia et Geonames. DBpedia, la version "linked data" de Wikipédia de domaine transversal, contient (au moment de l'expérimentation) 3 708 696 instances, 241 classes, et 1 385 propriétés ; et Geonames, de domaine géographique, contient (au moment de l'expérimentation) 7 480 462 instances, 428 classes, et 31 propriétés.\medskip\vspace{4px}

$\circleddash$ Au final, leur ontologie de sortie contenait seulement 135 classes et 453 propriétés. Nous concluons qu’ils ont fait une sorte d’intersection entre les deux ontologies plutôt qu’une intégration, car la conservation des informations initiales des ontologies d’entrée n’est pas respectée.\\

$\circleddash$ Puisqu’ils comptent sur l’analyse des instances inter-liées (sameAs) pour découvrir les alignements entre les data sets, leur outil ne pourra pas fonctionner s’il n’existe pas (ou s’il n’existe que peu) de liens "sameAs" entre les instances. Ainsi, l’inconvénient de FITON est qu’il ne peut fonctionner efficacement que lorsqu’au moins 4\% des instances d’un data set soient liées (identiques) aux instances de l’autre data set.

\section{Différences entre la fusion et l’intégration}

L'intégration et la fusion des ontologies sont tous les deux des processus de construction d'une nouvelle ontologie en se basant sur les informations de deux ou plusieurs ontologies sources. Cependant, dans la fusion, il y a beaucoup de connaissances en chevauchement entre les entités des ontologies sources, alors que dans l’intégration (la composition), il y a peu ou pas de chevauchement \cite{pinto1999towards}.\medskip\vspace{4px}

Par conséquent, la différence principale entre ces deux processus est que, après le processus d’intégration (composition), nous pouvons identifier dans l'ontologie résultante les régions issues des ontologies sources, car les connaissances ont été laissées plus ou moins inchangées. En effet, l'ontologie résultante est composée de modules (sous-ontologies). Tandis qu'après le processus de fusion, il est généralement difficile d'identifier dans l’ontologie résultante les régions issues des ontologies sources car les connaissances ont été mêlées ou unifiées et homogénéisées ainsi modifiées.

\section{Avantages de ces deux processus}
Dans le contexte de l’ingénierie des ontologies, la réutilisation des modèles de connaissances existants est recommandée comme étant un facteur clé pour le développement d’ontologies rentables et de haute qualité. \cite{ziemba2015integration} ont conclu que la construction d'une ontologie à l’aide de l'intégration ou de la fusion des ontologies sources, est beaucoup moins complexe que son processus de construction à partir de zéro.\medskip\vspace{5px}

L’intégration des ontologies réduit le coût et le temps requis pour la conceptualisation des domaines à partir de zéro, et améliore la qualité des ontologies nouvellement développées pour une application particulière en réutilisant des composants déjà validés.

La fusion des ontologies évite la confusion qui peut être générée à partir de plusieurs représentations du même domaine et renforce l'orchestration et l'harmonisation des connaissances.\medskip\vspace{4px}

Ces deux processus sont aussi particulièrement utilisés lorsqu'il est nécessaire d'effectuer un raisonnement impliquant plusieurs ontologies.

\section{Conséquences sémantiques de ces deux processus}
\cite{jimenez2009ontology} affirment que quand nous raisonnons sur les ontologies à intégrer et leurs correspondances, sur l’ontologie produite suite à l'intégration (contenant les axiomes des correspondances), ou sur l’ontologie produite suite à la fusion, il est souvent nécessaire de détecter des conflits; ainsi des erreurs vont probablement se manifester. Ces erreurs sont dues à deux causes principales :\vspace{4px}

\begin{itemize}
\item[$\bullet$] Les correspondances (générées généralement par un outil de matching automatique) peuvent comporter quelques fautes ou être erronées (incorrectes).\vspace{4px}
\item[$\bullet$] Même si les correspondances trouvées étaient toutes correctes, les ontologies à intégrer peuvent contenir des descriptions contradictoires des entités correspondues, et ceci est à cause de la représentation variable des ontologies sources. En effet, selon \cite{cheatham2017semantic}, les correspondances qui forment l’alignement ne sont pas indépendantes les unes des autres :\vspace{4px}
\begin{itemize}
\item[$\circ$] Il y a des cas dans lesquels seulement une parmi plusieurs correspondances peut être vraie (parmi les correspondances ayant la même entité source).
\item[$\circ$] Dans d’autres cas, plusieurs correspondances, réunies ensemble, peuvent conduire à une inférence involontaire et indésirable ou une classe insatisfiable.
\end{itemize}\vspace{4px}
\item[$\bullet$] Ou la combinaison de ces deux causes.
\end{itemize}\medskip\vspace{4px}

Ces erreurs sont des conséquences logiques imprévues (\textit{e.g.} des classes insatisfiables, de nouvelles subsomptions (suite aux axiomes d’équivalence), \textit{etc.}) difficiles à détecter, à comprendre, et à réparer. Par conséquent, l’ontologie globale créée, associée aux règles de correspondances, est très prédisposée aux erreurs et nécessite une supervision d’un expert de domaine ou une supervision automatique supportée par les applications.\medskip\vspace{4px}

Récemment, la recherche a été conduite au debugage et à la révision des alignements, ainsi que le debugage et la réparation des insatisfiabilités dans les ontologies OWL. Cependant, la suppression de certaines insatisfiabilités peut entraîner une perte d'informations précieuses provenant des ontologies sources. D’ailleurs, c'est le plus grand inconvénient de l’intégration ou de la fusion des ontologies dans les travaux actuels.

\subsection{Exemples}

Les mêmes données peuvent être décrites par des ontologies de différentes perspectives. Cependant, même en se concentrant sur une même perspective, la multitude d’ontologies actuellement utilisées pour les décrire, empêche leur intégration transparente. En effet, l’intégration de deux ontologies de différents modèles peut causer des incohérences logiques.\medskip

Supposons que $A$ est une classe satisfiable dans $O1$, et $B$ et $C$ sont des classes disjointes dans $O2$, et supposons que deux correspondances (d’un alignement entre $O1$ et $O2$) disent que $A$ est une sous classe de $B$, et que $A$ une est sous classe de $C$. Si ces correspondances, exprimées sous forme d’axiomes, sont ajoutées à la composition des deux ontologies $O1$ et $O2$, ils vont créer un problème car une classe ne peut pas être une sous classe de deux classes mères disjointes. Par conséquent, $A$ sera insatisfiable et l’ontologie résultante sera incohérente. Et si jamais $A$ avait une instance, l’ontologie résultante serait inconsistante \cite{abbas2013creating}.\medskip\vspace{4px}

Voici un autre exemple :
\begin{figure}[!h]
\centering
\includegraphics[scale=0.5]{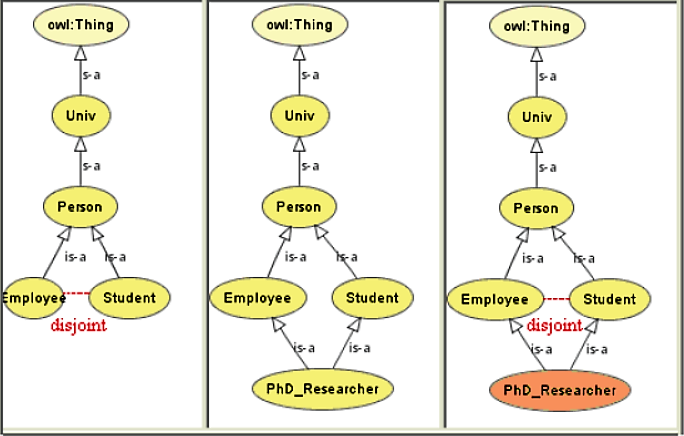}
\caption[Incohérence de l'ontologie fusionnée]{Incohérence de la fusion de $O1$ et $O2$ \cite{fahad2010disjoint}}
\end{figure}

\begin{figure}[!t]
\begin{center}
\includegraphics[scale=0.37]{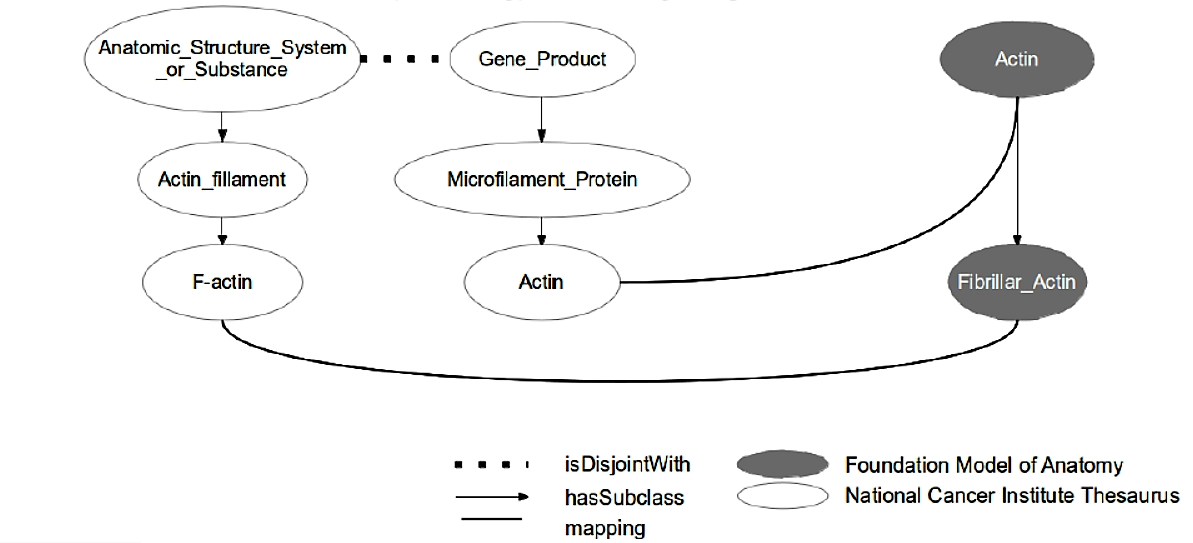}
\caption[Incohérence de l'ontologie de pont]{Incohérence de l'ontologie de pont \cite{cheatham2017semantic}}
\label{expl}
\end{center}
\end{figure}

L'exemple de la figure \ref{expl} illustre une incohérence logique causée par deux correspondances entre l’ontologie "National Cancer Institute Thesaurus" (NCIT) et l’ontologie "Foundational Model of Anatomy" (FMA). Cela se produit car, lors de l’intégration, la classe \textit{Fibrillar\_Actin} devient (suite à l’équivalence) une sous-classe de \textit{Anatomic\_Structure\_System\_or\_Substance} et de \textit{Gene\_Product}, qui sont deux classes disjointes.\\ 

\ding{220} Résoudre ces incohérences est loin d’être facile.

\subsection{Conséquences dans les outils existants}
D’après \cite{fahad2010disjoint}, les outils de fusion ou d’intégration existants ne préservent pas la connaissance disjointe des ontologies sources (pour but de produire à la fin une ontologie cohérente) et sont en général semi-automatiques nécessitant beaucoup d’intervention humaine pour la validation des correspondances suggérées et également la résolution des conflits générés durant ou après la construction de l’ontologie résultante. Par conséquent, le résultat est incomplet (il viole la description des disjonctions) et très dépendant de l’observation et de l’intelligence de l’utilisateur.\medskip\vspace{4px}

La raison de l’inexactitude de ces outils c’est qu’ils n'exploitent pas la sémantique cachée (et précisément les disjonctions) des ontologies sources pendant la phase de matching, et détectent ainsi des correspondances non fiables qui créent des situations erronées dans l’ontologie de sortie.\medskip\vspace{4px}

\cite{udrea2007leveraging} exigent l’exploitation de la sémantique des ontologies pendant la génération des correspondances entre elles (pendant le matching), pour parvenir à créer une ontologie consistante et cohérente suite au processus d’intégration (ou de fusion). Et \cite{fahad2010disjoint} exigent de prêter une attention particulière aux conflits sémantiques générés à cause des relations disjointes dans les ontologies sources.

\subsection*{Remarque}
Les alignements peuvent être réalisés pour aider l’interrogation distribuée ou bien le raisonnement logique \cite{cheatham2017semantic} :\medskip\vspace{4px}

\begin{itemize}
\item[$\bullet$] Pour l’\textbf{interrogation}, le rappel (\textit{i.e.} les résultats pertinents) des correspondances est un aspect important. Cela signifie que, pour les applications centrées sur les requêtes, ce n’est pas grave si les correspondances provoquent d’inconsistance logique, l’essentiel c’est que les relations soient correctes. Ce cas s’applique dans le contexte des linked data, et l’intégration des ontologies tout en les maintenant séparées.\medskip\vspace{4px}
\item[$\bullet$] Pour le \textbf{raisonnement} logique, le rappel n’est pas suffisant car les correspondances peuvent être correctes, mais générant de conflits. En effet, les applications qui ont l'intention d'employer un raisonneur sur les données intégrées ne peuvent pas utiliser un alignement qui génère une inconsistance logique. Ce cas s’applique dans le contexte de l’intégration des ontologies tout en les regroupant en une seule ontologie (la fusion et l’ontologie de pont entre autres).
\end{itemize}

\section{Discussion et synthèse}
Dans notre mémoire, nous allons nous intéresser aux ontologies de pont, qui, suivant les définitions ci-dessus, peuvent entrer dans le cadre de la fusion et de l’intégration, les deux à la fois, mais vu que le terme "intégration" peut être également un terme générique qui inclut la fusion, nous avons choisi d’utiliser le terme intégration, d’où le titre de notre mémoire.\medskip\vspace{4px}

L’ontologie de pont peut être considérée comme le plus faible niveau de "fusion" des ontologies car dans nos expérimentations, nous avons uni des ontologies ayant un domaine identique ou proche ; cependant, elle peut être considérée aussi comme une "intégration", car il s’agit bien d’une composition des ontologies sources de telle manière que les éléments de chacune des ontologies sources soient facilement reconnus dans l’ontologie résultante.\medskip\vspace{4px}

Ce type d’intégration est utilisé par exemple dans le cas où des entreprises en coopération veulent unir leurs connaissances sans tout de même changer leurs ontologies de base, ainsi changer toutes les données qui s’y conforment. En d’autres termes, elles veulent coopérer tout en restant indépendantes. Dans l’ontologie résultante, les noms et les descriptions des entités issues des ontologies sources restent comme ils le sont originairement, sans changer tout un système qui en est dépendant.

\section*{Conclusion}
L’intégration des ontologies est encore plus difficile avec les ontologies de domaines identiques, similaires, complémentaires, et surtout interdisciplinaires. Les difficultés apparaissent aussi avec les ontologies de différents niveaux formels (les ontologies légères et lourdes). Or, les ontologies qui ont beaucoup de points communs dans la structuration et l’organisation de leurs entités ont plus de chance de ne pas avoir des conflits et des difficultés d’intégration.\medskip\vspace{4px}

Il y a de multiples possibilités pour intégrer les ontologies. Les approches peuvent être distinguées par trois facteurs principaux : le niveau d’intégration (d’interopérabilité sémantique), le(s) domaine(s) des ontologies d’entrée, et la méthode d'intégration (incrémentale ou non).\medskip\vspace{4px}

Dans la littérature, le problème de l’intégration des ontologies a été largement étudié au cours des dernières années, mais il reste toujours un défi si nous voulons réaliser une intégration de manière \textbf{automatique}, \textbf{efficace}, sur de \textbf{grandes ontologies}, en \textbf{préservant toutes les données originales}, et \textbf{sans produire d'erreurs} (conflits sémantiques / logiques).
\chapter{Nouvelle méthode d’intégration des ontologies}

\section*{Introduction}
Dans ce chapitre, nous décrivons les étapes de notre méthode et celle de la référence, citons les conditions favorables pour avoir les meilleurs résultats, et proposons une nouvelle terminologie à utiliser au lieu des notions floues et mal appropriées de la littérature.

\section{Description de la nouvelle méthode}
\begin{figure}[p]
\begin{center}
\includegraphics[scale=0.773]{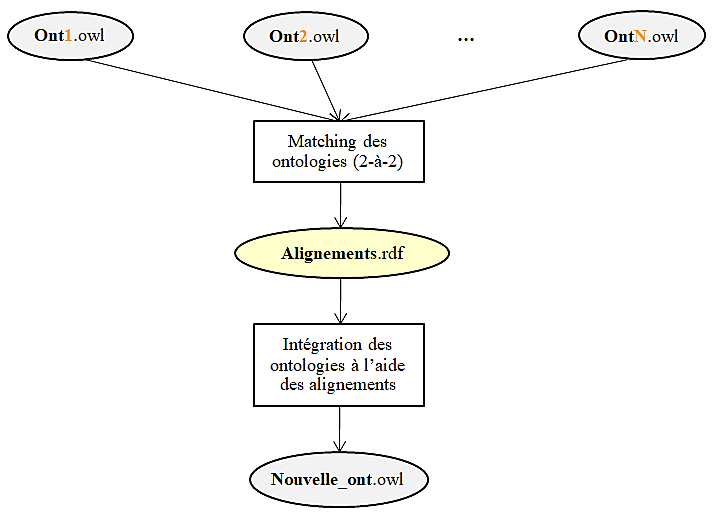}
\caption{Processus général de l'intégration des ontologies}
\label{fig}\vspace{31px}
\includegraphics[scale=0.773]{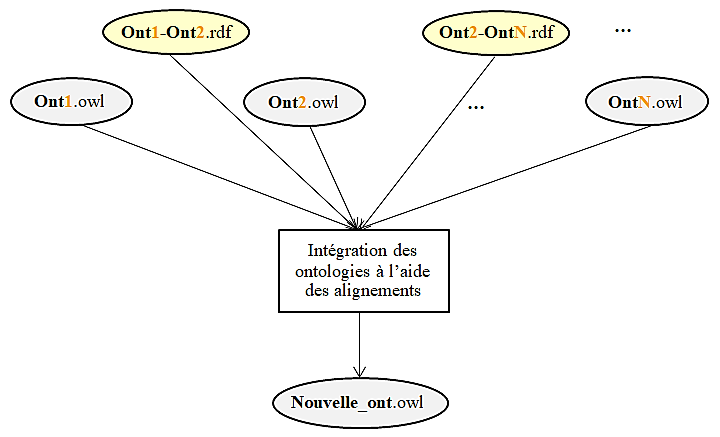}
\caption[Processus général de OIA2R]{Processus général de notre méthode d'intégration des ontologies (OIA2R)}
\label{fig2}
\end{center}
\end{figure}

En général, le processus d’intégration des ontologies passe par deux étapes majeures : le matching des ontologies d’entrée, puis la composition / l’union / l’agrégation de ces ontologies avec les mappings (ou les alignements) générés suite à l’étape de matching. La figure \ref{fig} illustre ce processus.\medskip\vspace{4px}

Mais dans notre travail, nous avons utilisé des alignements de référence comme entrée, sans avoir fait l’étape de matching par nous-mêmes, ainsi nous allégeons notre charge de travail et nous nous concentrons sur l’intégration concrète. Les temps d’exécution de nos expérimentations ne comprennent pas le temps de matching. En effet, le vrai temps d’exécution sera la somme de notre temps global et celui du matching (supposé fait). Ainsi, le temps d’exécution réel de notre méthode dépendra du temps d’exécution de l’algorithme du matching utilisé, et la qualité de l'ontologie résultante dépendra de la qualité des alignements d'entrée utilisés. La figure \ref{fig2} illustre notre processus.\medskip\vspace{4px}

\begin{figure}[p]
\begin{center}
\includegraphics[scale=0.65]{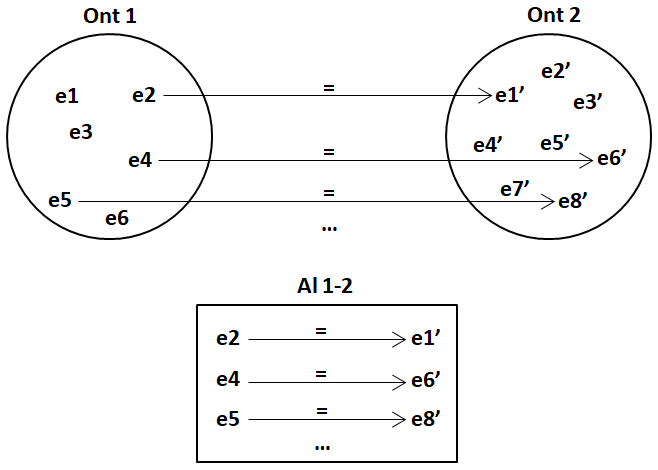}
\medskip\vspace{11px}
\includegraphics[scale=0.57]{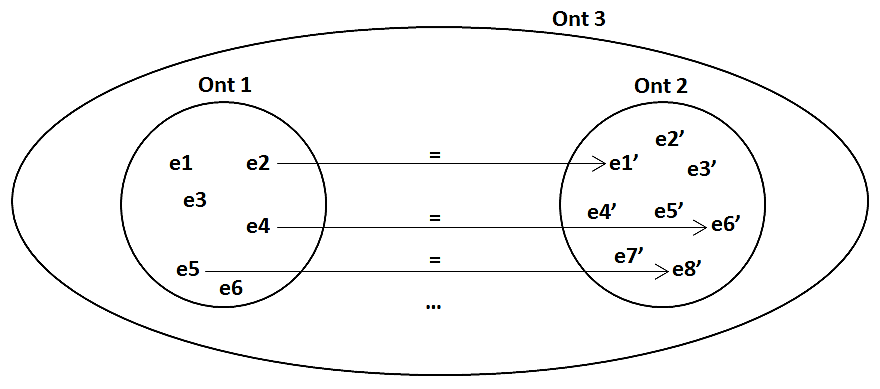}
\medskip\vspace{2px}
\includegraphics[scale=0.57]{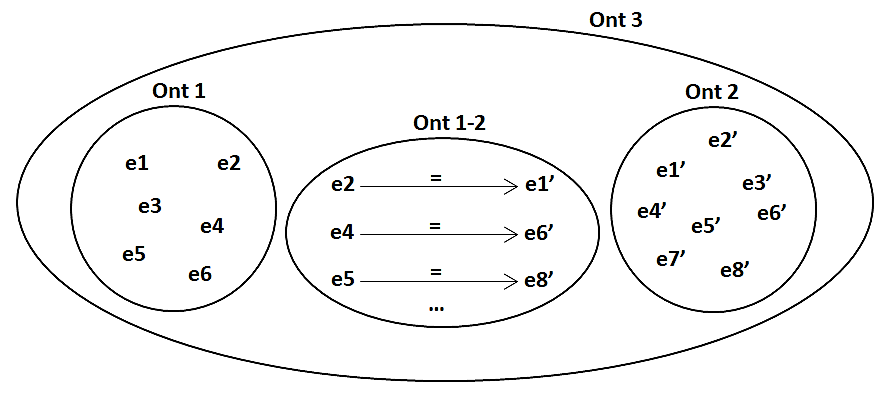}
\medskip\vspace{4px}
\caption[Ontologie de pont]{Ontologie de pont (Ont3)\label{fig3}}
\end{center}
\end{figure}

Nous avons appelé notre algorithme d'intégration "\textbf{OIA2R}" (\textbf{O}ntology \textbf{I}ntegration : \textbf{A}lignment \textbf{R}euse and \textbf{R}efactoring). Ce que notre algorithme génère est une ontologie de pont, \textit{i.e.}, l’union des ontologies d’entrée et des alignements entre eux. Puisque nous allons convertir les correspondances contenues dans les alignements d’entrée en des axiomes OWL, ces alignements sont considérés comme des ontologies OWL intermédiaires (constituées d’entités ayant des relations d’équivalence). Ainsi, implicitement, c’est une union des ontologies sources et des ontologies intermédiaires (l’articulation). Dans le cas de deux ontologies $O_1$ et $O_2$, ayant un alignement $A$ qui peut être vu comme une ontologie $O_A$, le résultat sera une nouvelle ontologie $O_3$ de sorte que $O_3 = O_1 + O_2 + A$, ou plutôt $O_3 = O_1 + O_2 + O_A$. Le schéma de la figure \ref{fig3} illustre nos dires.

\subsection{Approche générale de OIA2R}
Notre implémentation et également celle de la référence sont divisées en deux parties majeures :
\begin{enumerate}
\item La première consiste à composer (assembler) les ontologies d’entrée. Avec ce code seul, nous obtiendrons une ontologie composée des sous-ontologies sources, ainsi contenant les axiomes de description des entités des ontologies sources, sans "bridging" axiomes entre elles. \ding{220} Pour l'instant, il s’agit d’une intégration (composition) simple sans interopérabilité sémantique.
\item La deuxième consiste à ajouter (aux axiomes créés dans la première étape) des axiomes de pont qui sont en fait des axiomes d’équivalence entre les différentes entités. Ce sont des axiomes qui traduisent fidèlement les correspondances provenant des alignements entre les ontologies sources. \ding{220} Ces deux étapes font une intégration qui produit une ontologie dite "ontologie de pont" qui permet l'interopérabilité sémantique.\\
\end{enumerate}

\framebox[0.95\textwidth]{\textbf{Ontologie de sortie = axiomes des entités sources + "bridging" axiomes}}

\subsection{Algorithme général de OIA2R}

\noindent\fbox{\begin{minipage}{\textwidth}
\texttt{-- Saisie des ontologies d'entrée}

\texttt{-- Saisie des alignements d'entrée}\bigskip

\texttt{-- [Étape 1] Parsing et sauvegarde des axiomes correspondants aux entités et aux descriptions des entités des ontologies d'entrée, tout en personnalisant ces entités}\bigskip

\texttt{\textbackslash \textbackslash  -- Réparer les alignements d'entrée (avec LogMap, ou ALCOMO)}

\texttt{\textbackslash \textbackslash  -- Convertir les alignements d'entrée (1-à-N) en des mappings (1-à-1)}\bigskip

\texttt{-- [Étape 2] Parsing et sauvegarde des axiomes correspondants aux cellules (aux correspondances) des alignements d'entrée, tout en personnalisant leurs paires d'entités exactement comme nous l'avons fait pour ces mêmes entités dans l'étape 1}

\texttt{\textbackslash $\ast$ Si cette étape est supprimée, nous aurons juste une agrégation des ontologies entrée $\ast$\textbackslash}\bigskip

\texttt{-- Création de l'ontologie de sortie à partir de tous les axiomes sauvegardés}.

\end{minipage}
}

\section{Démarche détaillée de OIA2R}
\subsection{Introduction au refactoring}

Un IRI d’une ontologie (Internationalised Resource Identifier) identifie l’ontologie d’une façon unique. Il est considéré comme son nom. Il peut être un IRI physique, \textit{i.e.} un fichier .owl local, ou bien un URI à publier dans le Web, \textit{i.e.} une adresse Web contenant ce fichier .owl.\medskip\vspace{4px}

L'IRI d’une entité (classe, propriété, ou instance) est composé d’un "IRI de préfixe" (un préfixe) suivi du "nom" court de l’entité (un suffixe). En général, la partie "préfixe" de l’entité est exactement l’IRI de l’ontologie actuelle (ou d’une autre ontologie existante), mais elle peut aussi contenir en plus un "ID" (identifiant), juste après l’IRI de l’ontologie :

\begin{figure}[!ht]
\centering
\includegraphics[scale=0.5]{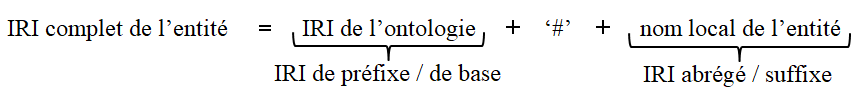}
\includegraphics[scale=0.5]{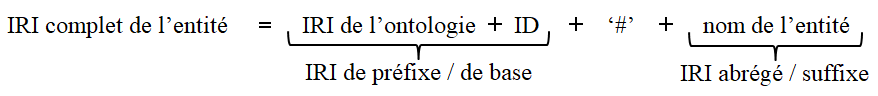}
\end{figure}

Dans une ontologie, nous ne pouvons pas avoir deux IRIs identiques, \textit{i.e.} si deux entités (deux objets créés dans OWL API) ont le même IRI, alors après leur création, il s’agira de la même entité. Ainsi, dans une ontologie, une entité (un IRI complet d’une entité) doit être unique. Mais quand nous allons intégrer des ontologies de même domaine en une seule ontologie, il y aura un grand risque de rencontrer des entités, même si originaires d’ontologies différentes (donc d’IRIs de préfixe différents), mais ayant exactement le même "nom local" (suffixe). Ceci nous causera un problème, car nous voulons, conformément aux standards, que les entités de notre future ontologie aient comme "IRI de préfixe" l’IRI de cette dernière. Dans ce cas, pendant la création de ces entités ayant le même "nom", seulement une sera créée, et elle aura dans sa description toutes les informations de ces entités. Elle sera bien évidemment sémantiquement erronée.\medskip\vspace{4px}

Voici un exemple pour concrétiser ce que nous étions en train de dire. Les entités de nom "\textbf{Paper}", "\textbf{Person}", et "\textbf{Conference}" existent dans au moins trois ontologies de la base "Conference" : \textbf{cmt} (Ont\textbf{1}), \textbf{conference} (Ont\textbf{2}), et \textbf{confOf} (Ont\textbf{3}) :\medskip\vspace{4px}

\begin{itemize}
\item[$\bullet$] Voici les IRIs complets originaux des entités ayant le nom "\textbf{Paper}" :
\begin{itemize}
\item \underline{http://cmt\#Paper}
\item \underline{http://conference\#Paper}
\item \underline{http://confOf\#Paper}
\end{itemize}\medskip\vspace{4px}

\item[$\bullet$] Voici les IRIs complets originaux des entités ayant le nom "\textbf{Person}" :
\begin{itemize}
\item \underline{http://cmt\#Person}
\item \underline{http://conference\#Person}
\item \underline{http://confOf\#Person}
\end{itemize}\medskip\vspace{4px}

\item[$\bullet$] Voici les IRIs complets originaux des entités ayant le nom "\textbf{Conference}" :
\begin{itemize}
\item \underline{http://cmt\#Conference}
\item \underline{http://conference\#Conference}
\item \underline{http://confOf\#Conference}\\
\end{itemize}
\end{itemize}

Si l’IRI de notre nouvelle ontologie à créer était : \underline{http://intégration},\medskip\vspace{4px}

\begin{itemize}
\item[$\bullet$] Voici comment allaient paraître les IRIs complets des entités de nom "\textbf{Paper}" dans la nouvelle ontologie :
\begin{itemize}
\item \underline{http://intégration\#Paper}
\item \underline{http://intégration\#Paper}
\item \underline{http://intégration\#Paper}
\end{itemize}\medskip\vspace{4px}

\item[$\bullet$] Voici comment allaient paraître les IRIs complets des entités de nom "\textbf{Person}" dans la nouvelle ontologie :
\begin{itemize}
\item \underline{http://intégration\#Person}
\item \underline{http://intégration\#Person}
\item \underline{http://intégration\#Person}
\end{itemize}\medskip\vspace{4px}

\item[$\bullet$] Voici comment allaient paraître les IRIs complets des entités de nom "\textbf{Conference}" dans la nouvelle ontologie :
\begin{itemize}
\item \underline{http://intégration\#Conference}
\item \underline{http://intégration\#Conference}
\item \underline{http://intégration\#Conference}\\
\end{itemize}
\end{itemize}
Mais ceci est impossible car un IRI d’une entité est unique, il ne se répète guère.\medskip\vspace{4px}

Pour pallier cette redondance et ne pas mettre les informations (les définitions) de toutes ces entités dans l’ontologie de sortie dans une seule entité ayant ce nom, nous avons choisi d’ajouter un ID aux IRIs de préfixe des entités de notre ontologie, pour pouvoir les différencier.\medskip\vspace{4px}

Nous allons attribuer à chaque ontologie un numéro ; celle qui sera parsée la première aura le numéro 1, la suivante aura le numéro 2, et ainsi de suite. L’ID sera le numéro de l’ontologie originale d’où venait l’entité en question. Par conséquent, nous pourrons garder intacte la partie "nom" des entités redondantes, sans être obligés de la modifier. C’est seulement la partie "IRI de préfixe" qui va changer. Nous avons défini l’ID sur quatre caractères, ainsi les quatre derniers caractères du préfixe de chaque entité seront réservés à l’ID.\medskip\vspace{4px}

\begin{itemize}
\item[$\bullet$] Voici comment vont paraître les IRIs complets des entités de nom "\textbf{Paper}" dans notre ontologie de sortie :
\begin{itemize}
\item \underline{http://intégration\textbf{/001}\#Paper}
\item \underline{http://intégration\textbf{/002}\#Paper}
\item \underline{http://intégration\textbf{/003}\#Paper}
\end{itemize}\medskip\vspace{4px}

\item[$\bullet$] Voici comment vont paraître les IRIs complets des entités de nom "\textbf{Person}" dans notre ontologie de sortie :
\begin{itemize}
\item \underline{http://intégration\textbf{/001}\#Person}
\item \underline{http://intégration\textbf{/002}\#Person}
\item \underline{http://intégration\textbf{/003}\#Person}
\end{itemize}\medskip\vspace{4px}

\item[$\bullet$] Voici comment vont paraître les IRIs complets des entités de nom "\textbf{Conference}" dans notre ontologie de sortie :
\begin{itemize}
\item \underline{http://intégration\textbf{/001}\#Conference}
\item \underline{http://intégration\textbf{/002}\#Conference}
\item \underline{http://intégration\textbf{/003}\#Conference}\\
\end{itemize}
\end{itemize}

De cette manière, le "nom" redondant d’une entité quelconque sera préservé, aura un IRI unique dans la nouvelle ontologie, et toutes ses informations liées (sa définition dans son ontologie originale) seront conservées correctement.

\subsection{Première étape}
\subsubsection{En général}
Nous faisons le parsing des \textbf{classes} (des ontologies d’entrée) et de leurs définitions (descriptions), et au fur et à mesure, nous créons les axiomes correspondants à ces classes et à leurs définitions dans notre nouvelle ontologie.\medskip\vspace{4px}

Nous faisons le parsing des \textbf{propriétés d’objet} (des ontologies d’entrée) et de leurs définitions, et au fur et à mesure, nous créons les axiomes correspondants à ces propriétés d’objet et à leurs définitions dans notre nouvelle ontologie.\medskip\vspace{4px}

Nous faisons le parsing des \textbf{propriétés de données} (des ontologies d’entrée) et de leurs définitions, et au fur et à mesure, nous créons les axiomes correspondants à ces propriétés de données et à leurs définitions dans notre nouvelle ontologie.\medskip\vspace{4px}

Nous faisons le parsing des \textbf{propriétés d’annotation} (des ontologies d’entrée) et de leurs définitions, et au fur et à mesure, nous créons les axiomes correspondants à ces propriétés d’annotation et à leurs définitions dans notre nouvelle ontologie.\medskip\vspace{4px}

Nous faisons le parsing des \textbf{individus / instances} (des ontologies d’entrée) et de leurs définitions, et au fur et à mesure, nous créons les axiomes correspondants à ces individus et à leurs définitions dans notre nouvelle ontologie.\medskip\vspace{4px}

Et, nous faisons le parsing des \textbf{individus anonymes} (des ontologies d’entrée) et de leurs définitions, et au fur et à mesure, nous créons les axiomes correspondants à ces individus et à leurs définitions dans notre nouvelle ontologie.

\subsubsection{En détail}
Pendant le parsing des\textbf{ classes} des ontologies d’entrée, nous extrayons pour chaque classe sa définition qui consiste en ses annotations utilisées (ses labels, ses commentaires, ses propriétés d’annotation), ses superclasses, ses classes équivalentes, et disjointes (informations avec lesquelles nous créons les classes de notre future ontologie). Et nous remplissons au fur et à mesure le HMap des classes qui contiendra comme "clé" l’URI original de la classe, et comme "valeur" le numéro de l’ontologie dont il est issu.\medskip\vspace{4px}

Pendant le parsing des \textbf{propriétés d’objet/data} des ontologies d’entrée, nous extrayons pour chaque propriété sa définition qui consiste en ses annotations utilisées (ses labels, ses commentaires, et ses propriétés d’annotation), ses super-propriétés, ses domaines, ses images, ses propriétés inverses (pour les propriétés d’objet seulement), équivalentes, disjointes, et son type (informations avec lesquelles nous créons les propriétés de notre future ontologie). Et nous remplissons au fur et à mesure les deux HMaps (des propriétés d’objet, et des propriétés data) qui contiendront comme "clé" l’URI original de la propriété, et comme "valeur" le numéro de l’ontologie dont il est issu.\medskip\vspace{4px}

Pendant le parsing des \textbf{instances} des ontologies d’entrée, nous extrayons pour chaque instance sa définition qui consiste en ses annotations utilisées (ses labels, ses commentaires, et ses propriétés d’annotation), ses classes qui l’instancient, ses instances identiques et différentes, les propriétés d’objet/data qu’elle appelle et leurs valeurs (informations avec lesquelles nous créons les instances de notre future ontologie). Et nous remplissons au fur et à mesure le HMap des instances qui contiendra comme "clé" l’URI original de l’instance, et comme "valeur" le numéro de l’ontologie dont il est issu.\medskip\vspace{4px}

Pendant le parsing des \textbf{propriétés d’annotation} des ontologies d’entrée, nous extrayons pour chaque propriété sa définition qui consiste en ses labels, ses commentaires, ses super-propriétés, ses domaines, et ses images (informations avec lesquelles nous créons les propriétés d’annotation de notre future ontologie).\medskip\vspace{4px}

Pendant le parsing des \textbf{individus anonymes} des ontologies d’entrée, nous extrayons pour chaque individu (qui est sous forme d’un ID local unique) sa définition qui consiste en ses labels et ses commentaires, \textit{etc.} (informations avec lesquelles nous créons les individus anonymes de notre future ontologie).\\

\ding{220}  Ainsi, lors du parsing des ontologies originales, nous avons créé des entités propres à notre future ontologie, et en même temps, nous avons rempli les quatre HMaps correspondants à chaque type d’entité –\textbf{classes}, \textbf{propriétés objet}, \textbf{propriété de données}, et \textbf{instances}– (où l’URI original de l’entité forme la "clé", et le numéro de son ontologie forme la "valeur"), et tout cela pour remédier au problème de redondance des entités expliqué dans la section précédente.

\subsection{Deuxième étape}
Nous allons représenter les correspondances (entre les différentes entités des ontologies sources) par des axiomes OWL, car cette représentation est sémantiquement correcte et permet de réutiliser l’infrastructure et le vocabulaire du langage OWL.

\subsubsection{En général}
Nous parcourons les correspondances (les paires d’entités) de chaque alignement d’entrée (\textit{i.e.} les cellules qui ont une mesure supérieure ou égale à un seuil que l'utilisateur a fixé), et au fur et à mesure, nous ajoutons à la nouvelle ontologie des axiomes de pont (des liens d’équivalence) traduisant ces correspondances entre les entités déjà créées dans notre ontologie.

\subsubsection{En détail}
Dans OWL API, nous ne pouvons pas lier les entités directement par des axiomes. En effet, il existe quatre types de méthodes de création d’axiomes, chacun destiné à un type d’entité (–\textbf{classes}, \textbf{propriétés d'objet}, \textbf{propriétés de données}, et \textbf{instances}–). Ceci s’applique entre autres pour les axiomes d’équivalence que nous allons utiliser dans notre cas. Les voici :

\begin{figure}[!ht]
\begin{center}
\includegraphics[width=\textwidth]{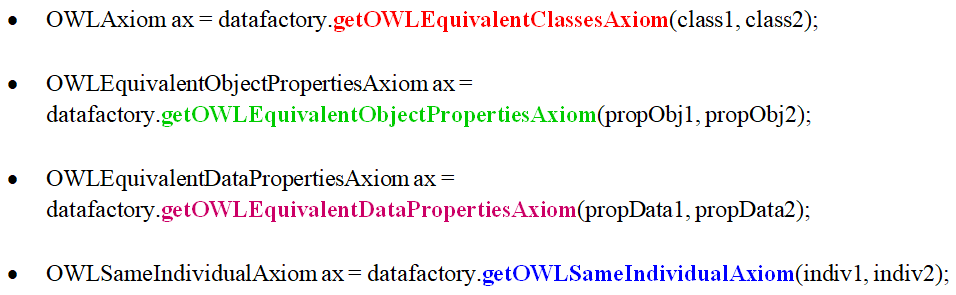}
\caption[Axiomes d'équivalence dans OWL API]{Création des axiomes d'équivalence dans OWL API}
\end{center}
\end{figure}

Sachant que dans les alignements, nous ne pouvons pas savoir le type des entités mises en correspondance, et que les entités sont citées par leurs URIs originaux complets (comme elles ont été définies dans leur ontologie originale), nous avons eu recours aux quatre HMaps remplis déjà dans l’étape précédente, pour savoir le type de chaque entité, et son ID (le numéro de l’ontologie dont elle est issue), afin de pouvoir créer des "bridging" axiomes en liant les couples d’entités (citées dans les cellules des alignements) par des axiomes d’équivalence, en changeant leurs URIs de préfixe originaux par l’URI de notre ontologie + l’ID correspondant.\medskip\vspace{4px}

A la fin, dans notre ontologie de sortie, les supposés "bridging" axiomes créés ne vont plus être considérés comme ceci. En effet, selon nous, ils vont plutôt être perçus comme des axiomes d’équivalence normaux et originaux liant des entités d’une nouvelle ontologie indépendante (la nôtre), comme si celle-ci n’était pas le résultat d’une intégration ; car les entités la composant ont toutes un "URI de préfixe" propre à elle (les URIs des entités de notre nouvelle ontologie ne sont pas des URIs des ontologies sources déjà publiées). Autrement dit, nous avons fait un refactoring (une personnalisation) des URIs des entités sources que nous avons réutilisées pour former notre ontologie.

\section{Démarche détaillée de la référence}

Comme il n’y a pas de benchmark pour les approches d’intégration des ontologies, nous avons créé une autre version d’intégration avec laquelle nous pourrons nous comparer.
Nous considérerons l'ontologie résultante comme notre ontologie de référence, car il s'agit d'une ontologie de pont sans perte de connaissances (\textit{i.e.}, complète).
Nous appellerons cette version comme l'intégration de "référence" ou la "pseudo-référence".\medskip

En principe, la démarche de référence consiste en la composition (l’union / l’intégration) automatique des ontologies sources et l’ontologie qui correspond aux alignements entre elles. Supposons que nous avons deux ontologies $O_1$ et $O_2$ à intégrer, et un alignement $A$ entre elles. L’ontologie résultante serait $O_3 = O_1 + O_2 + O_A$, après avoir converti $A$ en une ontologie. En effet, puisque le format d’alignement est exprimé en RDF, ainsi il est librement extensible, l’"Alignment API" permet de convertir les correspondances (les cellules) d’un alignement en des axiomes OWL d’équivalence, de subsomption, et de disjonction (à l’aide de la méthode OWLAxiomsRendererVisitor(ObjectAlignment)) pour générer une ontologie comprenant à la fois les entités alignées et les axiomes OWL de pont. Malheureusement, cette tâche n’a pas pu être effectuée correctement, et nous n’avons pas pu transformer directement l’alignement en une ontologie. Pour ce faire, nous avons appliqué l’idée de notre approche (déjà expliquée). 

\subsection{Première étape}
OWLOntologyMerger() est une méthode prédéfinie dans OWL API qui unit toutes les ontologies chargées (loaded) dans le OWLOntologyManager. Nous n’avons qu’à spécifier à la méthode createMergedOntology() le nouvel URI de l’ontologie que nous allons créer :\medskip\vspace{13px}

\begin{center}
OWLOntologyMerger integration = new \textbf{OWLOntologyMerger}(manager);\\
OWLOntology newOnto = integration.\textbf{createMergedOntology}(manager, newIRI);
\end{center}

\medskip\vspace{13px}
Il faut noter que les termes "Merger" et "MergedOntology" utilisés par OWL API sont faux et accentuent encore plus la mécompréhension du terme "fusion" dans la communauté. En effet, c’est une composition (union, intégration, association), ce n’est pas une fusion. D’ailleurs, Protégé fait exactement la même erreur avec le terme "Merge ontologies" dans le menu "refactor".\medskip\vspace{4px}

Nous aurons avec seulement cette étape une ontologie intégrée (composée, agrégé) qui ne manque aucune information des ontologies d’entrée, et qui maintient les URIs d’origine des entités sources en les mettant dans l’ontologie de sortie telles qu’elles sont originairement dans leurs ontologies d’entrée.

\subsection{Deuxième étape}
Nous parsons les \textbf{classes} des ontologies d’entrée, et nous remplissons au fur et à mesure le HSet des classes par tous les URIs originaux des classes.\medskip\vspace{4px}

Nous parsons les \textbf{propriétés d’objet} des ontologies d’entrée, et nous remplissons au fur et à mesure le HSet des propriétés d’objet par tous les URIs originaux des propriétés d’objet.\medskip\vspace{4px}

Nous parsons les \textbf{propriétés de données} des ontologies d’entrée, et nous remplissons au fur et à mesure le HSet des propriétés de données par tous les URIs originaux des propriétés de données.\medskip\vspace{4px}

Nous parsons les \textbf{individus} des ontologies d’entrée, et nous remplissons au fur et à mesure le HSet des individus par tous les URIs originaux des individus.\medskip\vspace{4px}

Nous parcourons les correspondances de chaque alignement d’entrée (\textit{i.e.} les cellules qui ont une mesure supérieure ou égale au seuil que nous avons fixé), et au fur et à mesure, nous ajoutons à la nouvelle ontologie des liens d’équivalence (des "bridging" axiomes) qui traduisent fidèlement ces correspondances entre les entités.\medskip\vspace{4px}

\ding{220} Sachant que dans OWL API, nous ne pouvons pas lier les entités directement par des axiomes, mais plutôt à travers quatre types de méthodes de création d’axiomes pour chacun des types d’entité (–\textbf{classes}, \textbf{propriétés objet}, \textbf{propriétés de données}, et \textbf{instances}–), et sachant que dans les alignements, nous ne pouvons pas savoir le type des entités correspondues, nous avons pu, à partir des quatre HSets déjà créés et remplis, savoir le type de chaque entité des cellules parcourues et créer des "bridging" axiomes (lier les entités citées dans les cellules des alignements par des axiomes d’équivalence en gardant leurs URIs originaux (tels qu’ils sont cités dans les alignements)).

\section{Comparaison entre OIA2R et la référence}
L’intégration de référence n’a absolument aucune perte d’informations des ontologies originales, pas le moindre axiome perdu, par contre la nôtre ne parvient pas à tout parser, nous perdons les entités anonymes et les restrictions, car toutes les entités que nous parsons sont nommées.\medskip\vspace{4px}

La seule différence entre ce travail et la référence, c’est que toutes les entités de notre ontologie résultante ont un URI de préfixe propre à nous, contrairement à l’intégration de référence qui garde les URIs originaux des entités et qui ne fait aucun refactoring. Ainsi notre ontologie est tout à fait originale et ne pointe pas sur des entités appartenant à des ontologies externes déjà existantes.

\section{Conditions favorables pour de meilleurs résultats}

\subsection[Mapping au lieu d’alignement]{Mapping (1-à-1) au lieu d’alignement (1-à-N)}
Comme expliqué dans le chapitre 1, le matching retourne un alignement ou un mapping. Dans l’alignement, une entité d’une première ontologie peut être mise en correspondance avec une ou plusieurs entités d’une deuxième ontologie (ces correspondances n’ont pas la même pertinence). Il s’agit de correspondances "1-à-N". Tandis que dans le mapping, une entité d’une première ontologie peut être mise en correspondance avec zéro ou une seule entité d’une deuxième ontologie. Il s’agit de correspondances "1-à-1".\medskip\vspace{4px}

D’après les expérimentations (que nous détaillerons dans le chapitre 4), nous avons remarqué que, dans l’ontologie résultant d’une intégration qui utilise des alignements, le nombre de classes insatisfiables est beaucoup plus important que celui de l’ontologie résultant d’une intégration qui utilise des mappings.\medskip


Ci-dessous un exemple qui montre comment se forme l’insatisfiabilité d’une classe dans une ontologie de pont. La figure \ref{just} montre les justifications que nous a affichées le debugueur de OWL API (à l’aide du raisonneur HermiT) pour une des classes insatisfiables (002\#Tissue\_Dissection) générées dans une de nos expérimentations.\medskip

\begin{figure}[!ht]
\centering
\includegraphics[width=1.005\textwidth]{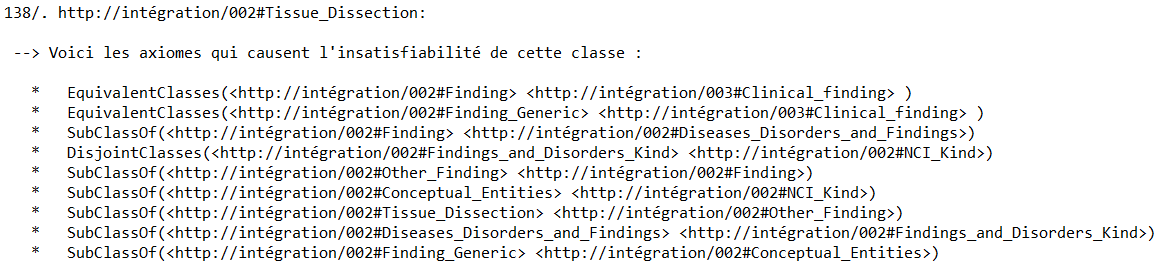}
\caption[Debugage d'une classe insatisfiable 1]{Debugage d'une classe insatisfiable dans une ontologie de pont}
\label{deb}
\end{figure}

La figure \ref{just} est la représentation graphique de axiomes de justification générés par le debugueur, où les classes de couleur rouge sont les classes insatisfiables.\medskip

\begin{figure}[!ht]
\centering
\includegraphics[scale=0.755]{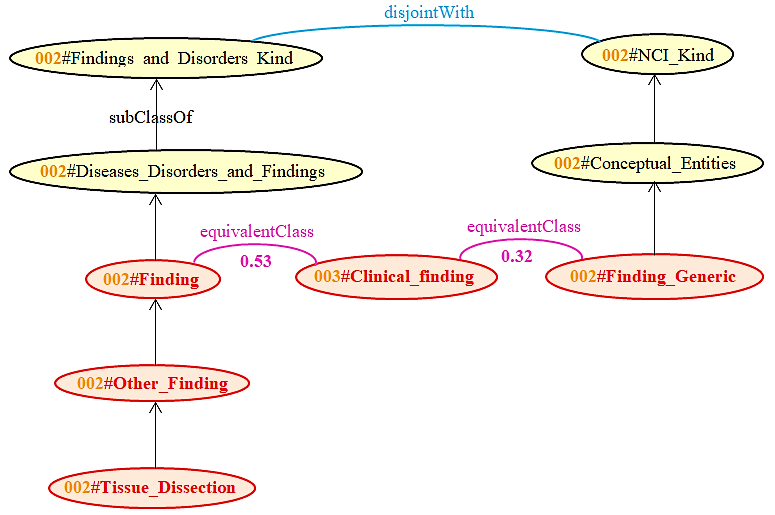}
\caption[Formation des classes insatisfiables 1]{Formation des classes insatisfiables}
\label{just}
\end{figure}

\begin{figure}[!ht]
\centering
\includegraphics[width=1.004\textwidth]{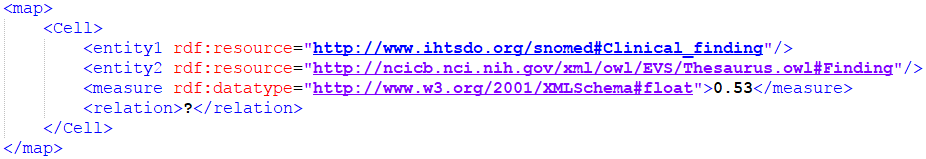}
\includegraphics[width=1.004\textwidth]{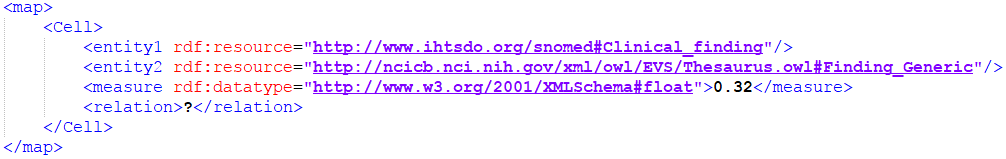}
\caption[Correspondances ayant la même source]{Deux correspondances ayant la même entité source}
\label{cores}
\end{figure}

La figure \ref{cores} montre les deux correspondances qui ont causées toutes  ces incohérences. La relation "?" veut dire une relation d'équivalence "=" correcte mais qui génère des insatisfiabilités dans l'ontologie intégrée.\medskip

\begin{figure}[!ht]
\centering
\includegraphics[scale=0.8]{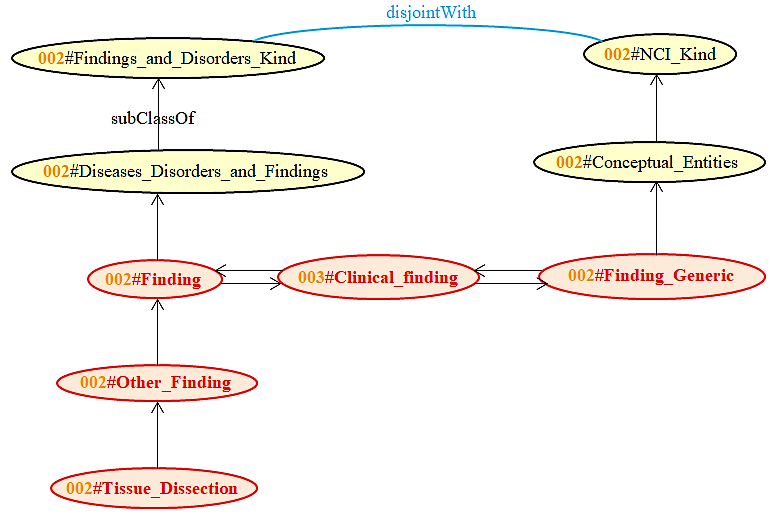}
\caption[Cause de l'incohérence 1]{Cause de l'incohérence d'une ontologie de pont}
\label{cauz}
\end{figure}

Sachant que la relation d’équivalence (des "bridging" axiomes) est en réalité égale à deux relations de subsomption dans les deux sens, le schéma devient comme le montre la figure \ref{cauz}.\medskip



\sloppy{
Nous remarquons qu’après l’ajout des "bridging" axiomes d’équivalence, la classe "002\#Tissue\_Dissection" devient par inférence une sous-classe des deux classes "002\#Findings\_and\_Disorders\_Kind" et "002\#NCI\_Kind" qui sont disjointes (information extraite de l’ontologie originale (Ont\textbf{2})). Ceci est contradictoire, car une classe ne peut pas être une sous-classe de deux classes disjointes. Aucune instance ne peut la satisfaire. La même chose s’applique pour les autres classes coloriées en rouge.} Si la classe "003\#Clinical\_finding" (provenant de l'ontologie Ont\textbf{3}) avait été correspondue avec une seule classe de l’autre ontologie (00\textbf{2}), précisément la classe avec laquelle elle a la plus grande mesure de similarité, nous aurons évité toutes ces incohérences.\medskip\vspace{4px}

Le pire avec les alignements, c’est que les entités cibles (ou sources) correspondues avec la même entité source (ou cible) sont généralement toutes proches (\textit{i.e.}, voisines) ainsi susceptibles d’avoir entre elles des relations de disjonction qui sont la première source des incohérences.\medskip\vspace{4px}

\begin{figure}[!ht]
\centering
\includegraphics[scale=0.8]{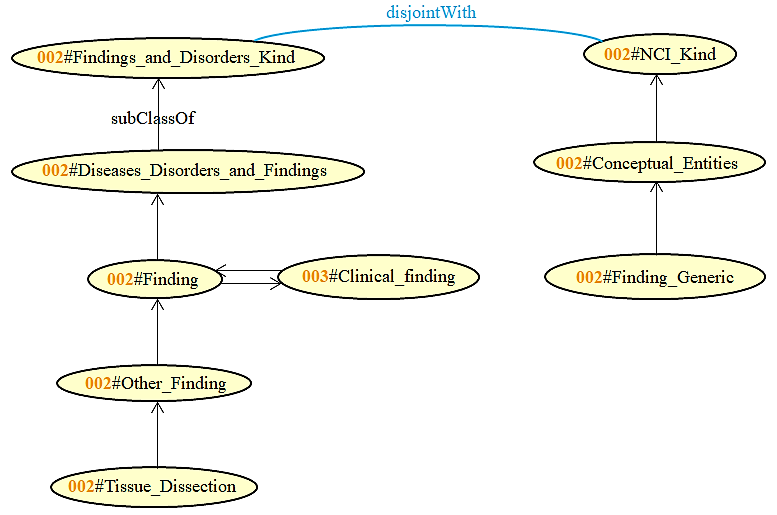}
\caption{Résolution des insatisfiabilités}
\label{rez}
\end{figure}

La figure \ref{rez} montre ce que devient si nous ne gardons qu’une seule correspondance pour l’entité source "003\#Clinical\_finding".\medskip

\begin{figure}[!ht]
\centering
\subfloat[][Étape 1]{\includegraphics[width=0.49\linewidth]{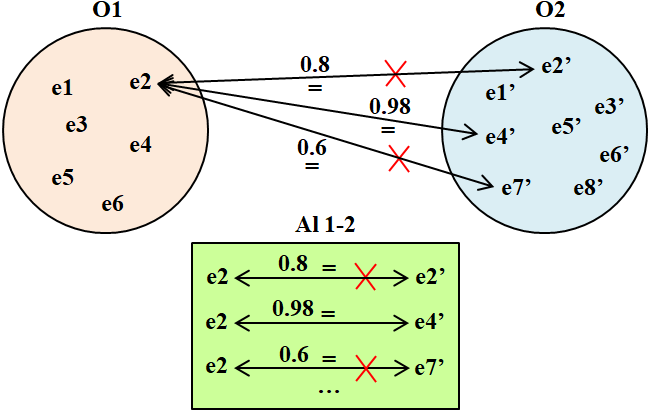}}~~~~
\subfloat[][Étape 2]{\includegraphics[width=0.49\linewidth]{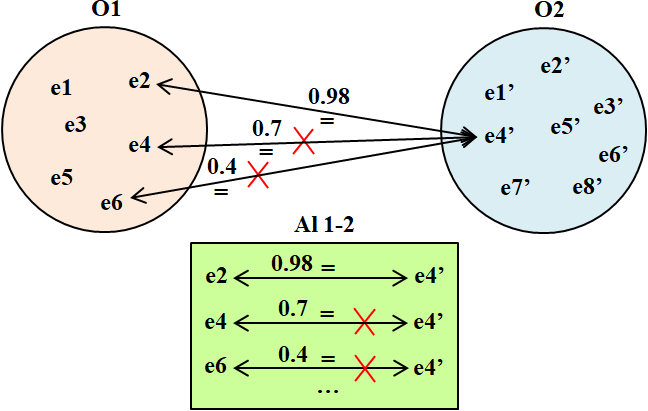}}
\caption[Transformation d'un alignement en un mapping]{Transformation d'un alignement (1-à-N) en un mapping (1-à-1)}
\label{mapppp}
\end{figure}

Dans notre approche, nous avons la possibilité de filtrer les correspondances ayant la même entité source ou la même entité cible en gardant uniquement la correspondance ayant la plus grande confiance (\textit{c.f.}, figure \ref{mapppp}).\medskip\vspace{4px}

En premier lieu, nous créons deux HMaps : la première contiendra comme clé l’IRI de l’entité source de chaque cellule d’un alignement, et comme valeur l’IRI de l’entité cible. La deuxième contiendra comme clé l’IRI de l’entité source de chaque cellule d’un alignement, et comme valeur la mesure de confiance de la relation. Pendant le parsing des alignements d’entrée, nous remplissons ces deux HMaps au fur et à mesure, de telle sorte que si nous trouvons une cellule dont l’entité source est déjà existante comme clé dans les HMaps et dont la mesure de confiance est supérieure à celle rencontrée avant, nous mettons à jour les valeurs correspondantes à cette clé dans les deux HMaps ; sinon nous ne faisons rien (\textit{i.e.} si la cellule dont l’entité source est déjà existante comme clé dans les HMaps et dont la mesure de confiance est inférieure ou égale à celle rencontrée avant, elle ne sera pas stockée, car nous avons celle qui est plus pertinente déjà enregistrée dans les HMaps).\medskip\vspace{4px}

En deuxième lieu, nous allons refaire la même chose, mais à l'envers. Nous créons deux HMaps "inverses" : la première contiendra comme clé l’IRI de l’entité cible de chaque entrée de la première HMap issue de l'étape précédente, et comme valeur l’IRI de l’entité source. La deuxième contiendra comme clé l’IRI de l’entité cible de chaque entrée de la première HMap issue de l'étape précédente, et comme valeur la mesure de confiance de la relation. Pendant le parcours des entrées de la première HMap déjà remplie dans l'étape 1, nous remplissons ces deux HMaps "inverses" au fur et à mesure, de telle sorte que si nous trouvons une entrée dont l’entité cible est déjà existante comme clé dans les HMaps "inverses" et dont la mesure de confiance est supérieure à celle rencontrée avant, nous mettons à jour les valeurs correspondantes à cette clé dans les deux HMaps "inverses".\medskip

Au lieu d'utiliser les alignements originaux (\textit{i.e.}, d'entrée), nous utiliserons le premier HMap "inverse" qui contiendra toutes les correspondances filtrées des alignements d'entrée. Ce HMap exprime les correspondances supposées former des mappings.

\subsection{Réparation des alignements}
La réparation des alignements ou des mappings (en supprimant les correspondances qui sont susceptibles d’engendrer des insatisfibilités lorsqu’elles seront associées aux ontologies d’entrée) aide à diminuer les incohérences dans l’ontologie résultante.\medskip\vspace{4px}

D’après \cite{cheatham2017semantic}, actuellement, peu de systèmes de matching d'ontologies ne supportent la gestion de l’incohérence logique, et encore moins pour les grandes ontologies. L'approche la plus basique consiste à filtrer les correspondances qui violent une série de règles sémantiques (comme le fait YAM++ (2012)). Des approches plus sophistiquées reposent sur des procédures automatisées capables d'identifier les correspondances impliquées dans l'incohérence logique et de sélectionner celles à supprimer pour atteindre la cohérence, comme AML (2015) et LogMap (2011).\medskip\vspace{4px}

\cite{abbas2013creating} remarquent deux perceptions de réparation et de debugage dans les travaux de fusion ou d'intégration des ontologies :\vspace{4px}

\begin{itemize}
\item[$\bullet$] Quelques auteurs considèrent que les ontologies (à intégrer ou fusionner) sont correctes et toujours plus fiables que les alignements, et s'il y a d'incohérence ou d'inconsistance, c’est forcément à cause des alignements (LogMap et ALCOMO). Ils cherchent alors à trouver l'ensemble minimal de conflits entraînant l’incohérence de l’alignement, et suppriment les correspondances qui causent les insatisfiabilités dans les classes de l’ontologie résultante pour minimiser leur impact.\vspace{4px}
\item[$\bullet$] D’autres auteurs trouvent que les incohérences peuvent être causées soit par les alignements, soit par les ontologies (ContentMap de \cite{jimenez2009ontology}). Ils ont déduit que lorsque les correspondances sont celles qui sont prévues (sont correctes) et lorsque l’ontologie résultante contient quand même des incohérences logiques, alors ces incohérences doivent être forcément dues aux ontologies sources qui sont incompatibles à cause des différences de leurs conceptualisations. Ils proposent alors une solution pour réparer les ontologies (supprimer les axiomes qui causent des contradictions dans l'ontologie de sortie) à l'aide d’un ingénieur du domaine.
\end{itemize}\medskip\vspace{4px}

Dans notre approche, nous avons exploité les outils LogMap et ALCOMO qui prennent en entrée deux ontologies sources et un alignement entre eux, et génèrent un alignement réparé (après avoir fait un calcul des inférences entre eux pour décider quelles correspondances supprimer). Nous utilisons l’un de ces deux outils pour tous nos alignements d’entrée afin de minimiser au maximum les incohérences dans notre future ontologie résultante.\medskip

LogMap\footnote{\url{https://github.com/ernestojimenezruiz/logmap-matcher}} est un système de matching d'ontologies basé sur la logique créé par \cite{jimenez2011logmap}. Il effectue une réparation des alignements (\textit{i.e.}, une transformation d'un alignement incohérent en un alignement cohérent) en exécutant un raisonnement (parfois incomplet). Il supprime ou modifie les correspondances qui causent l'apparition des classes insatisfiables.\medskip\vspace{4px}

Créé par \cite{meilicke2011alignment}, ALCOMO\footnote{\url{http://web.informatik.uni-mannheim.de/alcomo/}} est un système de debugage des alignements qui permet de transformer un alignement incohérent en un alignement cohérent en lui supprimant certaines correspondances. Il est complet car il détecte toute forme d'insatisfiabilité entre les ontologies causée par les alignements.

\section{Nouvelle définition de la notion d'\textit{intégration}}
Dans la littérature, il n’y pas un accord général sur les définitions de l’intégration et la fusion des ontologies. \cite{flouris2006classification}, et \cite{euzenat2007ontology} ont essayé de faire une clarification et une désambiguïsation de toutes les terminologies de l’ingénierie des ontologies. Et \cite{pinto1999towards} a fait la même chose pour les termes "intégration" et "fusion". Mais malgré leurs efforts, \textbf{les termes "intégration" et "fusion" sont toujours mal définis, mal compris, et mal placés}.\medskip\vspace{4px}

Comme le dit \cite{pinto1999towards}, l’intégration concerne des ontologies de différents ou de proches domaines pour former une ontologie de domaine plus large englobant tous les domaines sources ; et la fusion concerne des ontologies de même domaine pour former une ontologie décrivant mieux ce domaine-là. La confusion réelle réside dans le sens naturel de ces termes. En effet, dans la littérature, la plupart des auteurs parlent de fusion lorsqu’ils vont fondre les entités équivalentes pour les remplacer par une seule, ou lorsqu’ils vont changer et mêler les hiérarchies des ontologies sources en répartissant leurs entités autrement dans l’ontologie cible ; et parlent d’intégration ou de composition lorsqu’ils vont regrouper et assembler directement les ontologies, telles qu'elles sont, dans une autre ontologie, sans fusionner leurs entités équivalentes et sans toucher leurs hiérarchie originale. Les processus de fusion ou d'intégration peuvent s'appliquer concrètement à toute ontologie (de domaine $\neq$ ou $=$). C'est seulement le but à atteindre qui différencie vraiment les deux définitions consensuelles. Autrement dit, le problème réside dans le fait que nous pourrons faire une fusion (au sens propre du mot) pour deux ontologies de domaines différents (car elles peuvent contenir quand même des chevauchements entre elles), et une composition / intégration (au sens propre du mot) pour des ontologies de même domaine. Ce qui est contradictoire aux définitions soi-disant standardisées.\medskip\vspace{4px}

Dans le cas d'une ontologie de pont créée à partir d'ontologies de même domaine, si nous nous conformons à ces définitions, nous sommes en train de faire une fusion des ontologies, car il s’agit bien d’ontologies de même domaine, mais réellement, nous ne fusionnons pas les entités, ne mélangeons pas leurs emplacements, et ne modifions pas leurs structures, nous faisons une ontologie de pont où les entités originales et leurs hiérarchies sont intactes. Pas de fusion dans un processus de fusion. En contre parti, notre travail respecte les règles de la définition de l’intégration des ontologies qui dicte que les parties provenant des ontologies sources soient identifiables facilement dans l’ontologie de sortie et qu’il s’agit juste d’une inclusion, d'une agrégation, ou d'un assemblage des ontologies sources pour former un tout. Avions-nous fait une intégration ou une fusion ?\medskip\vspace{4px}

Par conséquent, nous proposons que le terme "intégration des ontologies" soit le terme général de toutes les définitions rencontrées dans le chapitre 2, par analogie avec le terme "intégration des données", et nous proposons qu’il soit appliqué sur des ontologies de \textbf{mêmes ou de différents sujets}, ainsi \textbf{pour de différents objectifs}. Bien évidemment, les ontologies de même sujet seront les plus dures à intégrer ; les ontologies de très différents domaines n’auront pas (beaucoup) de correspondances entre elles, donc seront toujours plus faciles à intégrer. Nous distinguons \textbf{cinq niveaux d’intégration selon le niveau d’interopérabilité sémantique}. Nous expliquons chaque type avec le plus simple cas qui intègre seulement deux ontologies (\textit{c.f.}, figure \ref{new}):\bigskip

\begin{figure}[!ht]
\centering
\includegraphics[width=1.001\textwidth]{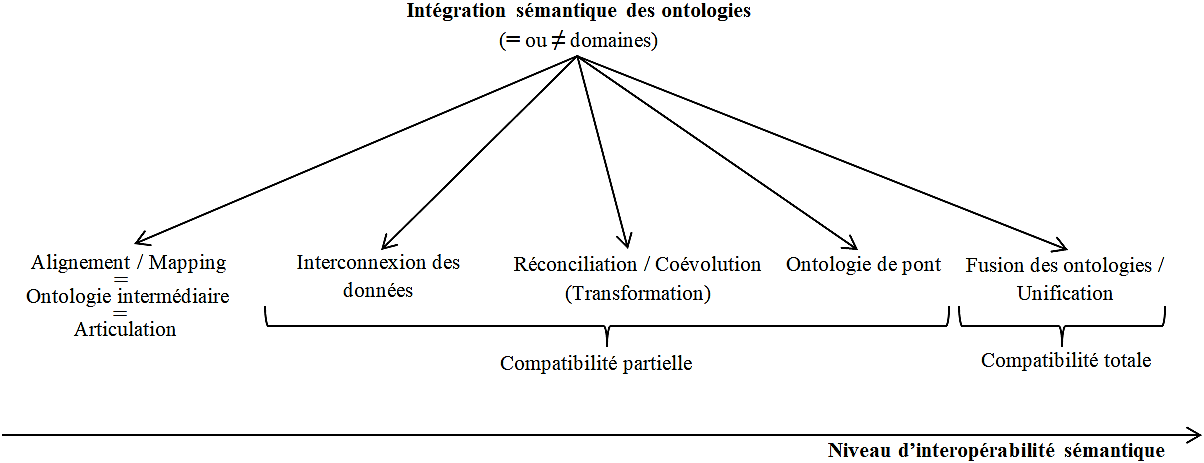}
\caption{Nouvelle définition de l'intégration sémantique}
\label{new}
\end{figure}

\paragraph{Alignement}
Défini par \cite{noy1999algorithm} (définition 1), il implique deux ontologies séparées et un alignement (deux mapping dans les deux sens, une articulation, ou une ontologie intermédiaire comme le nomme \cite{kalfoglou2003ontology}) à travers lequel une ontologie peut interroger l’autre et vice-versa.

\paragraph{Interconnexion des données}
Nommée aussi "révisons de mapping" par \cite{heflin2000dynamic}, terme qui peut être confondu avec la notion de debugage et de réparation des mappings, elle consiste en l’ajout de correspondances d’un alignement ou d’un mapping dans les deux ontologies. En d'autres termes, c'est l'ajout des relations sémantiques entre les entités de l'ontologie en question et les entités de l'autre ontologie comme prescrit dans l'alignement ou le mapping entre elles.

\paragraph{Réconciliation / Coévolution}
Définie par \cite{euzenat2007ontology}, elle peut être une transformation des entités de l’une des ontologies par les entités de l'autre comme le prescrit l’alignement entre les deux ontologies, ou bien une transformation des entités des deux ontologies, nommée "intersection d’ontologies" par \cite{heflin2000dynamic}, après la standardisation des termes correspondus.

\paragraph{Ontologie de pont}
Introduite par \cite{de2006ontology}, elle met les deux ontologies et les correspondances de leur alignement dans une même ontologie qui les englobe sans rien modifier.

\paragraph{Fusion / Unification}
Appelée par \cite{pinto1999towards} "Fusion", et appelée "unification" ou "compatibilité totale" par \cite{sowa1997electronic}, elle génère une ontologie en sortie et peut se faire de différentes manières ; la plus simple est d’unir les ontologies sources et de fusionner leurs entités équivalentes (comme décrites dans l’alignement); et la plus difficile est d’exploiter, en plus des correspondances d’équivalence et de disjonction, des correspondances de subsomption qui changeront énormément la hiérarchie originale des ontologies.\medskip

Selon \cite{calvanese2001framework}, dans le cas de plusieurs ontologies sources, l'intégration des ontologies peut impliquer soit une approche "\textit{global-centric}", où les entités de l'ontologie globale sont mises en correspondance avec les entités des ontologies locales, soit une approche "\textit{local-centric}", où les entités des ontologies locales sont mises en correspondance avec les entités de l'ontologie globale. Mais nous ajoutons une autre approche que nous nous permettons d'appeler "\textit{non-centric}" où les entités des paires d'ontologies sont mises en correspondance. Ainsi, il n'existe pas d'ontologies globale et locales, elles ont toutes la même priorité.\medskip\vspace{4px}

\begin{figure}[!ht]
\centering
\includegraphics[width=1.004\textwidth]{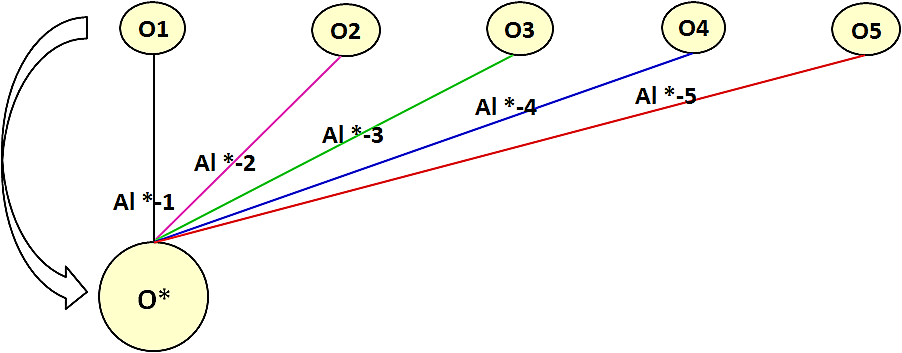}
\caption{Intégration incrémentale (cas 1)}
\label{noninc}
\end{figure}

\begin{figure}[!ht]
\centering
\includegraphics[width=1.004\textwidth]{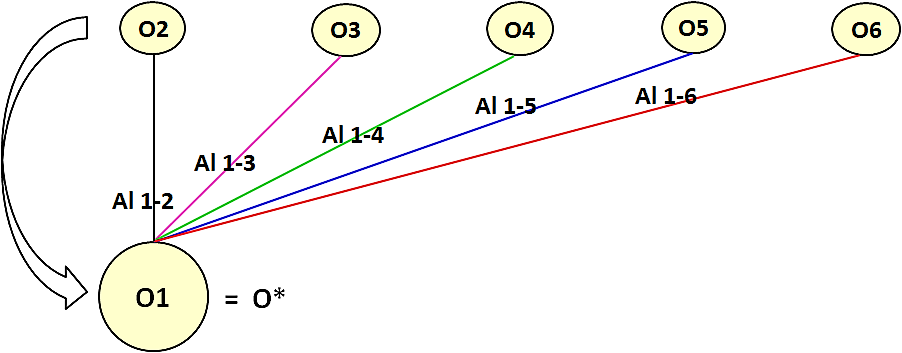}
\caption{Intégration incrémentale (cas 2)}
\label{noninc1}
\end{figure}

Concernant l'ontologie de pont et l'ontologie de fusion, le processus d'intégration de plusieurs ontologies peut être fait soit d'une manière \textit{incrémentale}, où il y a une ontologie cible prioritaire ou une ontologie vide (nommée \textit{ontologie globale}) dans laquelle les autres ontologies sources (\textit{locales}) seront intégrées l'une après l'autre (\textit{c.f.}, figure \ref{noninc} and \ref{noninc1}); soit d'une manière \textit{non incrémentale}, où toutes les ontologies sources ont la même priorité et seront intégrées les unes avec les autres en même temps pour former ensemble l'ontologie cible.

\begin{figure}[!ht]
\centering
\includegraphics[width=\linewidth]{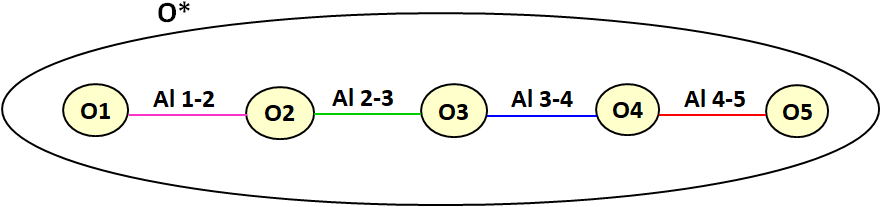}
\caption{Intégration non incrémentale (2-to-2)}
\label{22}
\end{figure}

\begin{figure}[!ht]
\centering\includegraphics[width=\linewidth]{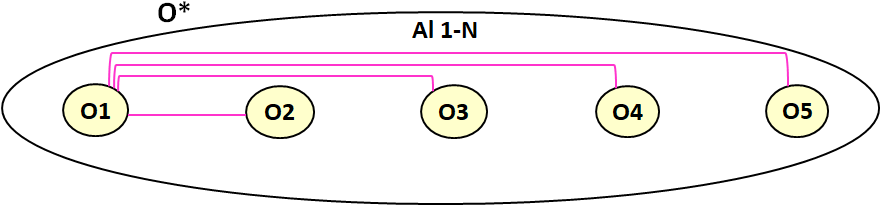}
\caption{Intégration non incrémentale (1-to-N)}
\label{1N}
\end{figure}

\begin{figure}[!ht]
\centering
\includegraphics[width=\linewidth]{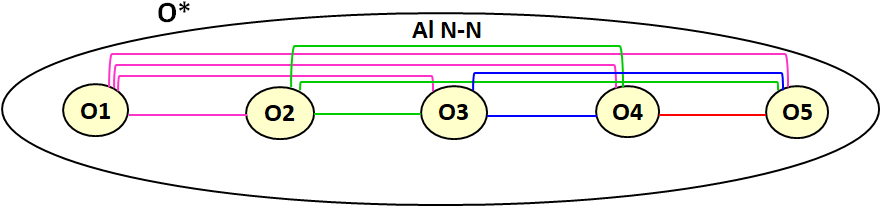}
\caption{Intégration non incrémentale (N-to-N)}
\label{NN}
\end{figure}

\newpage
Nous proposons trois types d'intégration non incrémentale :\medskip

\begin{itemize}
\item[$\bullet$] Intégration 2-à-2: Les ontologies sont intégrées uniquement à l'aide des alignements entre les paires d’ontologies consécutives (\textit{c.f.}, figure \ref{22}),
\item[$\bullet$] Intégration 1-à-N: Les ontologies sont intégrées à l'aide des alignements entre une ontologie choisie (préférée ou prioritaire) et les autres ontologies (\textit{c.f.}, figure \ref{1N}),
\item[$\bullet$] Intégration N-à-N: Les ontologies sont intégrées à l'aide des alignements entre toute paire d’ontologies possible (\textit{c.f.}, figure \ref{NN}).
\end{itemize}
\medskip

L'intégration N-à-N est la plus complète, par contre les deux autres types d'intégration (1-à-N et 2-à-2) ne le sont pas (\textit{i.e.}, n'assurent pas une interopérabilité complète). Tenons l'exemple de cinq ontologies appartenant au même domaine ($O_1$, $O_2$, $O_3$, $O_4$, and $O_5$) pour faire une intégration 2-à-2. Supposons que toutes les ontologies contiennent une classe $A$, à part l'ontologie $O_4$. Si $O_4$ était dans la quatrième position, alors la classe $A$ de l'ontologie $O_5$ ne va pas avoir une correspondance d'équivalence avec une classe de l'ontologie $O_4$. Ce qui fait qu'elle ne sera pas intégrée avec les autres classes $A$ des autres ontologies (ayant la position 1, 2 et 3). L'intégration 1-à-N  par contre résout ce problème. Tenons le même exemple en choisissant l'ontologie $O_1$ comme l'ontologie préférée, avec qui les quatre autres ontologies ($O_2$, $O_3$, $O_4$, and $O_5$) seront alignées. Dans ce cas, la classe $A$ de l'ontologie $O_1$ va être correspondue à toutes les classes $A$ existantes, y compris celle de l'ontologie $O_5$. Malheureusement, cette méthode ne va pas garantir une interopérabilité complète. Par exemple, si une classe $B$ existe seulement dans l'ontologie $O_3$ et $O_4$, ces deux classes ne vont pas être intégrées (parce que ce type d'intégration n'utilise pas l'alignement entre $O_3$ and $O_4$). Ces inconvénients sont tous palliés par l'intégration N-à-N. Mais dans le cas d'une ontologie de pont, cette dernière va générer de multiple redondances et cycles. Par exemple, parmi les cinq ontologies d'entrée, les ontologies $O_1$, $O_2$, et $O_3$ ont une entité $A$ en commun. Dans une ontologie de pont, nous aurons des liens d'équivalence entre \NoAutoSpacing{$O1$:$A$ et $O2$:$A$, $O2$:$A$ et $O3$:$A$, $O1$:$A$ et $O3$:$A$. L'équivalence entre $O1$:$A$ et $O3$:$A$} peut être déduite des deux autres équivalences. Il s'agit donc d'un lien redondant. Et sachant qu'une relation d'équivalence est formellement composée de deux relations de subsumption (chacune dans un sens), il y aura deux cycles de subsomption entre ces trois entités (un cycle dans chaque direction).

\subsection*{Remarque}
\begin{figure}[!ht]
\centering
\includegraphics[width=\linewidth]{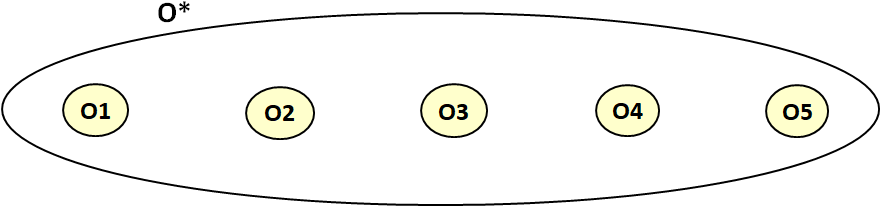}
\caption{Agrégation des ontologies}
\label{agreg}
\end{figure}

Rappelons que les termes "Merger" et "MergedOntology" utilisés par des méthodes dans OWL API,  et "Merge ontologies" dans le menu "refactor" de Protégé sont faux, car ils font juste une composition/union/agrégation d’ontologies (\textit{c.f.}, figure \ref{agreg}); ils mettent les ontologies sources, telles qu’elles sont, dans une nouvelle ontologie qui les englobe, sans créer des liens de pont en elles et sans les fusionner. Nous n’avons pas ajouté ce simple processus dans les types d’intégration, car il ne permet aucune interopérabilité sémantique entre les agrégats (les sous-ontologies sources) de l’ontologie résultante.

\section*{Conclusion}

L’ontologie de pont que nous allons implémenter est une intégration d’interopérabilité moyenne. Elle peut être exploitée réellement quand deux entreprises veulent coopérer, collaborer et intégrer leurs ontologies utilisées sans modifier les noms des entités et leurs descriptions, pour ne pas être obligés de modifier tout un système (ou une application) alimenté avec ces ontologies. Les liens de pont ajoutés dans l’ontologie de pont vont permettre l’interopérabilité sémantique entre elles.\medskip\vspace{4px}

En général, le type d’intégration choisi dépend des circonstances à faire face et des buts des applications. Tenons l’exemple des linked data dans le Web distribué et réparti, l’intégration dans ce contexte consiste à ajouter à chaque ontologie des liens d’équivalence et d’identité (sameAs) qui pointent vers des entités d’autres ontologies. C’est l’un des niveaux les plus bas d’interopérabilité, mais qui convient le mieux à cette situation.\medskip\vspace{4px}

Nous allons passer tout de suite à la concrétisation de notre approche décrite dans ce chapitre.
\chapter{Réalisation et expérimentations}

\section*{Introduction}
Dans ce chapitre, nous allons présenter notre environnement de travail, les bases de test utilisées dans les expérimentations, et les critères d’évaluation de la qualité de l’ontologie produite suite à l’intégration. Par la suite, nous allons étaler et évaluer les résultats de nos expérimentations faites sur des ontologies de différentes tailles. Ce volet d’expérimentations va nous permettre de valider notre méthode d’intégration qui s’est avérée efficace, et de prouver sa capacité de produire une ontologie de bonne qualité.

\section{Environnement de réalisation}
L’environnement de développement de notre méthode est constitué des outils suivants :\\

$\bullet$ \textbf{Java} : un langage de programmation orienté objet.\vspace{4px}

$\bullet$ \textbf{Eclipse} : un environnement de développement intégré (IDE) Java, libre, gratuit et multiplateforme.\vspace{4px}

$\bullet$ \textbf{OWL API}\footnote{\url{https://github.com/owlcs/owlapi/wiki/Documentation}} (Version 4.1.4) : une interface de programmation pour le développement, la manipulation, et la sérialisation des ontologies OWL.\vspace{4px}

$\bullet$ \textbf{HermiT}\footnote{\url{http://www.hermit-reasoner.com/}} (Version 1.3.8) : un moteur d'inférence OWL 2 DL (Description Logic) créé à l’université d’Oxford et publié sous la licence LGPL. Il est supporté par OWL API et Protégé (qui est un outil de création et de gestion d’ontologies). Il permet de réaliser les services de raisonnement suivants : l’inférence, la classification, la satisfiabilité, et la consistance. Mais il ne fournit pas un diagnostic ou une solution pour les deux derniers problèmes.\vspace{4px}

$\bullet$ \textbf{Alignment API}\footnote{\url{http://alignapi.gforge.inria.fr/}} (Version 4.9) : une interface de programmation développée en java permettant d'exprimer, d'accéder et de manipuler des alignements ontologiques sous le format d’alignement (qui est le format le plus utilisé pour représenter les alignements). Les fondateurs de cette API ont conçu le format d’alignement pour exprimer les alignements disponibles de manière uniforme et pouvoir les partager sur le Web. Ce format est écrit en langage RDF, ainsi il est librement extensible. Dans sa représentation, chaque correspondance entre deux ontologies (nommée "cellule") contient l'URI de l'entité source, l'URI de l'entité cible, la relation qui existe entre ces deux entités (égalité, subsumption, exclusion, ou instanciation \textit{etc.}), et la force de cette relation (une valeur décimale comprise entre 0 et 1, inclusivement).\\

Les tests ont été effectués sur un PC doté d’un système d’exploitation Windows 10, d’une mémoire centrale de 4 Go, et d’une horloge possédant une fréquence de 2 GHz.

\section{Critères d’évaluation}
Selon \cite{flouris2006classification}, le problème de l'évaluation des techniques d’intégration ou de fusion des ontologies est encore ouvert. Une comparaison générale et objective est difficile, car nous ne savons pas comment l’évaluation de ces outils pourrait être mesurée.\medskip\vspace{4px}

En effet, déterminer la qualité d'un résultat d’intégration ou de fusion nécessiterait de le comparer avec un résultat parfait ou presque parfait. Mais ce résultat est impossible à obtenir manuellement pour les grandes ontologies, et même inexistant car il pourrait y avoir plus qu’un résultat idéal. Il n'y a pas de benchmark qui pourrait être utilisé pour évaluer la qualité de l'approche proposée, \textit{e.g.}, en utilisant des mesures de qualité telles que la précision, le rappel ou la F-Mesure \cite{raunich2012towards}.\medskip\vspace{4px}

Un benchmark de la fusion ou de l’intégration des ontologies devrait être en mesure d'évaluer équitablement la qualité des différents outils. Pour ce faire, \cite{raunich2012towards} ont défini des métriques de qualité telles que :\\

$\bullet$ \large{\textbf{La qualité de l’ontologie de sortie} :} \normalsize Elle se reflète par la quantité de ses chevauchements sémantiques qui peuvent être palliés en évitant l'introduction de chemins supplémentaires (relations redondantes). La qualité de l’ontologie dépend fortement de la qualité des ontologies d’entrée ; Idéalement, les ontologies d'entrée sont correctes et ne représentent pas (ou très peu) de conflits et d’incohérences ; idéalement, les alignements d’entrée sont aussi corrects bien que ce n’est pas évident d’en obtenir pour les grandes ontologies.\medskip\vspace{4px}

$\bullet$ \large{\textbf{La couverture (Préservation de l'information)} :} \normalsize C’est une exigence clé, afin que toutes les informations représentées dans les ontologies d'entrée soient conservées dans l’ontologie résultante :\medskip\vspace{4px}

\begin{itemize}
\item[$\ast$] Pour le "\textbf{Full Merge}" où chaque paire d’entités équivalentes devient une entité fusionnée, la taille de l’ontologie résultante doit être égale à la somme du nombre d’entités des deux ontologies d'entrée, moins le nombre d’entités fusionnées, ou plutôt moins le nombre de correspondances d'équivalence (=) dans l’alignement d'entrée.\vspace{4px}

\hspace{7px}\ding{220} En dépit que ce soit considéré comme une perte d’information, le fait de ne pas couvrir toutes les entités d’entrée peut être un choix volontaire pour éviter les conflits qui sont dus à l'héritage multiple dans l’ontologie de sortie (ce qui est appelé "fusion asymétrique" par \cite{raunich2012towards}).\medskip\vspace{4px}

\item[$\ast$] Pour l’\textbf{ontologie de pont} (notre cas), la taille de l’ontologie résultante doit être égale à la somme du nombre d’entités des deux ontologies d'entrée.
\end{itemize}\medskip\vspace{4px}

$\bullet$ \large{\textbf{L'efficacité} :} \normalsize C'est le temps d’exécution de l'algorithme d'intégration ou de fusion.\medskip\vspace{4px}

$\bullet$ \large{\textbf{L'effort manuel} :} \normalsize C'est l'intervention de l’utilisateur ou de l’expert nécessaire pour le bon déroulement du processus.

\section{Présentation des ontologies utilisées}
L’expérimentation est réalisée à l'aide des bases de test Conference, Anatomy, et Large Biomedical disponibles dans le cadre de la compagne OAEI (Ontology Alignment Evaluation Initiative). Menée depuis 2004 par un groupe de chercheurs sur le matching des ontologies, l'initiative de l’évaluation des alignements d’ontologies (OAEI) est une plate-forme internationale standard d'évaluation des outils de matching. Elle vise à améliorer les différents matchers d'ontologies en évaluant et en comparant leurs forces et leurs faiblesses à l’aide d’une suite d’alignements de référence qu’elle fournit. Les résultats des campagnes d'évaluation de chaque année, ainsi que l’ensemble des bases et des alignements de référence peuvent être téléchargés sur le site Web de OAEI.\medskip\vspace{4px}

La base de test "\textbf{Conference}" est composée essentiellement de sept petites ontologies (cmt, conference, confOf, edas, ekaw, iasted, et sigkdd) décrivant le contexte de l’organisation des conférences. OAEI fournit un alignement de référence entre chaque paire de ces sept ontologies, pour avoir en tout, 21 alignements de référence.\medskip\vspace{4px}

La base de test "\textbf{Anatomy}" est composée de deux ontologies de taille moyenne : "mouse" qui décrit l’anatomie de la souris adulte, et "human" (une partie de NCIT) qui décrit l’anatomie humaine. OAEI fournit un alignement de référence entre elles.\medskip\vspace{4px}

La base de test "\textbf{Large Biomedical Ontologies}" est composée de trois ontologies volumineuses et sémantiquement riches : FMA (Foundational Model of Anatomy), SNOMED CT (Clinical Terms), et NCI (National Cancer Institute Thesaurus) contenant des dizaines de milliers de classes. La base Large Bio se subdivise en trois catégories de taille croissante (qu’on nommera 1, 2, et 3) dont la troisième est la complète. Pour \textit{FMA}, il y a \textit{FMA1} (5\%), \textit{FMA2} (13\%), et \textit{FMA3} (100\%). Pour \textit{NCI}, il y a \textit{NCI1} (10\%), \textit{NCI2} (36\%), and \textit{NCI3} (100\%). Et pour \textit{SNOMED}, il y a \textit{SNOMED1} (5\%), \textit{SNOMED2} (17\%), et \textit{SNOMED3} (40\%). OAEI fournit trois alignements de référence pour la troisième catégorie, \textit{i.e.} entre chaque paire des trois ontologies complètes. Les relations "?" contenues dans les correspondances des alignements de référence veulent dire des relations d'équivalence "=" correctes mais qui génèrent des insatisfiabilités dans l'ontologie intégrée. Ces correspondances qui causent l'incohérence de l'alignement sont détectées par le système de débugage ALCOMO et/ou les systèmes de réparation de Logmap et/ou AML.\bigskip

\ding{50} Nous n'avons pas testé notre Framework sur d'autres ontologies de plus grandes tailles, car il n'existe pas d'alignements disponibles publiés sur Internet pour de telles ontologies.

\begin{table}[!ht]
\centering
\caption{Caractéristiques des ontologies de la base Conference}\label{t1}
\resizebox{\linewidth}{!}{
\begin{tabular}{|l||r|r||r|r||r|r||r||r|}
\hline
\textbf{Conference} & Classes & Niv & Prop Obj & Niv & Prop Data & Niv & Instances & Axioms \\ \hline\hline
cmt        & 29      & 4   & 49       & 1   & 10        & 1   & 0         & 226    \\ 
conference & 59      & 7   & 46       & 2   & 18        & 1   & 0         & 285    \\ 
confOf     & 38      & 3   & 13       & 1   & 23        & 1   & 0         & 196    \\ 
edas       & 103     & 4   & 30       & 1   & 20        & 1   & 114       & 739    \\ 
ekaw       & 73      & 6   & 33       & 2   & 0         & 0   & 0         & 233    \\ 
iasted     & 140     & 6   & 38       & 1   & 3         & 1   & 4         & 358    \\ 
sigkdd     & 49      & 4   & 17       & 1   & 11        & 1   & 0         & 116    \\ \hline\hline
\textbf{Total}/Max      & 491     & 7   & 226      & 2   & 85        & 1   & 118       & 2 153  \\ \hline
\end{tabular}}
\end{table}

\begin{table}[!ht]
\centering
\caption{Caractéristiques des ontologies de la base Anatomy}\label{t3}
\resizebox{\linewidth}{!}{
\begin{tabular}{|l||r|r||r|r||r|r||r|r|}
\hline
\textbf{Anatomy}   & Classes & Niv & Prop Obj & Niv & Prop Data & Niv & Instances & Axioms \\ \hline\hline
human     & 3 304   & 13  & 2        & 1   & 0         & 0   & 0         & 11 545 \\ 
mouse     & 2 743   & 7   & 3        & 1   & 0         & 0   & 0         & 4 838  \\ \hline\hline
\textbf{Total}/Max & 6 047   & 13  & 5        & 1   & 0         & 0   & 0         & 16 383 \\ \hline
\end{tabular}}
\end{table}


%

\begin{table}[!ht]
\centering
\caption{Caractéristiques des ontologies de la base LargeBio}
\label{t5}
\resizebox{\linewidth}{!}{
\begin{tabular}{|l||r|r||r|r||r|r||r||r|}
\hline
\textbf{LargeBio}  & Classes & Niv & Prop Obj & Niv & Prop Data & Niv & Instances & Axioms  \\ \hline\hline
FMA       & 78 988  & 21  & 0        & 0   & 54        & 1   & 0         & 79 218  \\ 
NCI       & 66 724  & 17  & 123      & 6   & 67        & 1   & 0         & 96 046  \\ 
SNOMED    & 122 464 & 34  & 55       & 3   & 0         & 0   & 0         & 191 203 \\ \hline\hline
\textbf{Total}/Max & 268 176 & 34  & 178      & 6   & 121       & 1   & 0         & 366 467 \\ \hline
\end{tabular}}
\end{table}

\begin{table}[!ht]
\centering
\caption{Les alignements de référence de la base Conference}\label{t2}
\begin{threeparttable}
\begin{tabular}{|l||r|}
\hline
\textbf{Alignment}  & \textbf{Cellules} \\ \hline\hline
cmt–conference    & 15 (14)                \\ 
cmt–confOf                 & 16                 \\ 
cmt–edas                   & 13                 \\ 
cmt–ekaw                   & 11                 \\ 
cmt–iasted                 & 4                  \\ 
cmt–sigkdd                 & 12                 \\ 
conference–confOf & 15                 \\ 
conference–edas            & 17                 \\ 
conference–ekaw            & 25                 \\ 
conference–iasted          & 14                 \\ 
conference–sigkdd          & 15                 \\ 
confOf–edas       & 19                 \\ 
confOf–ekaw                & 20 (19)                \\ 
confOf–iasted              & 9                  \\ 
confOf–sigkdd              & 7                  \\ 
edas–ekaw         & 23                 \\ 
edas–iasted                & 19                 \\ 
edas–sigkdd                & 15                 \\ 
ekaw–iasted       & 10                 \\ 
ekaw–sigkdd                & 11                 \\ 
iasted–sigkdd     & 15                 \\ \hline\hline
\textbf{Total} & 305 (303)\\ \hline
\end{tabular}
    \begin{tablenotes}
        \item () : filtré
    \end{tablenotes}
    \end{threeparttable}
\end{table}

\begin{table}[!ht]
\centering
\caption{Les alignements de référence de la base Anatomy}\label{t4}
\begin{tabular}{|l||r|r|}
\hline
\multirow{-1}{*}{\textbf{Alignment}} & \multicolumn{2}{c|}{\textbf{Nombre de cellules}}  \\ \cline{2-3} 
                           & Original (1-to-N) & Filtré (1-to-1) \\ \hline\hline
human-mouse                & 1 516             & 1 491             \\ \hline
\end{tabular}
\end{table}

\begin{table}[!ht]
\centering
\caption{Les alignemens de référence de la base LargeBio}
\label{t6}
\begin{tabular}{|l||r|r||r||r|r||r|}
\hline
\multirow{2}{*}{\textbf{Alignment}} & \multicolumn{3}{c||}{Original (1-to-N)} & \multicolumn{3}{c|}{Filtré (1-to-1)} \\ \cline{2-7} 
                          & =           & ?          & \textbf{Taille}        & =           & ?          & \textbf{Taille}        \\ \hline\hline
FMA-NCI                    & 2 686       & 338        & 3 024       & 2 337       & 170        & 2 507       \\ 
FMA-SNOMED                 & 6 026       & 2 982      & 9 008       & 5 186       & 2 568      & 7 754       \\ 
SNOMED-NCI                 & 17 210      & 1 634      & 18 844      & 13 358      & 740      & 14 098      \\ \hline\hline
\textbf{Total}                      & 25 922      & 4 954      & 30 876      & 20 881      & 3 478      & 24 359      \\ \hline
\end{tabular}
\end{table}

\clearpage
\section{Captures d’écran des résultats}
Pour but de mettre en relief les axiomes d’équivalence ajoutés à l’union des ontologies sources et bien les montrer dans nos captures, nous avons choisi de faire une intégration N-à-N. Nous avons choisi d’imprimer le résultat de l’intégration des plus grandes ontologies pour prouver que notre Framework monte à l’échelle facilement (comme le montreront les temps d’exécution).\bigskip

\begin{enumerate}
\item La première ontologie insérée en entrée est "FMA 3" (whole);
\item La deuxième ontologie est "NCI 3" (whole);
\item La troisième ontologie est "SNOMED 3" (whole).
\end{enumerate}

\bigskip\bigskip\bigskip

\begin{figure}[!ht]
\centering
\includegraphics[width=1.006\textwidth]{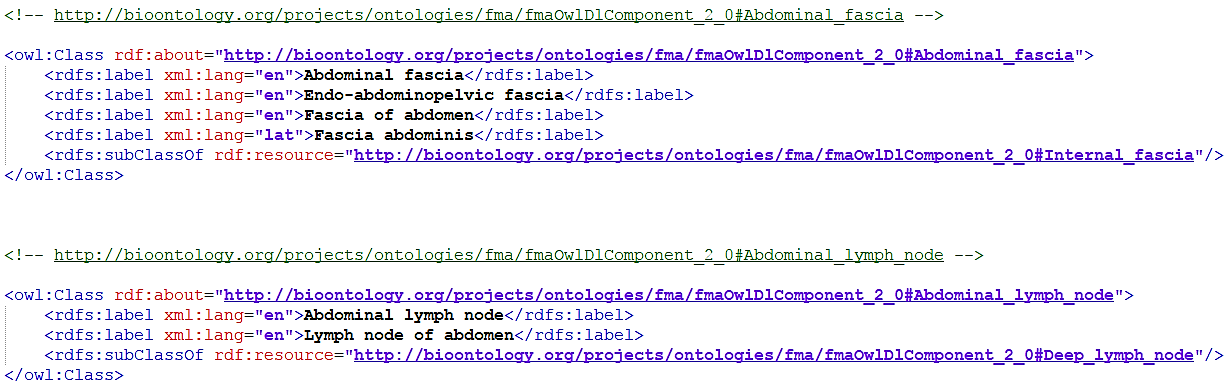}
\caption{Agrégation de "référence"}
\label{cap1}
\end{figure}

\begin{figure}[!ht]
\centering
\includegraphics[width=1.006\textwidth]{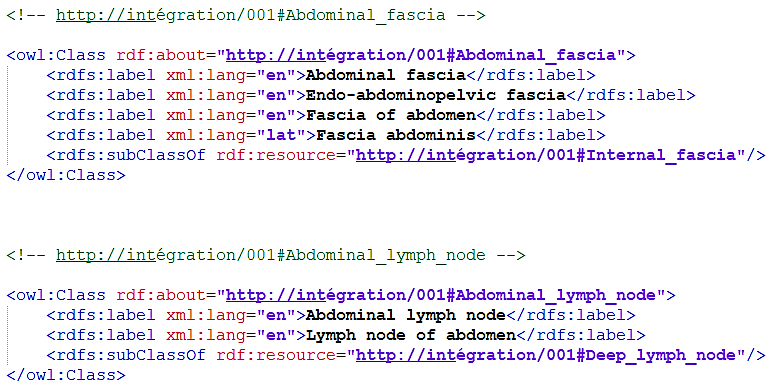}
\caption{Agrégation de OIA2R}
\label{cap2}
\end{figure}

Voici ce que donne l’exécution de la première partie du code qui réalise une agrégation / composition simple des ontologies d'entrée (\textit{i.e.}, sans les "bridging" axiomes) :\bigskip

\begin{itemize}
\item[$\diamond$] \textbf{Pour la référence} : les axiomes sont exactement identiques aux originaux (Figure \ref{cap1}).
\item[$\diamond$] \textbf{Pour OIA2R} : les axiomes sont décrits exactement comme les originaux sauf que nous personnalisons les IRIs de toutes les entités mentionnées (Figure \ref{cap2}).
\end{itemize}

\newpage

Voici ce que donne l’exécution de tout le code (avec ses deux parties) qui réalise une intégration (\textit{i.e.}, une ontologie de pont avec --les "bridging" axiomes--) :\\

\begin{figure}[!ht]
\centering
\includegraphics[width=1.006\textwidth]{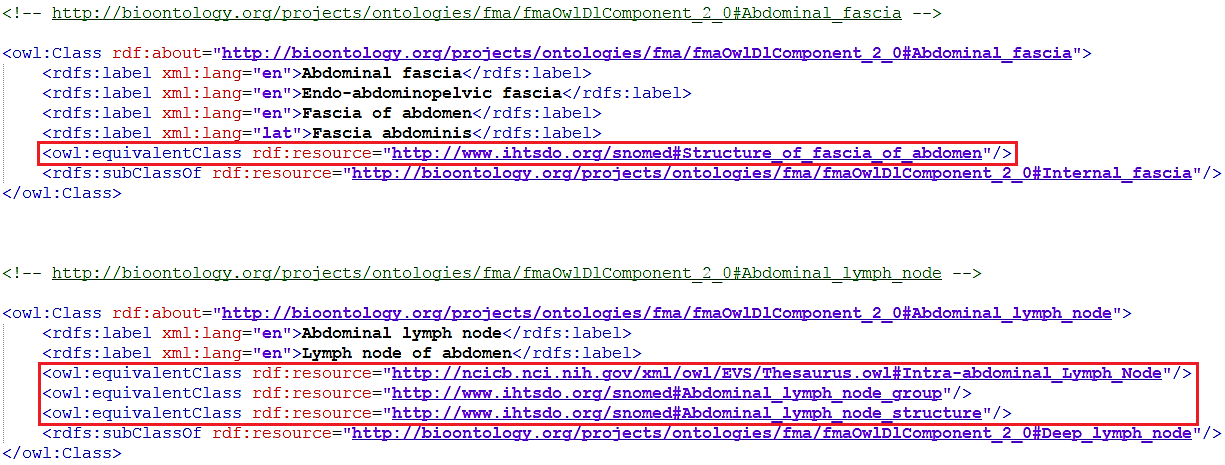}
\caption{Intégration de "référence"}
\label{cap3}
\end{figure}

\begin{figure}[!ht]
\centering
\includegraphics[width=1.006\textwidth]{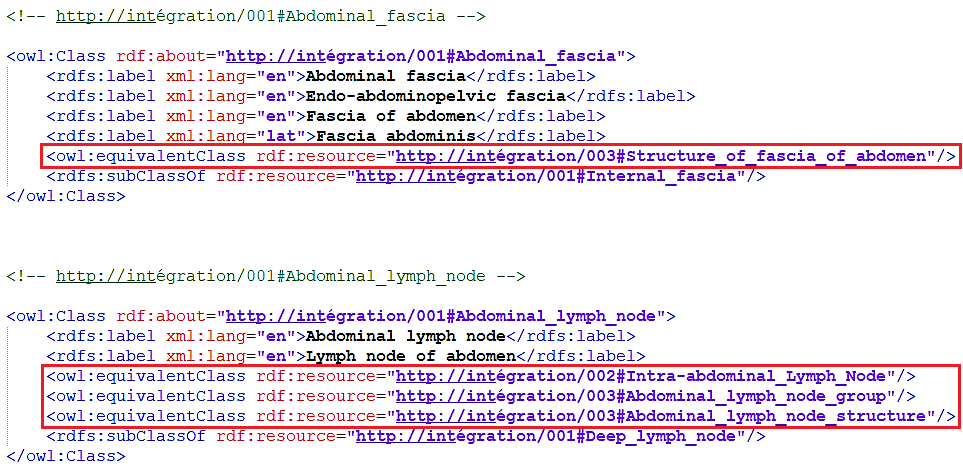}
\caption{Intégration de OIA2R}
\label{cap4}
\end{figure}

\begin{itemize}
\item[$\diamond$] \textbf{Pour la référence} : les axiomes sont exactement identiques aux originaux, mais des axiomes d’équivalence (de pont) s’y ajoutent en plus (Figure \ref{cap3}).

\item[$\diamond$] \textbf{Pour OIA2R} : les axiomes sont décrits exactement comme les originaux, mais des axiomes d’équivalence (de pont) s’y ajoutent, tout en personnalisant les IRIs de toutes les entités mentionnées (Figure \ref{cap4}).
\end{itemize}

\newpage
\section{Notions à clarifier}
\subsection{Insatisfiabilité}
C’est un terme dédié aux entités. Une classe insatisfiable est une classe ayant une description fausse (contradictoire), ce qui signifie qu'il n'est pas possible pour une instance de répondre à toutes les exigences requises pour être membre de cette classe. Elle ne peut et ne doit jamais avoir d’instances (tout à fait comme la classe "{\NoAutoSpacing owl:Nothing}"), car il n’existera aucune instance qui pourra la satisfaire \cite{greycite33925}.

\subsection{Inconsistance}
C’est un terme dédié aux ontologies. Une ontologie est consistante s'il lui existe une interprétation satisfaisante, \textit{e.g.} une ontologie à partir de laquelle nous pouvons déduire que l’individu x est différent de l’individu y et qu'il est en même temps identique à lui, ne peut pas avoir une interprétation satisfaisante. L’inconsistance peut se manifester lorsqu’il y a au moins une violation des restrictions d’une classe, une instanciation d’une classe insatisfiable, une instanciation de deux classes disjointes, ou une contradiction sémantique entre les individus \textit{etc.} \cite{greycite33921}.\medskip\vspace{4px}

Dans une ontologie inconsistante, toutes les classes sont insatisfiables, \textit{i.e.} aucune de ses classes ne peut avoir d’individu. En effet, elle n’a pas de modèle. Elle est considérée comme une ontologie sévèrement endommagée contenant une grave erreur qui doit être réparée car aucune connaissance utile ne peut en être inférée. Elle ne peut pas être publiée et utilisée dans les applications.

\subsection{Incohérence}
C’est un terme dédié aux ontologies. Une ontologie incohérente est une ontologie qui contient  quelques classes nommées insatisfiables. S’il y a au moins une classe insatisfiable (à part "{\NoAutoSpacing owl:Nothing}") dans l’ontologie, elle est incohérente \cite{greycite33925}.\medskip\vspace{4px}

Les ontologies incohérentes peuvent être publiées et utilisées dans les applications, car l’ontologie incohérente reste tout de même une ontologie consistante. Mais si nous instancions une de ses classes insatisfiables, l’ontologie incohérente deviendra inconsistante. Par conséquent, dans une ontologie incohérente, une contradiction risque de survenir.

\subsection{Raisonneur}
Un raisonneur est un composant clé dans le domaine des ontologies. Puisque la connaissance dans une ontologie OWL peut ne pas être explicite, la classification et l’interrogation d’une ontologie (qui sont les deux tâches basiques d’un raisonneur) doivent être faites par un raisonneur, pour pouvoir déduire les connaissances implicites et obtenir des résultats d’interrogation corrects. Les raisonneurs existants détectent l’inconsistance et l’incohérence, mais ne leur fournissent pas un diagnostic et une solution \cite{greycite33925}.\medskip

Voici les travaux qui ont été menés en collaboration avec notre laboratoire LIPAH concernant l'évaluation des performances des raisonneurs existants : \cite{alaya2015makes, alaya2015predicting, alaya2015towards, alaya2015ranking, alaya2016raksor, alaya2017multi}.

\subsection{Classification}
Selon \cite{greycite33925}, un raisonneur détermine toutes les inférences de la forme\\"A subClassOf B" d’une ontologie donnée, \textit{i.e.} il détermine sa hiérarchie, en appliquant les tests suivants :
\shorthandoff{:}
\begin{itemize}
\item Si A = "owl:Thing", et B = "owl:Nothing", alors il s’agit d’un test de consistance.
\item Si A = "une classe", et B = "owl:Nothing", il s’agit d’un test de satisfiabilité.
\item Si A et B sont toutes les deux des classes, il s’agit d’un test de subsomption.
\end{itemize}
\shorthandon{:}

\section{Résultats et évaluation}

\subsection{Résultats}

\begin{table}[!ht]
\centering
\caption[Caractéristiques de l’ontologie intégrée ou agrégée]{Caractéristiques de l’ontologie résultant d'une intégration ou d'une agrégation}
\label{tab9}
\begin{tabular}{|l||r|r|r|r|}
\hline
\multirow{2}{*}{\begin{tabular}[c]{@{}c@{}}\textbf{Entrées} =\\ ontologies (+ alignements)\end{tabular}} & \multicolumn{4}{c|}{\textbf{Sortie} = Ontologie résultante}         \\ \cline{2-5} 
                                                                                               & Classes & Propriétés d’objet & Propriétés data & Individus \\ \hline\hline
« \textbf{Conference} »                                                                                     & 491     & 226                & 85              & 118       \\ 
« \textbf{Anatomy} »                                                                                        & 6 047   & 5                  & 0               & 0         \\ 
« \textbf{Large Bio} »                                                                               & 268 176 & 178                & 121             & 0         \\ \hline
\end{tabular}\vspace{7px}
\end{table}

\begin{table}[!ht]
\centering
\caption{Qualité de l'ontologie résultant d'une agrégation}
\label{my-bel}
\begin{tabular}{|l||r|r||r|r||r|r|}
\hline
\multirow{3}{*}{\begin{tabular}[c]{@{}c@{}}\\\textbf{Ontologies}\\ \textbf{d'entrée}\end{tabular}} & \multicolumn{6}{c|}{\textbf{Ontologie de sortie}}                                                                                                                                                                    \\ \cline{2-7} 
                                                                            & \multicolumn{2}{c||}{\begin{tabular}[c]{@{}c@{}}\textbf{Classes}\\ \textbf{insatisfiables}\end{tabular}} & \multicolumn{2}{c||}{\begin{tabular}[c]{@{}c@{}}\textbf{Axioms}\\ \textbf{logiques}\end{tabular}} & \multicolumn{2}{c|}{\textbf{Consistance}} \\ \cline{2-7} 
                                                                            & OIA2R                                      & Réf                                     & OIA2R                                 & Réf                                   & OIA2R            & Réf           \\ \hline\hline
Conference                                                                  & 0                                          & 0                                       & 1 860                                 & 2 153                                 & \ding{51}              & \ding{51}           \\ \hline
Anatomy                                                                     & 0                                          & 0                                       & 6 635                                 & 16 383                                & \ding{51}              & \ding{51}           \\ \hline
LargeBio                                                                    & 0                                          & 0                                       & 244 942                               & 366 467                               & \ding{51}              & \ding{51}           \\ \hline
\end{tabular}
\end{table}

\begin{table}[!ht]
\centering
\caption[Qualité de l'ontologie résultant de l'intégration de Conference]{Qualité de l'ontologie résultant de l'intégration des ontologies de Conference}
\begin{tabular}{|l||r|r||r|r|}
\hline
\multirow{3}{*}{\begin{tabular}[c]{@{}c@{}}\textbf{Conference}\end{tabular}} & \multicolumn{4}{c|}{\textbf{Nombre des classes insatisfiables}}                                            \\ \cline{2-5} 
                                                                                  & \multicolumn{2}{c||}{Al originaux (1-à-N)} & \multicolumn{2}{c|}{Al filtrés (1-à-1)} \\ \cline{2-5} 
                                                                                  & OIA2R                 & Réf               & OIA2R                & Réf                \\ \hline\hline
\begin{tabular}[c]{@{}c@{}}Intégration\\ 2-à-2\end{tabular}                      & 0                     & 5                 & 0                    & 5                  \\ \hline
\begin{tabular}[c]{@{}c@{}}Intégration\\ 1-à-N\end{tabular}                      & 0                     & 0                 & 0                    & 0                  \\ \hline
\begin{tabular}[c]{@{}c@{}}Intégration\\ N-à-N\end{tabular}                      & 54                    & \ding{53}                & 54                   & \ding{53}                 \\ \hline
\end{tabular}
\end{table}

\begin{table}[!ht]
\centering
\caption[Préservation des axiomes après l'intégration de Conference]{Préservation des axiomes après l'intégration des ontologies de Conference}
\begin{tabular}{|l||r|r|r||r|r|r|}
\hline
\multirow{3}{*}{\begin{tabular}[c]{@{}c@{}}\textbf{Conference}\end{tabular}} & \multicolumn{6}{c|}{\textbf{Nombre des axiomes logiques}}                                                                                                                 \\ \cline{2-7} 
                                                                                  & \multicolumn{3}{c||}{Al originaux}                                   & \multicolumn{3}{c|}{Al filtrés}                                   \\ \cline{2-7} 
                                                                                  & OIA2R & Réf   & Attendus                                                    & OIA2R & Réf   & Attendus                                                    \\ \hline\hline
\begin{tabular}[c]{@{}c@{}}Intégration\\ 2-à-2\end{tabular}                            & 1 957 & 2 250 & \begin{tabular}[c]{@{}c@{}}\textbf{2 250}\\ (2 153 +\\97)\end{tabular}  & 1 956 & 2 249 & \begin{tabular}[c]{@{}c@{}}\textbf{2 249}\\ (2 153 +\\96)\end{tabular}  \\ \hline
\begin{tabular}[c]{@{}c@{}}Intégration\\ 1-à-N\end{tabular}                            & 1 931 & 2 224 & \begin{tabular}[c]{@{}c@{}}\textbf{2 224}\\ (2 153 +\\71)\end{tabular}  & 1 930 & 2 223 & \begin{tabular}[c]{@{}c@{}}\textbf{2 223}\\ (2 153 +\\70)\end{tabular}  \\ \hline
\begin{tabular}[c]{@{}c@{}}Intégration\\ N-à-N\end{tabular}                            & 2 165 & 2 458 & \begin{tabular}[c]{@{}c@{}}\textbf{2 458}\\ (2 153 +\\305)\end{tabular} & 2 163 & 2 456 & \begin{tabular}[c]{@{}c@{}}\textbf{2 456}\\ (2 153 +\\303)\end{tabular} \\ \hline
\end{tabular}
\end{table}

\begin{table}[!ht]
\centering
\caption[Qualité de l'ontologie résultant de l'intégration de Anatomy]{Qualité de l'ontologie résultant de l'intégration des ontologies de Anatomy}
\begin{tabular}{|l||r|r||r|r|}
\hline
\multirow{3}{*}{\begin{tabular}[c]{@{}c@{}}\textbf{Anatomy}\end{tabular}} & \multicolumn{4}{c|}{\textbf{Nombre de classes insatisfiables}}                                            \\ \cline{2-5} 
                                                                               & \multicolumn{2}{c||}{Al originaux (1-à-N)} & \multicolumn{2}{c|}{Al filtrés (1-à-1)} \\ \cline{2-5} 
                                                                               & OIA2R                 & Réf               & OIA2R                & Réf                \\ \hline\hline
\begin{tabular}[c]{@{}c@{}}Intégration\\ 2-à-2\end{tabular}                   & 0                     & 0                 & 0                    & 0                  \\ \hline
\end{tabular}
\end{table}

\begin{table}[!ht]
\centering
\caption[Préservation des axiomes après l'intégration de Anatomy]{Préservation des axiomes après l'intégration des ontologies de Anatomy}
\begin{tabular}{|l||r|r|r||r|r|r|}
\hline
\multirow{3}{*}{\begin{tabular}[c]{@{}c@{}}\textbf{Anatomy}\end{tabular}} & \multicolumn{6}{c|}{\textbf{Nombre des axiomes logiques}}                                                                                                                       \\ \cline{2-7} 
                                                                               & \multicolumn{3}{c||}{Al originaux}                                        & \multicolumn{3}{c|}{Al filtrés}                                        \\ \cline{2-7} 
                                                                               & OIA2R & Réf    & Attendus                                                        & OIA2R & Réf    & Attendus                                                        \\ \hline\hline
\begin{tabular}[c]{@{}c@{}}Intégration\\ 2-à-2\end{tabular}                   & 8 151 & 17 899 & \begin{tabular}[c]{@{}c@{}}\textbf{17 899}\\ (16 383\\+ 1 516)\end{tabular} & 8 126 & 17 874 & \begin{tabular}[c]{@{}c@{}}\textbf{17 874}\\ (16 383\\+ 1 491)\end{tabular} \\ \hline
\end{tabular}
\end{table}

\begin{table}[!ht]
\centering
\caption[Qualité de l'ontologie résultant de l'intégration de LargeBio]{Qualité de l'ontologie résultant de l'intégration des ontologies de LargeBio}
\begin{tabular}{|l|l||r|r||r|r|}
\hline
\multicolumn{2}{|c||}{\multirow{3}{*}{\begin{tabular}[c]{@{}c@{}}\textbf{LargeBio}\end{tabular}}}                                    & \multicolumn{4}{c|}{\textbf{Nombre de classes insatisfiables}}                                            \\ \cline{3-6} 
\multicolumn{2}{|c||}{}                                                                                                                  & \multicolumn{2}{c||}{Al originaux} & \multicolumn{2}{c|}{Al filtrés} \\ \cline{3-6} 
\multicolumn{2}{|c||}{}                                                                                                                  & OIA2R                & Réf                & OIA2R               & Réf                 \\ \hline\hline
\multirow{2}{*}{\begin{tabular}[c]{@{}c@{}}Intég\\ 2-à-2\end{tabular}} & \begin{tabular}[c]{@{}c@{}}Al\\ originaux\end{tabular} & 120 743              & 190 486            & 67 342              & 141 941             \\ \cline{2-6} 
                                                                              & \begin{tabular}[c]{@{}c@{}}Al\\ réparés\end{tabular}   & 11 978               & -                  & \textbf{11 078}              & \textbf{-}                   \\ \hline\hline
\multirow{2}{*}{\begin{tabular}[c]{@{}c@{}}Intég\\ 1-à-N\end{tabular}} & \begin{tabular}[c]{@{}c@{}}Al\\ originaux\end{tabular} & 58 608               & 118 579            & 27 773              & 65 043              \\ \cline{2-6} 
                                                                              & \begin{tabular}[c]{@{}c@{}}Al\\ réparés\end{tabular}   & 56                   & -                  & \textbf{48}                  & \textbf{96}                 \\ \hline\hline
\multirow{2}{*}{\begin{tabular}[c]{@{}c@{}}Intég\\ N-à-N\end{tabular}} & \begin{tabular}[c]{@{}c@{}}Al\\ originaux\end{tabular} & 136 301              & 206 232            & 80 320             & 157 121             \\ \cline{2-6} 
                                                                              & \begin{tabular}[c]{@{}c@{}}Al\\ réparés\end{tabular}   & 14 655               & -                  & \textbf{12 919}             & \textbf{-}                   \\ \hline
\end{tabular}
\end{table}

\begin{table}[!ht]
\centering
\caption[Préservation des axiomes après l'intégration de LargeBio]{Préservation des axiomes après l'intégration des ontologies de LargeBio}
\begin{tabular}{|l|l||r|r|r||r|r|r|}
\hline
\multicolumn{2}{|c||}{\multirow{3}{*}{\begin{tabular}[c]{@{}c@{}}\textbf{LargeBio}\end{tabular}}}                                & \multicolumn{6}{c|}{\textbf{Nombre des axiomes logiques}}                                                                                                                                       \\ \cline{3-8} 
\multicolumn{2}{|c||}{}                                                                                                          & \multicolumn{3}{c||}{Al originaux}                                                & \multicolumn{3}{c|}{Al filtrés}                                                \\ \cline{3-8} 
\multicolumn{2}{|c||}{}                                                                                                          & OIA2R   & Réf     & Attendus                                                             & OIA2R   & Réf     & Attendus                                                             \\ \hline\hline
\multirow{2}{*}{\begin{tabular}[c]{@{}c@{}}Intég\\ 2-à-2\end{tabular}} & \begin{tabular}[c]{@{}c@{}}Al\\ originaux\end{tabular} & 266 810 & 388 335 & \begin{tabular}[c]{@{}c@{}}\textbf{388 335}\\ (366 467\\+ 21 868)\end{tabular} & 261 547 & 383 072 & \begin{tabular}[c]{@{}c@{}}\textbf{383 072}\\ (366 467\\+ 16 605)\end{tabular} \\ \cline{2-8} 
                                                                        & \begin{tabular}[c]{@{}c@{}}Al\\ réparés\end{tabular} & 264 838 & 386 363 & \begin{tabular}[c]{@{}c@{}}\textbf{386 363}\\ (366 467\\+ 19 896)\end{tabular} & 260 637 & 382 162 & \begin{tabular}[c]{@{}c@{}}\textbf{382 162}\\ (366 467\\+ 15 695)\end{tabular} \\ \hline\hline
\multirow{2}{*}{\begin{tabular}[c]{@{}c@{}}Intég\\ 1-à-N\end{tabular}} & \begin{tabular}[c]{@{}c@{}}Al\\ originaux\end{tabular} & 256 974 & 378 499 & \begin{tabular}[c]{@{}c@{}}\textbf{378 499}\\ (366 467\\+ 12 032)\end{tabular} & 255 203 & 376 728 & \begin{tabular}[c]{@{}c@{}}\textbf{376 728}\\ (366 467\\+ 10 261)\end{tabular} \\ \cline{2-8} 
                                                                        & \begin{tabular}[c]{@{}c@{}}Al\\ réparés\end{tabular} & 253 654 & 375 179 & \begin{tabular}[c]{@{}c@{}}\textbf{375 179}\\ (366 467\\+ 8 712)\end{tabular}  & 252 465 & 373 990 & \begin{tabular}[c]{@{}c@{}}\textbf{373 990}\\ (366 467\\+ 7 523)\end{tabular}  \\ \hline\hline
\multirow{2}{*}{\begin{tabular}[c]{@{}c@{}}Intég\\ N-à-N\end{tabular}} & \begin{tabular}[c]{@{}c@{}}Al\\ originaux\end{tabular} & 275 818 & 397 343 & \begin{tabular}[c]{@{}c@{}}\textbf{397 343}\\ (366 467\\+ 30 876)\end{tabular} & 269 301 & 390 826 & \begin{tabular}[c]{@{}c@{}}\textbf{390 826}\\ (366 467\\+ 24 359)\end{tabular} \\ \cline{2-8} 
                                                                        & \begin{tabular}[c]{@{}c@{}}Al\\ réparés\end{tabular} & 270 864 & 392 389 & \begin{tabular}[c]{@{}c@{}}\textbf{392 389}\\ (366 467\\+ 25 922)\end{tabular} & 265 823 & 387 348 & \begin{tabular}[c]{@{}c@{}}\textbf{387 348}\\ (366 467\\+ 20 881)\end{tabular} \\ \hline
\end{tabular}
\end{table}

\clearpage
Notez que toutes les classes insatisfiables sont causées par la préservation des connaissances de disjonction des alignements d’entrée.
En effet, si nous ne gardons pas d'axiomes de disjonction, nous n'obtiendrons aucune classe insatisfaisable, et toutes nos ontologies de sortie seraient cohérentes et consistantes, mais incomplètes (\textit{i.e.}, manquant des informations de disjonction précieuses).
Nous concluons que, lorsque toutes les correspondances sont des correspondances d'équivalence, la seule cause de conflits est les relations "\textit{DisjointWith}" issues des ontologies d'entrée.

\subsection{Constatations}
Dans tous les cas d’intégration, notre ontologie finale est complète dans le sens où elle conserve toutes les entités et la hiérarchie des ontologies d'entrée, et toutes les correspondances des alignements d'entrée. Elle ne parvient pas pourtant à préserver tous les axiomes des ontologies d'entrée, contrairement à l'ontologie de "référence". Dans l’agrégation (sans "bridging" axiomes), notre ontologie n’aura aucun ajout de classe insatisfiable, et le nombre de niveaux de sa hiérarchie sera toujours égal au nombre maximal des niveaux de hiérarchie des ontologies d’entrée. Dans l'ontologie de pont, nous constatons que suite à l’ajout des "bridging" axiomes, le raisonneur HermiT génère beaucoup trop de classes insatisfiables. Nous remarquons aussi que dans l’ontologie résultant d’une intégration N-à-N, le nombre de classes insatisfiables est beaucoup plus important que celui de l’ontologie résultant d’une intégration 2-à-2 ou 1-à-N. L’ontologie de référence contient toujours plus de classes insatisfiables que notre ontologie. Par conséquent, nous serions dans le risque d’avoir une inconsistance possible si jamais il y avait un individu instancié par une des classes insatisfiables, ou si jamais des individus avaient un conflit entre eux suite à des axiomes de pont de type "sameAs". Dans notre cas, les ontologies sources utilisées ne contiennent pas d’instances, à part "edas" et "iasted" dont toutes les instances sont instanciées par des classes qui n’ont aucune correspondance avec d’autres classes, donc sont hors de danger.\medskip\vspace{4px}

Nous constatons aussi qu’une intégration d’ontologies de différents domaines (comme dans "Anatomy") génère toujours moins de conflits qu’une intégration d’ontologies de même domaine (comme dans "Conference" et "Large Bio").\medskip\vspace{4px}

A part cela, nous ne parvenons pas toujours à avoir un nombre de niveaux fixe (un niveau maximal) dans la hiérarchie des classes de notre ontologie, car le raisonneur qui découvre les niveaux (et les classes de chaque niveau) ne termine pas son raisonnement. On dirait que, suite à l’ajout des "bridging" axiomes, le raisonneur rencontre une boucle infinie (un cercle vicieux) dans son raisonnement.

\subsection{Interprétations}
Ces insatisfiabilités sont dues au fait que l’axiome d’équivalence entre deux entités est formellement équivaut à deux axiomes de subsomption réciproques :

\begin{center}
\framebox{\textbf{equivalentClass (C1 C2) = subClassOf (C1 C2) + subClassOf (C2 C1)}}
\end{center}

\ding{220} Et l’ajout de ces subsomptions implicites aux équivalences infère de nouvelles connaissances qui peuvent être contradictoires.\medskip\vspace{4px}

\ding{220} Ces axiomes d’équivalence vont aussi affecter la hiérarchie (la classification) des classes et des propriétés de l’ontologie résultante. En effet, le raisonneur pourra ne pas pouvoir s’arrêter (dans son calcul) et ainsi ne pas avoir un nombre précis de niveaux dans la hiérarchie.\\

\begin{figure}[!ht]
\centering
\includegraphics[width=1.005\textwidth]{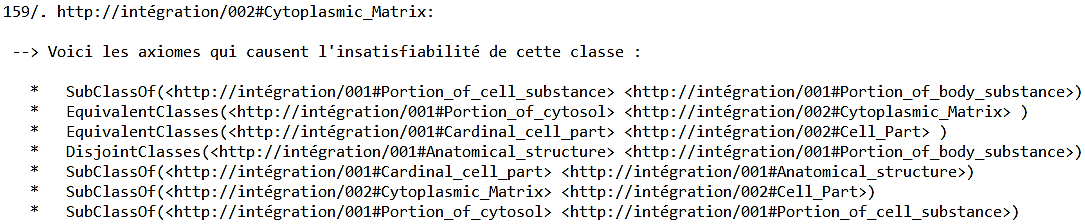}
\caption[Debugage d'une classe insatisfiable 2]{Debugage d'une classe insatisfiable dans une ontologie de pont}
\label{justt}
\end{figure}

Ci-dessous un exemple qui montre comment se forme l’insatisfiabilité d’une classe dans une ontologie de pont. La figure \ref{justt} montre les justifications que nous a affichées le debugueur de OWL API à l’aide du raisonneur HermiT pour une des classes insatisfiables (Cytoplasmic\_Matrix) de l’ontologie résultant d’une intégration des ontologies FMA 1 (Ont\textbf{1}) et NCI 1 (Ont\textbf{2}).\medskip\vspace{4px}

\begin{figure}[!ht]
\centering
\includegraphics[scale=0.59]{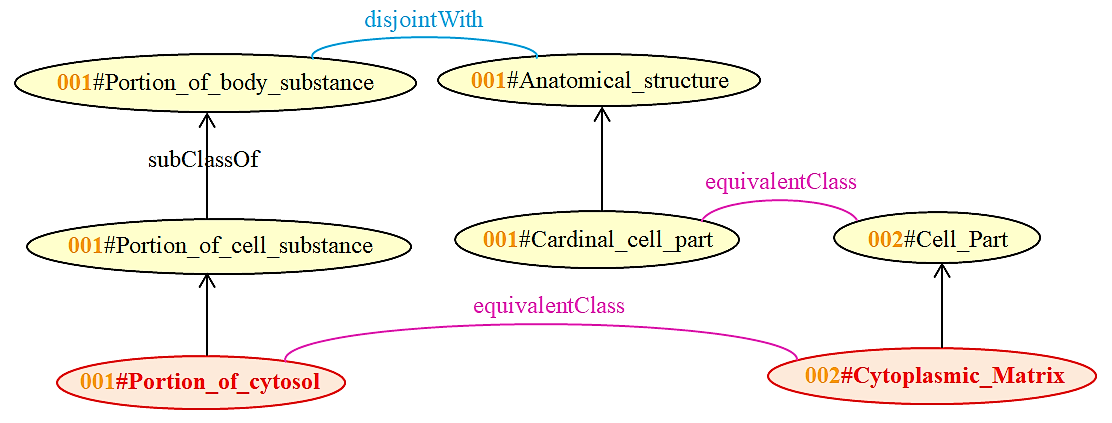}
\caption[Formation des classes insatisfiables 2]{Formation des classes insatisfiables}
\label{i}
\end{figure}

Dans la figure \ref{i}, nous modélisons la représentation graphique des axiomes de justification ci-dessus, où les classes en rouge sont les classes insatisfiables.

\begin{figure}[!ht]
\centering
\includegraphics[scale=0.59]{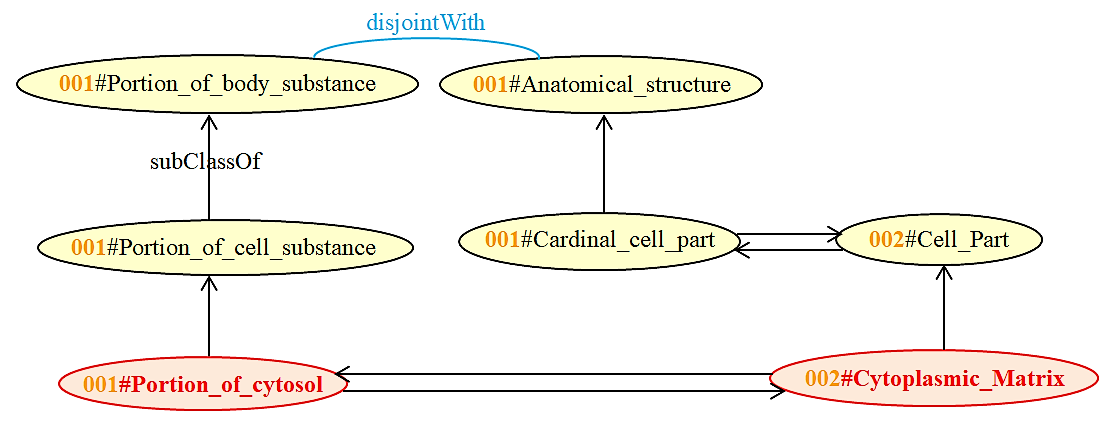}
\caption[Cause de l'incohérence 2]{Cause de l'incohérence d'une ontologie de pont}
\label{ins}
\end{figure}

Sachant que la relation d’équivalence (des "bridging" axiomes) est en fait égale à deux relations de subsomption dans les deux sens, le schéma devient comme le montre la figure \ref{ins}.


\newpage
\sloppy{
Nous remarquons qu’après l’ajout des "bridging" axiomes d’équivalence, la classe "002\#Cytoplasmic\_Matrix" (provenant de l'ontologie NCI 1 (Ont\textbf{2})) devient par inférence sous-classe des deux classes "001\#Portion\_of\_body\_structure" et "001\#Anatomical\_structure" qui sont disjointes (information extraite de l’ontologie FMA 1 (Ont\textbf{1})). Ceci est contradictoire, car une classe ne peut pas être sous-classe de deux classes disjointes. La même chose s’applique pour l’autre classe coloriée en rouge.}\medskip

La référence ne manque aucun axiome des ontologies d’entrée, alors que la nôtre a des pertes d’informations telles que les entités anonymes et les restrictions. C'est pour cette raison que le raisonneur infère plus d’informations et ainsi détecte plus de classes insatisfiables dans l'ontologie de référence. D’ailleurs, pour calculer toutes ses classes insatisfiables, le raisonnement HermiT prend beaucoup plus de temps pour terminer son calcul.\medskip

L’ontologie d’une intégration N-à-N produit beaucoup plus de classes insatisfiables que l’ontologie d’une intégration 2-à-2 ou 1-à-N à cause des relations redondantes. Plus il y a de relations redondantes (dans ce cas, des relations d’équivalence redondantes entre les entités sources), plus il y a d’incohérence. Les correspondances redondantes peuvent être déduites automatiquement par un raisonneur, ainsi elles sont inutiles, et surtout source de conflits sémantiques.\medskip

Il est important également de noter que la réparation de l’alignement est dédiée à l’intégration de deux ontologies seulement. Autrement dit, si nous intégrons deux ontologies en utilisant un alignement réparé entre elles, nous obtiendrons une ontologie consistante et cohérente sans aucune classe insatisfiable. Cependant, si nous intégrons plusieurs ontologies à l'aide de plusieurs alignements réparés (\textit{i.e.}, entre les paires d'ontologies), nous obtiendrons une ontologie comportant plusieurs classes insatisfiables. En effet, les systèmes actuels de réparation des alignement prennent en entrée deux ontologies (à intégrer ultérieurement) et un alignement entre elles. Ils ne sont pas en mesure de prendre en compte plusieurs ontologies et alignements entre elles pour but de les intégrer simultanément.

\section{Atouts de OIA2R}

Notre framework est automatique et générique. Il prend en entrée toute ontologie et tout alignement avec lesquels il produira une nouvelle ontologie qui les englobe tous. Ce processus est rapide même pour les plus grandes ontologies et les plus grands alignements. L’ontologie résultante est assez complète et cohérente.\medskip\vspace{4px}

Nous donnons la possibilité de convertir les alignements sources en des mappings, et également de les réparer à l’aide d’outils externes, afin de minimiser les erreurs dans l’ontologie de sortie.\medskip\vspace{4px}

Notre approche permet un refactoring (une personnalisation) de toutes les entités des ontologies et des alignements sources pour former une ontologie propre à nous (dont les entités ne pointent pas sur les URIs externes des ontologies sources déjà publiées). En effet, l’utilisateur n’a qu’à entrer l’URI qu’il désire pour la future ontologie, et par la suite toutes les entités l’auront comme URI de préfixe.

\section{Temps d’exécution}
Ce sont les temps d’exécution moyens d’une intégration complète N-à-N avec des alignements transformés en mappings, mais qui sont déjà réparés à l'avance. Nous voulons dire par "loading", le temps de chargement des ontologies dans le manager de OWL API, car pour les grandes ontologies, leur loading prend une bonne part du temps d’exécution (comme le montrera le tableau \ref{tab23}), et ce temps-là ne fait pas partie du temps effectif de notre intégration. Le temps "avec loading" est le temps d’exécution de tout le programme (du début jusqu'à la fin), et le temps "sans loading" est le temps d’exécution exact de notre framework. Rappelons que notre framework prend en entrée des alignements pré-établis, ainsi, les temps de d'exécution fournis ne comprennent pas le temps du matching.

\begin{table}[!ht]
\centering
\caption{Temps d’exécution d’une intégration N-à-N}
\label{tab23}
\begin{tabular}{|l||r|r||r|r|}
\hline
\multirow{2}{*}{\begin{tabular}[c]{@{}c@{}}\textbf{Tps d’exécution}\\ \textbf{CPU} (\textbf{s})\end{tabular}} & \multicolumn{2}{c||}{\textbf{Notre intégration}} & \multicolumn{2}{c|}{\textbf{La référence}} \\ \cline{2-5} 
                                                                                 & +loading           & -loading          & +loading              & -loading              \\ \hline\hline
Conference                                                                       & 1,531              & 0,406             & 1,375                 & 0,171                 \\ 
Anatomy                                                                          & 3,093              & 0,703             & 3,562                 & 0,453                 \\ 
Large Bio                                                                & 36,859              & 8,406             & 41,375                 & 4,890                 \\ \hline
\end{tabular}\bigskip
\end{table}

Les temps d’exécution CPU ne dépassent pas 0,7 min pour les plus grandes ontologies. Le temps global de l'intégration de référence est plus long, car la référence prend plus de temps pour sauvegarder tous les axiomes d'entrée (à la fin du programme) et créer l'ontologie de pont complète. Cependant, le temps global de l'intégration de OIA2R prend moins de temps pour sauvegarder les axiomes d'entrée, car OIA2R manque les axiomes complexes. Le temps effectif de l'intégration de référence prend moins de temps, car la référence ne perd pas du temps à parser, refactoriser les axiomes des ontologies d'entrée et créer de nouveaux axiomes refactorisés comme le fait OIA2R, elle assemble directement les axiomes des ontologies d’entrée et ajoute les axiomes de pont.

\section{Ouvrir des horizons}
Pour découvrir le niveau d’intégration qui génère plus d’incohérences dans son ontologie résultante (soit l’ontologie de pont, soit la fusion), nous avons réalisé un autre travail qui fait une fusion de deux ontologies (uniquement deux), \textit{i.e.} il fusionne les paires d’entités équivalentes en une seule entité.

\begin{table}[!ht]
\centering
\caption{Qualité de l'ontologie intégrée (LargeBio OAEI Task 1)}
\begin{tabular}{|l||r|r||r|r|}
\hline
\multirow{3}{*}{\begin{tabular}[c]{@{}c@{}}\textbf{\textit{FMA}1-\textit{NCI}1}\end{tabular}} & \multicolumn{4}{c|}{\textbf{Nombre de classes insatisfiables}}                                  \\ \cline{2-5} 
                                                                                 & \multicolumn{2}{c||}{Al original} & \multicolumn{2}{c|}{Al filtré} \\ \cline{2-5} 
                                                                                 & OIA2R             & Full merge            & OIA2R             & Full Merge            \\ \hline\hline
Al original                                                                      & 1 727             & 826                   & 410               & 173                   \\ \hline
Al réparé                                                                      & 0                 & 0                     & 0                 & 0                     \\ \hline
\end{tabular}
\end{table}

\begin{table}[!ht]
\centering
\caption{Qualité de l'ontologie intégrée (LargeBio OAEI Task 3)}
\begin{tabular}{|l||r|r||r|r|}
\hline
\multirow{3}{*}{\begin{tabular}[c]{@{}c@{}}\textbf{\textit{FMA}2-\textit{SNOMED}1}\end{tabular}} & \multicolumn{4}{c|}{\textbf{Nombre de classes insatisfiables}}                                  \\ \cline{2-5} 
                                                                                 & \multicolumn{2}{c||}{Al original} & \multicolumn{2}{c|}{Al filtré} \\ \cline{2-5} 
                                                                                 & OIA2R             & Full merge            & OIA2R             & Full Merge            \\ \hline\hline
Al original                                                                      & 13 508             & 7 212                   & 10 048               & 4 379                   \\ \hline
Al réparé                                                                      & 0                 & 0                     & 0                 & 0                     \\ \hline
\end{tabular}
\end{table}

\begin{table}[!ht]
\centering
\caption{Qualité de l'ontologie intégrée (LargeBio OAEI Task 5)}
\begin{tabular}{|l||r|r||r|r|}
\hline
\multirow{3}{*}{\begin{tabular}[c]{@{}c@{}}\textbf{\textit{NCI}2-\textit{SNOMED}2}\end{tabular}} & \multicolumn{4}{c|}{\textbf{Nombre de classes insatisfiables}}                                  \\ \cline{2-5} 
                                                                                 & \multicolumn{2}{c||}{Al original} & \multicolumn{2}{c|}{Al filtré} \\ \cline{2-5} 
                                                                                 & OIA2R             & Full merge            & OIA2R             & Full Merge            \\ \hline\hline
Al original                                                                      & 34 639             &  19 132                  & 25 637               & 12 990                   \\ \hline
Al réparé                                                                      & 0                 & 0                     & 0                 & 0                     \\ \hline
\end{tabular}
\end{table}

\begin{table}[!ht]
\centering
\caption{Qualité de l'ontologie intégrée (LargeBio OAEI Task 2)}
\begin{tabular}{|l||r|r||r|r|}
\hline
\multirow{3}{*}{\begin{tabular}[c]{@{}c@{}}\textbf{\textit{FMA}3-\textit{NCI}3}\end{tabular}} & \multicolumn{4}{c|}{\textbf{Nombre de classes insatisfiables}}                                  \\ \cline{2-5} 
                                                                                 & \multicolumn{2}{c||}{Al original} & \multicolumn{2}{c|}{Al filtré} \\ \cline{2-5} 
                                                                                 & OIA2R             & Full merge            & OIA2R             & Full Merge            \\ \hline\hline
Al original                                                                      & 7 175             & 6 272                   &  1 158              &  995                  \\ \hline
Al réparé                                                                      & 0                 & 0                     & 0                 & 0                     \\ \hline
\end{tabular}
\end{table}

\begin{table}[!ht]
\centering
\caption{Qualité de l'ontologie intégrée (LargeBio OAEI Task 6)}
\begin{tabular}{|l||r|r||r|r|}
\hline
\multirow{3}{*}{\begin{tabular}[c]{@{}c@{}}\textbf{\textit{NCI}3-\textit{SNOMED}3}\end{tabular}} & \multicolumn{4}{c|}{\textbf{Nombre de classes insatisfiables}}                                  \\ \cline{2-5} 
                                                                                 & \multicolumn{2}{c||}{Al original} & \multicolumn{2}{c|}{Al filtré} \\ \cline{2-5} 
                                                                                 & OIA2R             & Full merge            & OIA2R             & Full Merge            \\ \hline\hline
Al original                                                                      &    92 149          &      76 280             &    49 825            &   42 331                 \\ \hline
Al réparé                                                                      & 0                 & 0                     & 0                 & 0                     \\ \hline
\end{tabular}
\end{table}

\newpage
\ding{220} Nous remarquons (comme l’avaient dit \cite{raunich2012towards}) que la fusion complète des ontologies génère toujours moins de conflits qu’une ontologie de pont (appelée "fusion directe" selon eux).\medskip

\begin{figure}[!ht]
\centering
\includegraphics[width=1.005\textwidth]{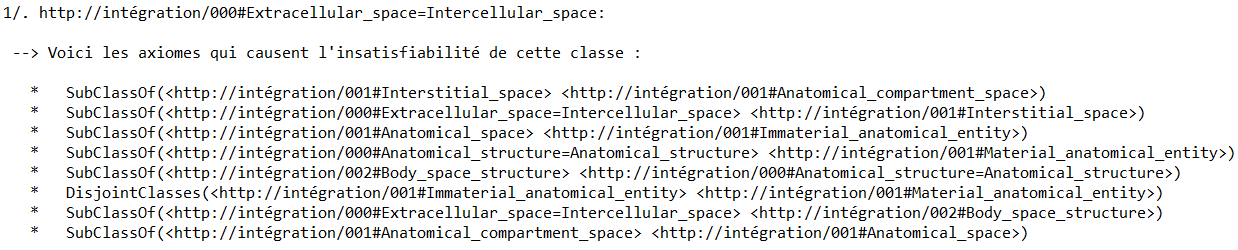}
\caption[Debugage d'une classe insatisfiable 3]{Debugage d'une classe insatisfiable après une fusion}
\label{de}
\end{figure}

Ci-dessous un exemple qui montre comment se forme l’insatisfiabilité d’une classe dans une ontologie résultante d'un processus de fusion complète. Dans la figure \ref{de}, nous schématisons les justifications que nous a affichées le debugueur de OWL API (à l’aide du raisonneur HermiT) pour une des classes insatisfiables générées suite à la fusion de FMA 1 (Ont\textbf{1}) et SNOMED 1 (Ont\textbf{2}).

Nous avons choisi le \textbf{ID} "/000" pour les entités issues de la fusion de deux entités équivalentes. Et nous avons choisi de leur donner comme noms, une concaténation des noms des paires d'entités fusionnées (juste pour pouvoir voir clairement dans l'ontologie résultante ce qui a été fusionné). Dans l'état de l'art, les auteurs choisissent généralement l'un des deux noms des entités fusionnées (peut-être après avoir défini une ontologie prioritaire), ou bien créent un code unique tout en ajoutant les deux noms originaux comme labels.\medskip

\begin{figure}[!ht]
\centering
\includegraphics[width=1.007\textwidth]{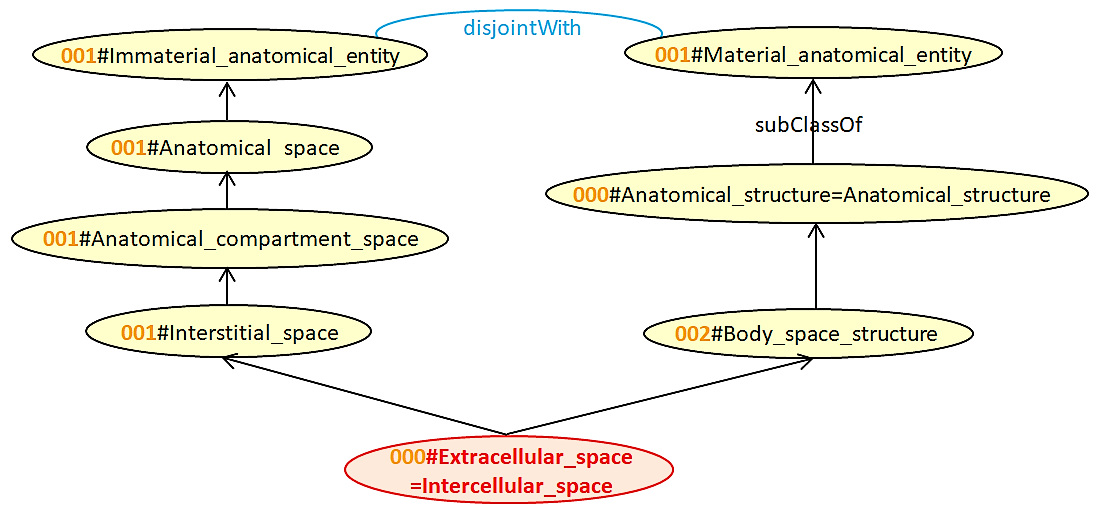}
\caption[Formation d'une classe insatisfiable 3]{Formation d'une classe insatisfiable après une fusion complète}
\label{fusion}
\end{figure}

Dans la figure \ref{fusion}, nous modélisons la représentation graphique des axiomes de justification ci-dessus, où la classe en rouge est la classe insatisfiable.

\newpage
Nous remarquons qu’après la fusion des classes "001\#Extracellular\_space" provenant de FMA 1 (Ont\textbf{1}) et "002\#Intercellular\_space" provenant de SNOMED 1 (Ont\textbf{2}), la classe "000\#Extracellular\_space=Intercellular\_space" devient par inférence sous-classe des deux classes "001\#Immaterial\_anatomical\_entity" et "001\#Material\_anatomical\_entity" qui sont disjointes (information extraite de l’ontologie FMA 1 (00\textbf{1})). Ceci est contradictoire, car une classe ne peut pas être sous-classe de deux classes disjointes.

\section*{Conclusion}
Enfin, nous déduisons que si les ontologies modélisent des vues différentes et incompatibles du même domaine, il est impossible de les intégrer aveuglément et d'assurer à la fois la complétude et la cohérence dans l'ontologie résultante. Dans un contexte d'intégration d'ontologies, assurer la cohérence et la consistance est une priorité car l'ontologie résultante doit être logiquement correcte pour être réellement utile. Dans ce cas, nous ne pourrons jamais réaliser une interopérabilité sémantique complète entre les ontologies d’entrée car nous serons dans l'obligation d'abandonner des correspondances sémantiques. Et au cas où nous souhaiterions atteindre la complétude, les incompatibilités des ontologies sont insolvables automatiquement et l'intervention d'un expert devient nécessaire, ce qui est impossible pour les grandes ontologies. Dans l’ensemble, l’intégration des ontologies ayant des vues incompatibles reste toujours un problème ouvert.\medskip

Dans ce chapitre, nous avons présenté les expérimentations sur notre nouvelle approche d’intégration des ontologies OIA2R. L’analyse des résultats a montré les performances de notre approche et la validité de l’ontologie résultante. En effet, OIA2R a produit des résultats encourageants dans des temps minimes. De même, les expérimentations ont montré une possibilité d’amélioration de ces résultats pour avoir une qualité optimale de l’ontologie de sortie.
\chapter*{Conclusion générale}
\addcontentsline{toc}{chapter}{Conclusion générale}

\markboth{CONCLUSION}{}
Les services Web et les moteurs de recherche peuvent améliorer leurs performances dans l'échange des informations et la précision des résultats de recherche en exploitant la représentation sémantiquement enrichie des informations qu'ils partagent à travers les ontologies.\medskip\vspace{4px}

Actuellement, la recherche et le développement dans le domaine du Web sémantique (qui est un Web distribué et ouvert) ont atteint un stade où un grand nombre d'applications et de services tels que le commerce électronique, le renseignement gouvernemental, la médecine, la fabrication, \textit{etc}, sont alimentés par des ontologies de toute taille, développées par différentes personnes, groupes de recherche, ou organisations, et contenant beaucoup de chevauchements (similarités) sémantiques entre elles. En effet, les différentes sources de données modélisent leurs ontologies de différentes manières selon leurs propres besoins, exigences, et buts, et n’utilisent pas nécessairement des ontologies déjà existantes. Par conséquent, il devient difficile de récupérer les informations provenant de différentes sources dans le Web.\medskip\vspace{4px}

Pour une représentation efficace et homogène des domaines de connaissances, il serait alors nécessaire d’intégrer (ou de fusionner) toutes les ontologies pour former de nouvelles ontologies plus complètes et mieux modélisées qui les remplacera.\medskip\vspace{4px}

Dans le \textbf{premier chapitre}, nous avons passé en revue les principales définitions des notions essentielles pour cerner le champ d’étude. Une étude bibliographique sur le \textit{Web Sémantique}, l’\textit{ontologie}, et l’\textit{ingénierie ontologique} a été menée.\medskip\vspace{4px}

Nous avons établi au \textbf{deuxième chapitre} un bilan des définitions et des méthodes existantes qui s’appliquent dans le cadre de l’intégration des ontologies. Notamment, parmi ces méthodes ou définitions, il y a celles qui sont restées uniquement théoriques.\medskip\vspace{4px}

Dans le \textbf{troisième chapitre}, nous avons mis au point une nouvelle méthode qui permet d’intégrer deux ou plusieurs ontologies à l’aide des alignements entre elles, pour générer une nouvelle ontologie qui les englobe.\medskip\vspace{4px}

Dans le \textbf{quatrième chapitre}, nous avons décrit notre environnement de travail et les critères d'évaluation des outils d'intégration, nous avons appliqué et évalué notre approche dans la pratique, et nous avons prouvé qu'elle est générique, efficace et scalable.

\section*{Perspectives :}
Dans nos prochains travaux, nous allons nous projeter sur la fusion des ontologies qui constitue le plus haut niveau d’interopérabilité sémantique entre les ontologies, et cela pour but de minimiser au maximum les erreurs de l'ontologie résultante de ce processus. En effet, la fusion génère toujours moins d’insatisfiablités que l’ontologie de pont que nous avons réalisée dans ce mémoire.\medskip

Nous exploiterons aussi d’autres relations sémantiques à part la relation d’équivalence dans les alignements, telles que la subsomption et la disjonction, pour que l’interopérabilité soit maximale et que toute hétérogénéité soit traitée.\medskip

Suivant l'exemple de FCA-Merge \cite{stumme2001fca}, nous comptons exploiter le domaine de la fouille de données dans nos prochains travaux de fusion ou d'intégration des ontologies. En effet, les contributions de notre laboratoire LIPAH ont atteint un niveau avancé dans ce domaine, ce qui va énormément nous aider. Citons en quelques travaux intéressants : \cite{bouzouita2006garc, gasmi2007extraction, cellier2008parameterized, othman2008yet, hamrouni2008succinct, ayouni2010mining, ayouni2011extracting, brahmi2010mad, brahmi2011towards, hamdi2013trust, bouker2014mining}\medskip

Il faut également noter qu’un autre axe de recherche très intéressant et plus difficile se présente. C’est le domaine de réparation, de debugage, ou de révision des ontologies et des alignements. Ce domaine aidera énormément à avoir des ontologies de bonne qualité suite à l'intégration des ontologies.

\pagenumbering{gobble}
\bibliographystyle{apalike}
\bibliography{biblio}

\begin{thebibliography}{}

\bibitem[Abbas and Berio, 2013]{abbas2013creating}
Abbas, M.~A. and Berio, G. (2013).
\newblock {C}reating {O}ntologies {u}sing {O}ntology {M}appings: {C}ompatible
  and {I}ncompatible {O}ntology {M}appings.
\newblock In {\em Web Intelligence (WI) and Intelligent Agent Technology (IAT)
  Workshops}, pages 143--146. IEEE.

\bibitem[Abels et~al., 2005]{abels2005identification}
Abels, S., Haak, L., and Hahn, A. (2005).
\newblock Identification of common methods used for ontology integration tasks.
\newblock In {\em Proceedings of the first international workshop on
  Interoperability of heterogeneous information systems}, pages 75--78. ACM.

\bibitem[Alaya et~al., 2015a]{alaya2015predicting}
Alaya, N., {Ben Yahia}, S., and Lamolle, M. (2015a).
\newblock Predicting the {E}mpirical {R}obustness of the {O}ntology {R}easoners
  based on {M}achine {L}earning {T}echniques.
\newblock In {\em the International Conference on Knowledge Engineering and
  Ontology Development {KEOD}}, pages 61--73. SciTePress.

\bibitem[Alaya et~al., 2015b]{alaya2015ranking}
Alaya, N., {Ben Yahia}, S., and Lamolle, M. (2015b).
\newblock Ranking with {T}ies of {OWL} {O}ntology {R}easoners {B}ased on
  {L}earned {P}erformances.
\newblock In {\em the 7th International Joint Conference on Knowledge
  Discovery, Knowledge Engineering and Knowledge Management, {IC3K}}, volume
  631, pages 234--259. Springer.

\bibitem[Alaya et~al., 2015c]{alaya2015towards}
Alaya, N., {Ben Yahia}, S., and Lamolle, M. (2015c).
\newblock Towards {U}nveiling the {O}ntology {K}ey {F}eatures {A}ltering
  {R}easoner {P}erformances.
\newblock {\em CoRR}, abs/1509.08717.

\bibitem[Alaya et~al., 2015d]{alaya2015makes}
Alaya, N., {Ben Yahia}, S., and Lamolle, M. (2015d).
\newblock What {M}akes {O}ntology {R}easoning so {A}rduous?: {U}nveiling the
  key ontological features.
\newblock In {\em the 5th International Conference on Web Intelligence, Mining
  and Semantics, {WIMS}}, pages 4:1--4:12. {ACM}.

\bibitem[Alaya et~al., 2016]{alaya2016raksor}
Alaya, N., {Ben Yahia}, S., and Lamolle, M. (2016).
\newblock Rak{SOR}: {R}anking of {O}ntology {R}easoners {B}ased on {P}redicted
  {P}erformances.
\newblock In {\em the 28th {IEEE} International Conference on Tools with
  Artificial Intelligence, {ICTAI}}, pages 1076--1083. {IEEE}.

\bibitem[Alaya et~al., 2017]{alaya2017multi}
Alaya, N., Lamolle, M., and {Ben Yahia}, S. (2017).
\newblock Multi-{L}abel {B}ased {L}earning for {B}etter {M}ulti-{C}riteria
  {R}anking of {O}ntology {R}easoners.
\newblock In {\em International Semantic Web Conference}, pages 3--19.
  Springer.

\bibitem[Ayouni et~al., 2011]{ayouni2011extracting}
Ayouni, S., Ben~Yahia, S., and Laurent, A. (2011).
\newblock Extracting compact and information lossless sets of fuzzy association
  rules.
\newblock {\em Fuzzy Sets and Systems}, 183(1):1--25.

\bibitem[Ayouni et~al., 2010]{ayouni2010mining}
Ayouni, S., Laurent, A., Ben~Yahia, S., and Poncelet, P. (2010).
\newblock Mining closed gradual patterns.
\newblock In {\em International Conference on Artificial Intelligence and Soft
  Computing {(ICAISC)} {(1)}}, pages 267--274. Springer.

\bibitem[Bail, 2013]{greycite33921}
Bail, S. (2013).
\newblock Common reasons for ontology inconsistency.
\newblock \url{http://ontogenesis.knowledgeblog.org/1343}.
\newblock [Online; accessed 18-Feb-2018].

\bibitem[Bouker et~al., 2014]{bouker2014mining}
Bouker, S., Saidi, R., Ben~Yahia, S., and Mephu~Nguifo, E. (2014).
\newblock Mining undominated association rules through interestingness
  measures.
\newblock {\em International Journal on Artificial Intelligence Tools},
  23(04):1460011.

\bibitem[Bouzouita et~al., 2006]{bouzouita2006garc}
Bouzouita, I., Elloumi, S., and Ben~Yahia, S. (2006).
\newblock {GARC: A} new associative classification approach.
\newblock In {\em International Conference on Data Warehousing and Knowledge
  Discovery {(DaWaK)}}, pages 554--565. Springer.

\bibitem[Brahmi et~al., 2011]{brahmi2011towards}
Brahmi, I., Ben~Yahia, S., Aouadi, H., and Poncelet, P. (2011).
\newblock Towards a multiagent-based distributed intrusion detection system
  using data mining approaches.
\newblock In {\em International Workshop on Agents and Data Mining Interaction
  {(ADMI)}}, pages 173--194. Springer.

\bibitem[Brahmi et~al., 2010]{brahmi2010mad}
Brahmi, I., Ben~Yahia, S., and Poncelet, P. (2010).
\newblock {MAD-IDS}: novel intrusion detection system using mobile agents and
  data mining approaches.
\newblock In {\em Pacific-Asia Workshop on Intelligence and Security
  Informatics {(PAISI)}}, pages 73--76. Springer.

\bibitem[Caldarola and Rinaldi, 2016]{caldarola2016}
Caldarola, E.~G. and Rinaldi, A.~M. (2016).
\newblock {A}n {A}pproach to {O}ntology {I}ntegration for {O}ntology {R}euse.
\newblock In {\em the 17th {IEEE} International Conference on Information Reuse
  and Integration, {IRI} 2016}, pages 384--393. IEEE.

\bibitem[Calvanese et~al., 2001]{calvanese2001framework}
Calvanese, D., De~Giacomo, G., and Lenzerini, M. (2001).
\newblock A framework for ontology integration.
\newblock In {\em Proceedings of the First International Conference on Semantic
  Web Working Symposium, SWWS'01}, pages 303--316. CEUR-WS.org.

\bibitem[Cellier et~al., 2008]{cellier2008parameterized}
Cellier, P., Ferr{\'e}, S., Ridoux, O., and Ducasse, M. (2008).
\newblock A parameterized algorithm to explore formal contexts with a taxonomy.
\newblock {\em International Journal of Foundations of Computer Science},
  19(02):319--343.

\bibitem[Chatterjee et~al., 2017]{chatterjee2017ontology}
Chatterjee, N., Kaushik, N., Gupta, D., and Bhatia, R. (2017).
\newblock {O}ntology {M}erging: {A} {P}ractical {P}erspective.
\newblock In {\em International Conference on Information and Communication
  Technology for Intelligent Systems}, pages 136--145. Springer.

\bibitem[Cheatham and Pesquita, 2017]{cheatham2017semantic}
Cheatham, M. and Pesquita, C. (2017).
\newblock {S}emantic {D}ata {I}ntegration.
\newblock In {\em Handbook of Big Data Technologies}, pages 263--305. Springer.

\bibitem[Choi et~al., 2006]{choi2006survey}
Choi, N., Song, I.-Y., and Han, H. (2006).
\newblock A survey on ontology mapping.
\newblock {\em ACM SIGMOD Record}, 35(3):34--41.

\bibitem[De~Bruijn et~al., 2006]{de2006ontology}
De~Bruijn, J., Ehrig, M., Feier, C., Mart{\'\i}n-Recuerda, F., Scharffe, F.,
  and Weiten, M. (2006).
\newblock {O}ntology mediation, merging and aligning.
\newblock {\em Semantic Web Technologies}, pages 95--113.

\bibitem[Djeddi et~al., 2015]{djeddi2015xmap}
Djeddi, W.~E., Khadir, M.~T., and {Ben Yahia}, S. (2015).
\newblock X{M}ap: results for {OAEI} 2015.
\newblock In {\em the 10th International Workshop on Ontology Matching (OM)
  collocated with the 14th International Semantic Web Conference {(ISWC)}},
  volume 1545, pages 216--221. CEUR-WS.org.

\bibitem[{El Abdi} et~al., 2015]{el2015clona}
{El Abdi}, M., Souid, H., Kachroudi, M., and {Ben Yahia}, S. (2015).
\newblock {CLONA} results for {OAEI} 2015.
\newblock In {\em the 10th International Workshop on Ontology Matching
  collocated with the 14th International Semantic Web Conference {(ISWC)}},
  volume 1545, pages 124--129. CEUR-WS.org.

\bibitem[Euzenat and Shvaiko, 2013]{euzenat2007ontology}
Euzenat, J. and Shvaiko, P. (2013).
\newblock {\em Ontology Matching, Second Edition}.
\newblock Springer.

\bibitem[Fahad et~al., 2010]{fahad2010disjoint}
Fahad, M., Moalla, N., Bouras, A., Qadir, M.~A., and Farukh, M. (2010).
\newblock {D}isjoint-{K}nowledge {A}nalysis and {P}reservation in {O}ntology
  {M}erging {P}rocess.
\newblock In {\em The Fifth International Conference on Software Engineering
  Advances, {ICSEA}}, pages 422--428. IEEE.

\bibitem[Flouris et~al., 2006]{flouris2006classification}
Flouris, G., Plexousakis, D., and Antoniou, G. (2006).
\newblock {A} {C}lassification of {O}ntology {C}hange.
\newblock In {\em Semantic Web Applications and Perspectives {(SWAP)}}, volume
  201. CEUR-WS.org.

\bibitem[Gasmi et~al., 2007]{gasmi2007extraction}
Gasmi, G., Ben~Yahia, S., Nguifo, E.~M., and Bouker, S. (2007).
\newblock Extraction of association rules based on literalsets.
\newblock In {\em International Conference on Data Warehousing and Knowledge
  Discovery {(DaWaK)}}, pages 293--302. Springer.

\bibitem[Gruber et~al., 2009]{gruber2009encyclopedia}
Gruber, T., Ontology, I. L.~L., and {\"O}zsu, M.~T. (2009).
\newblock Encyclopedia of {D}atabase {S}ystems.
\newblock {\em Ontology}.

\bibitem[Hamdi et~al., 2013]{hamdi2013trust}
Hamdi, S., Bouzeghoub, A., Gancarski, A.~L., and Ben~Yahia, S. (2013).
\newblock Trust inference computation for online social networks.
\newblock In {\em 12th IEEE International Conference on Trust, Security and
  Privacy in Computing and Communications (TrustCom)}, pages 210--217. IEEE.

\bibitem[Hamdi et~al., 2012]{hamdi2012enriching}
Hamdi, S., Gan{\c{c}}arski, A.~L., Bouzeghoub, A., and {Ben Yahia}, S. (2012).
\newblock Enriching the {DB}pedia {O}ntology with {S}hared {C}onceptualizations
  from {F}olksonomies.
\newblock In {\em the 36th Annual {IEEE} Computer Software and Applications
  Conference, {COMPSAC}}, pages 551--556. {IEEE}.

\bibitem[Hamdi et~al., 2015]{hamdi2015linking}
Hamdi, S., Gancarski, A.~L., Bouzeghoub, A., and {Ben Yahia}, S. (2015).
\newblock Linking trust in social networks with the semantic web: {FOAF} case.
\newblock In {\em International Machine Learning and Data Analytics Symposium
  (MLDAS)}.

\bibitem[Hamrouni et~al., 2008]{hamrouni2008succinct}
Hamrouni, T., Ben~Yahia, S., and Mephu~Nguifo, E. (2008).
\newblock Succinct minimal generators: Theoretical foundations and
  applications.
\newblock {\em International journal of foundations of computer science},
  19(02):271--296.

\bibitem[Heflin and Hendler, 2000]{heflin2000dynamic}
Heflin, J. and Hendler, J. (2000).
\newblock Dynamic {O}ntologies on the {W}eb.
\newblock In {\em AAAI/IAAI}, pages 443--449.

\bibitem[Jim{\'e}nez-Ruiz and Grau, 2011]{jimenez2011logmap}
Jim{\'e}nez-Ruiz, E. and Grau, B.~C. (2011).
\newblock {LogMap:} {L}ogic-based and {S}calable {O}ntology {M}atching.
\newblock In {\em the 10th International Semantic Web Conference (ISWC), Part
  {I}}, pages 273--288. Springer.

\bibitem[Jim{\'{e}}nez{-}Ruiz et~al., 2009]{jimenez2009ontology}
Jim{\'{e}}nez{-}Ruiz, E., Grau, B.~C., Horrocks, I., and Llavori, R.~B. (2009).
\newblock {O}ntology {I}ntegration {U}sing {M}appings: {T}owards {G}etting the
  {R}ight {L}ogical {C}onsequences.
\newblock In {\em the 6th European Semantic Web Conference {(ESWC)}}, pages
  173--187. Springer.

\bibitem[Kachroudi et~al., 2012]{kachroudi2012exploring}
Kachroudi, M., {Ben Yahia}, S., and Zghal, S. (2012).
\newblock Exploring {OWL-DL} {P}rimitives for {A}utomatic {O}ntology
  {A}lignment {T}ools {P}arametrization.
\newblock In {\em Eighth International Conference on Semantics, Knowledge and
  Grids {(SKG)}}, pages 261--264. {IEEE}.

\bibitem[Kachroudi et~al., 2015]{kachroudi2015f}
Kachroudi, M., Chelbi, A., Souid, H., and {Ben Yahia}, S. (2015).
\newblock Fiona: {A} {F}ramework for {I}ndirect {O}ntology {A}lignment.
\newblock In {\em the 22nd International Symposium on Methodologies for
  Intelligent Systems, {ISMIS}}, volume 9384, pages 224--229. Springer.

\bibitem[Kachroudi et~al., 2016]{kachroudi2016initiating}
Kachroudi, M., Diallo, G., and {Ben Yahia}, S. (2016).
\newblock Initiating cross-lingual ontology alignment with information
  retrieval techniques.
\newblock In {\em 6{\'e}mes Journ{\'e}es Francophones sur les Ontologies
  (JFO)}.

\bibitem[Kachroudi et~al., 2017a]{kachroudi2017oaei}
Kachroudi, M., Diallo, G., and {Ben Yahia}, S. (2017a).
\newblock {OAEI} 2017 results of {KEPLER}.
\newblock In {\em the 12th International Workshop on Ontology Matching (OM)
  co-located with the 16th International Semantic Web Conference {(ISWC)}},
  volume 2032, pages 138--145. CEUR-WS.org.

\bibitem[Kachroudi et~al., 2017b]{kachroudi2017composition}
Kachroudi, M., Diallo, G., and {Ben Yahia}, S. (2017b).
\newblock On the composition of large biomedical ontologies alignment.
\newblock In {\em the 7th International Conference on Web Intelligence, Mining
  and Semantics, {WIMS}}, pages 24:1--24:10. {ACM}.

\bibitem[Kachroudi et~al., 2011]{kachroudi2011ldoa}
Kachroudi, M., Moussa, E.~B., Zghal, S., and {Ben Yahia}, S. (2011).
\newblock {LDOA} results for {OAEI} 2011.
\newblock In {\em the 6th International Workshop on Ontology Matching (OM)},
  volume 814. CEUR-WS.org.

\bibitem[Kachroudi et~al., 2013a]{kachroudi2013ontopart}
Kachroudi, M., Zghal, S., and {Ben Yahia}, S. (2013a).
\newblock Onto{P}art: at the cross-roads of ontology partitioning and scalable
  ontology alignment systems.
\newblock {\em International Journal of Metadata, Semantics and Ontologies
  {(IJMSO)}}, 8(3):215--225.

\bibitem[Kachroudi et~al., 2013b]{kachroudi2013parametrage}
Kachroudi, M., Zghal, S., and {Ben Yahia}, S. (2013b).
\newblock Param{\'{e}}trage intelligent de l'alignement d'ontologies par
  l'int{\'{e}}grale de choquet.
\newblock In {\em Extraction et gestion des connaissances (EGC'13)}, pages
  377--382.

\bibitem[Kachroudi et~al., 2013c]{kachroudi2013using}
Kachroudi, M., Zghal, S., and {Ben Yahia}, S. (2013c).
\newblock Using {L}inguistic {R}esource for {C}ross-{L}ingual {O}ntology
  {A}lignment.
\newblock {\em International Journal of Recent Contributions from Engineering,
  Science \& IT {(iJES)}}, 1(1):21--27.

\bibitem[Kachroudi et~al., 2014]{kachroudi2014bridging}
Kachroudi, M., Zghal, S., and {Ben Yahia}, S. (2014).
\newblock Bridging the multilingualism gap in ontology alignment.
\newblock {\em International Journal of Metadata, Semantics and Ontologies
  {(IJMSO)}}, 9(3):252--262.

\bibitem[Kalfoglou and Schorlemmer, 2003]{kalfoglou2003ontology}
Kalfoglou, Y. and Schorlemmer, W.~M. (2003).
\newblock {O}ntology {M}apping: the state of the art.
\newblock {\em The knowledge engineering review}, 18(1):1--31.

\bibitem[Kamoun and {Ben Yahia}, 2012a]{kamoun2012automatic}
Kamoun, K. and {Ben Yahia}, S. (2012a).
\newblock Automatic {A}pproach for {O}ntology {E}volution based on {S}tability
  {E}valuation.
\newblock In {\em the 8th International Conference on Web Information Systems
  and Technologies {WEBIST}}, pages 452--455. SciTePress.

\bibitem[Kamoun and {Ben Yahia}, 2012b]{kamoun2012information}
Kamoun, K. and {Ben Yahia}, S. (2012b).
\newblock Information content similarity measure to assess stability during
  ontology enrichment.
\newblock {\em International Review on Computers and Software}, 7(3).

\bibitem[Kamoun and {Ben Yahia}, 2012c]{kamoun2012novel}
Kamoun, K. and {Ben Yahia}, S. (2012c).
\newblock A novel global measure approach based on ontology spectrum to
  evaluate ontology enrichment.
\newblock {\em Complexity}, 39(17).

\bibitem[Kamoun and {Ben Yahia}, 2014]{kamoun2014stability}
Kamoun, K. and {Ben Yahia}, S. (2014).
\newblock Stability {A}ssess {B}ased on {E}nhanced {I}nformation {C}ontent
  {S}imilarity {M}easure for {O}ntology {E}nrichment.
\newblock In {\em the 4th International Conference on Model and Data
  Engineering, {MEDI}}, volume 8748, pages 146--153. Springer.

\bibitem[Kamoun et~al., 2010]{kamoun2010evolution}
Kamoun, K., Harzallah, M., Kuntz, P., and {Ben Yahia}, S. (2010).
\newblock Evolution d'ontologies: revue et critiques.
\newblock In {\em Atelier Recherche d'informations personnalis{\'e}es sur le
  Web, Conf{\'e}rence EGC'10}.

\bibitem[Keet, 2004]{keet2004aspects}
Keet, C.~M. (2004).
\newblock Aspects of ontology integration.
\newblock {\em The PhD Proposal, School of Computing, Napier University,
  Scotland}.

\bibitem[Klein, 2001]{klein2001combining}
Klein, M. (2001).
\newblock Combining and relating ontologies: an analysis of problems and
  solutions.
\newblock In {\em IJCAI-01 Workshop on Ontologies and Information Sharing},
  pages 53--62. CEUR-WS.org.

\bibitem[Kokla, 2006]{kokla2006guidelines}
Kokla, M. (2006).
\newblock Guidelines on geographic ontology integration.
\newblock In {\em Proceedings of the ISPRS Technical Commission II Symposium},
  pages 67--72.

\bibitem[Leung et~al., 2014]{leung2014new}
Leung, N.~K., Lau, S.~K., and Tsang, N. (2014).
\newblock A new methodology to streamline ontology integration processes.
\newblock In {\em the Fourth International Conference on Digital Information
  and Communication Technology and its Applications {(DICTAP)}}, pages
  174--179. IEEE.

\bibitem[Malik et~al., 2010]{malik2010ontology}
Malik, S.~K., Prakash, N., and Rizvi, S. (2010).
\newblock Ontology merging using prompt plug-in of prot{\'e}g{\'e} in
  {S}emantic {W}eb.
\newblock In {\em International Conference on Computational Intelligence and
  Communication Networks {(CICN)}}, pages 476--481. IEEE.

\bibitem[McGuinness et~al., 2000]{mcguinness2000chimaera}
McGuinness, D.~L., Fikes, R., Rice, J., and Wilder, S. (2000).
\newblock The {C}himaera {O}ntology {E}nvironment.
\newblock {\em AAAI/IAAI}, 2000:1123--1124.

\bibitem[Meilicke, 2011]{meilicke2011alignment}
Meilicke, C. (2011).
\newblock {\em {A}lignment {I}ncoherence in {O}ntology {M}atching}.
\newblock PhD thesis, Universit{\"a}t Mannheim.

\bibitem[Mena et~al., 1996]{mena1996managing}
Mena, E., Kashyap, V., Illarramendi, A., and Sheth, A.~P. (1996).
\newblock Managing {M}ultiple {I}nformation {S}ources through {O}ntologies:
  {R}elationship between {V}ocabulary {H}eterogeneity and {L}oss of
  {I}nformation.
\newblock In {\em Knowledge Representation Meets Databases {(KRDB)}}.
  CEUR-WS.org.

\bibitem[Neches et~al., 1991]{neches1991enabling}
Neches, R., Fikes, R.~E., Finin, T., Gruber, T., Patil, R., Senator, T., and
  Swartout, W.~R. (1991).
\newblock Enabling {T}echnology for {K}nowledge {S}haring.
\newblock {\em AI magazine}, 12(3):36.

\bibitem[Noy and Musen, 2000]{noy1999algorithm}
Noy, N.~F. and Musen, M.~A. (2000).
\newblock {PROMPT}: Algorithm and {T}ool for {A}utomated {O}ntology {M}erging
  and {A}lignment.
\newblock In {\em {AAAI/IAAI}}, pages 450--455. {AAAI} Press / The {MIT} Press.

\bibitem[Othman and Ben~Yahia, 2008]{othman2008yet}
Othman, L.~B. and Ben~Yahia, S. (2008).
\newblock Yet another approach for completing missing values.
\newblock In {\em Concept Lattices and Their Applications}, pages 155--169.
  Springer.

\bibitem[Pinto, 1999]{pinto1999towards}
Pinto, H.~S. (1999).
\newblock {T}owards {O}ntology {R}euse.
\newblock In {\em Proceedings of AAAI99’s Workshop on Ontology Management,
  WS-99}, volume~13, pages 67--73.

\bibitem[Pinto and Martins, 2004]{pinto2004ontologies}
Pinto, H.~S. and Martins, J.~P. (2004).
\newblock Ontologies: {H}ow can they be built?
\newblock {\em Knowledge and Information Systems}, 6(4):441--464.

\bibitem[Raunich and Rahm, 2012]{raunich2012towards}
Raunich, S. and Rahm, E. (2012).
\newblock {T}owards a {B}enchmark for {O}ntology {M}erging.
\newblock In {\em {OTM} Confederated International Workshops}, pages 124--133.
  Springer.

\bibitem[Raunich and Rahm, 2014]{raunich2014target}
Raunich, S. and Rahm, E. (2014).
\newblock {T}arget-driven {M}erging of {T}axonomies with {ATOM}.
\newblock {\em Information Systems}, 42:1--14.

\bibitem[Sattler et~al., 2013]{greycite33925}
Sattler, U., Stevens, R., and Lord, P. (2013).
\newblock (i can't get no) satisfiability.
\newblock \url{http://ontogenesis.knowledgeblog.org/1329}.
\newblock [Online; accessed 18-Feb-2018].

\bibitem[Sowa, 1997]{sowa1997electronic}
Sowa, J. (1997).
\newblock Electronic communication in the onto-std mailing list.

\bibitem[Studer et~al., 1998]{studer1998knowledge}
Studer, R., Benjamins, V.~R., Fensel, D., et~al. (1998).
\newblock Knowledge {E}ngineering: {P}rinciples and {M}ethods.
\newblock {\em Data and knowledge engineering}, 25(1):161--198.

\bibitem[Stumme and Maedche, 2001]{stumme2001fca}
Stumme, G. and Maedche, A. (2001).
\newblock {FCA-Merge}: Bottom-{U}p {M}erging of {O}ntologies.
\newblock In {\em the 7th International Joint Conference on Artificial
  Intelligence {(IJCAI)}}, pages 225--234. Morgan Kaufmann.

\bibitem[Udrea et~al., 2007]{udrea2007leveraging}
Udrea, O., Getoor, L., and Miller, R.~J. (2007).
\newblock {L}everaging {D}ata and {S}tructure in {O}ntology {I}ntegration.
\newblock In {\em {SIGMOD} International Conference on Management of Data},
  pages 449--460. {ACM}.

\bibitem[Umer and Mundy, 2012]{umer2012semantically}
Umer, Q. and Mundy, D. (2012).
\newblock Semantically {I}ntelligent {S}emi-{A}utomated {O}ntology
  {I}ntegration.
\newblock In {\em Proc. of the World Congress on Eng}.

\bibitem[Visser et~al., 1998]{visser1998assessing}
Visser, P.~R., Jones, D.~M., Bench-Capon, T.~J., and Shave, M.~J. (1998).
\newblock Assessing heterogeneity by classifying ontology mismatches.
\newblock In {\em Proceedings of the FOIS}, volume~98.

\bibitem[Wr{\'o}blewska et~al., 2012]{wroblewska2012methods}
Wr{\'o}blewska, A., Podsiadly-Marczykowska, T., Bembenik, R., Protaziuk, G.,
  and Rybinski, H. (2012).
\newblock Methods and {T}ools for {O}ntology {B}uilding, {L}earning and
  {I}ntegration-{A}pplication in the {SYNAT} {P}roject.
\newblock {\em Intelligent Tools for Building a Scientific Information
  Platform}, 390:121--151.

\bibitem[Zghal, 2010]{zghal2010contributions}
Zghal, S. (2010).
\newblock {\em Contributions {\`a} l'alignement d'ontologies OWL par
  agr{\'e}gation de similarit{\'e}s}.
\newblock PhD thesis, Artois.

\bibitem[Zghal et~al., 2007a]{zghal2007soda}
Zghal, S., {Ben Yahia}, S., Nguifo, E.~M., and Slimani, Y. (2007a).
\newblock {SODA:} an {OWL-DL} based {O}ntology {M}atching {S}ystem.
\newblock In {\em the 2nd International Workshop on Ontology Matching
  {(OM-07)}}, volume 304. CEUR-WS.org.

\bibitem[Zghal et~al., 2011]{zghal2011oacas}
Zghal, S., Kachroudi, M., {Ben Yahia}, S., and Nguifo, E.~M. (2011).
\newblock {OACAS:} results for {OAEI} 2011.
\newblock In {\em the 6th International Workshop on Ontology Matching (OM)},
  volume 814. CEUR-WS.org.

\bibitem[Zghal et~al., 2007b]{zghal2007nouvelle}
Zghal, S., Kamoun, K., {Ben Yahia}, S., and Nguifo, E.~M. (2007b).
\newblock Une nouvelle m{\'{e}}thode d'alignement et de visualisation
  d'ontologies {OWL-L}ite.
\newblock In {\em Extraction et gestion des connaissances (EGC'07)}, volume
  {RNTI-E-9}, pages 197--198.

\bibitem[Zghal et~al., 2007c]{zghal2007edola}
Zghal, S., Kamoun, K., {Ben Yahia}, S., Nguifo, E.~M., and Slimani, Y. (2007c).
\newblock {EDOLA} : Une nouvelle m{\'{e}}thode d'alignement d'ontologies
  owl-lite.
\newblock In {\em COnf{\'{e}}rence en Recherche d'Infomations et Applications -
  {CORIA}}, pages 351--367. Universit{\'{e}} de Saint-{\'{E}}tienne.

\bibitem[Zghal et~al., 2007d]{zghal2007new}
Zghal, S., Nguifo, E.~M., Kamoun, K., {Ben Yahia}, S., and Slimani, Y. (2007d).
\newblock A new alignment method for {OWL-L}ite ontologies using propagation of
  similarity over the graph.
\newblock In {\em the 18th International Workshop on Database and Expert
  Systems Applications {(DEXA)}}, pages 524--528. {IEEE}.

\bibitem[Zhang et~al., 2017]{zhang2017oim}
Zhang, L., Ren, J., and Li, X. (2017).
\newblock {OIM-SM:} {A} method for ontology integration based on semantic
  mapping.
\newblock {\em Journal of Intelligent and Fuzzy Systems}, 32(3):1983--1995.

\bibitem[Zhao and Ichise, 2014]{zhao2014ontology}
Zhao, L. and Ichise, R. (2014).
\newblock {O}ntology {I}ntegration for {L}inked {D}ata.
\newblock {\em Journal on Data Semantics}, 3(4):237--254.

\bibitem[Zhu et~al., 2009]{zhu2009research}
Zhu, L., Yang, Q., and Chen, W. (2009).
\newblock Research on ontology integration combined with machine learning.
\newblock In {\em Second International Conference on Intelligent Computation
  Technology and Automation, ICICTA'09}, volume~1, pages 464--467. IEEE.

\bibitem[Ziemba et~al., 2015]{ziemba2015integration}
Ziemba, P., Jankowski, J., Watrobski, J., Wolski, W., and Becker, J. (2015).
\newblock {I}ntegration of {D}omain {O}ntologies in the {R}epository of
  {W}ebsite {E}valuation {M}ethods.
\newblock In {\em Federated Conference on Computer Science and Information
  Systems (FedCSIS)}, pages 1585--1595. {IEEE}.

\end{thebibliography}

\newpage
\thispagestyle{empty}

\paragraph{\\ \\ Résumé \\ \\}
Ce travail est accompli dans le cadre d’un projet de mémoire de mastère de recherche. Le but est d'intégrer deux ou plusieurs ontologies (de mêmes ou de différents domaines) dans une nouvelle ontologie OWL consistante et cohérente pour assurer leur interopérabilité sémantique. Pour ce faire, nous avons choisi de créer une ontologie de pont qui inclut toutes les ontologies sources et leurs axiomes de pont dans une nouvelle ontologie. Par la suite, nous avons introduit un critère qui aide à obtenir une ontologie de meilleure qualité (ayant le minimum de conflits sémantiques / logiques). Nous avons proposé également une nouvelle terminologie qui clarifie les notions floues et mal placées utilisées dans les travaux de l'état de l'art. Enfin, nous avons testé et évalué notre outil OIA2R à l'aide des ontologies et des alignements de référence de OAEI. Il s'est avéré qu'il est générique, efficace, scalable, et assez performant.\\

\textbf{Mots clés}: Ontologie, Intégration des ontologies, Fusion des ontologies, Matching, Alignement, Mapping, Consistance, Cohérence, Insatisfiabilité, OWL, Réparation des alignements, debugage des alignements.\\ \\ \\ 

\vspace{3pt}\hrule\vspace{6pt}

\paragraph{\\ \\ Abstract \\ \\ }

This work is done as part of a research master's thesis project. The goal is to integrate two or more ontologies (of the same or different domains) in a new consistent and coherent OWL ontology to insure semantic interoperability between them. To do this, we have chosen to create a bridge ontology that includes all source ontologies and their bridging axioms in a new ontology. Subsequently, we introduced a new criterion for obtaining an ontology of better quality (having the minimum of semantic / logical conflicts). We have also proposed a new terminology that clarifies the unclear and misplaced notions used in state-of-the-art works. Finally, we tested and evaluated our OIA2R tool using OAEI ontologies and reference alignments. It turned out that it is generic, efficient, scalable, and powerful enough.\\
 
\textbf{Keywords}: Ontology, Ontology Integration, Ontology Merging, Matching, Alignement, Mapping, Consistency, Coherence, Insatisfiability, OWL, Alignment Repair, Alignment Debugging. \\ \\ \\

\vspace{3pt}\hrule\vspace{6pt}

\end{document}